%% file: main.tex
\documentclass{article}

\PassOptionsToPackage{numbers, compress}{natbib}

\usepackage[final]{neurips_2019}

\usepackage[utf8]{inputenc} 
\usepackage[T1]{fontenc}    
\usepackage{hyperref}       
\usepackage{url}            
\usepackage{booktabs}       
\usepackage{amsfonts}       
\usepackage{nicefrac}       
\usepackage{microtype}      
\usepackage{subcaption}
\usepackage{tikz} 
\usetikzlibrary{arrows, positioning}
\usepackage{algorithm}
\usepackage[noend]{algpseudocode}

\graphicspath{{fig/}}

\title{Ultrametric Fitting by Gradient Descent}

\author{%
Giovanni Chierchia\thanks{Both authors contributed equally.}\\%
Université Paris-Est, LIGM (UMR 8049) \\
CNRS, ENPC, ESIEE Paris, UPEM \\
F-93162, Noisy-le-Grand, France\\
\texttt{giovanni.chierchia@esiee.fr}%
\And Benjamin Perret\footnotemark[1]\\%
Université Paris-Est, LIGM (UMR 8049) \\
CNRS, ENPC, ESIEE Paris, UPEM \\
F-93162, Noisy-le-Grand, France\\
\texttt{benjamin.perret@esiee.fr}%
}

\usepackage{amsmath}%
\usepackage{amssymb}%
\usepackage{xcolor}
\usepackage{array}
\usepackage{enumerate}
\usepackage{stmaryrd}
\usepackage{ifthen}
\usepackage{bbm}
\usepackage{enumitem}

\include{macros}

\usepackage[colorinlistoftodos, textsize=tiny]{todonotes}

\algnewcommand\algorithmicforeach{\textbf{for each}}
\algdef{S}[FOR]{ForEach}[1]{\algorithmicforeach\ #1\ \algorithmicdo}

\begin{document}

\maketitle

\begin{abstract}
We study the problem of fitting an ultrametric distance to a dissimilarity graph in the context of hierarchical cluster analysis. Standard hierarchical clustering methods are specified procedurally, rather than in terms of the cost function to be optimized. We aim to overcome this limitation by presenting a general optimization framework for ultrametric fitting. 
Our approach consists of modeling the latter as a constrained optimization problem over the continuous space of ultrametrics. 
So doing, we can leverage the simple, yet effective, idea of replacing the ultrametric constraint with a min-max operation injected directly into the cost function. 
The proposed reformulation leads to an unconstrained optimization problem that can be efficiently solved by gradient descent methods. 
The flexibility of our framework allows us to investigate several cost functions, following the classic paradigm of combining a data fidelity term with a regularization.
While we provide no theoretical guarantee to find the global optimum, the numerical results obtained  over a number of synthetic and real datasets demonstrate the good performance of our approach with respect to state-of-the-art agglomerative algorithms. 
This makes us believe that the proposed framework sheds new light on the way to design a new generation of hierarchical clustering methods. Our code is made publicly available at \url{https://github.com/PerretB/ultrametric-fitting}.
\end{abstract}

\section{Introduction}
\input{intro.tex}

\section{Ultrametric fitting}
\input{proposed.tex}

\section{Algorithms}\label{sec:algo}
\input{algo.tex}

\section{Experiments}
\input{closest.tex}

\input{clustering.tex}

\section{Conclusion}
\input{conclusion.tex}

\section{Acknowledgment}
This work was partly supported by the INS2I JCJC project under grant 2019OSCI. We are deeply grateful to Julian Yarkony and Charless Fowlkes for sharing their code, and to Fred Hamprecht for many insightful discussions.

\appendix
\section{Supplemental material}
\input{supplemental.tex}

\bibliographystyle{unsrtnat}
\bibliography{biblio}

\end{document}

%% file: macros.tex
\newcommand*{\V}{\ensuremath{V}}
\newcommand*{\E}{\ensuremath{E}}

\newcommand*{\G}{\mathcal{G}}
\newcommand*{\w}{w}

\newcommand*{\weightfuns}{\mathcal{W}}
\newcommand*{\Cycles}{\mathcal{C}}
\newcommand*{\Paths}{\mathcal{P}}
\newcommand*{\minmax}{\Phi}
\newcommand*{\mstEdge}[1]{\sigma(#1)}
\newcommand*{\ancestors}{\mathcal{A}}
\newcommand*{\children}{\rm Children}
\newcommand*{\otherChild}{c}
\newcommand*{\alt}{\textrm{alt}}
\newcommand*{\lca}{\textrm{lca}}

\newcommand*{\SL}{\textrm{SL}}
\definecolor{color1}{HTML}{D73027}
\definecolor{color2}{HTML}{FC8D59}
\definecolor{color3}{HTML}{FEE090}
\definecolor{color4}{HTML}{E0F3F8}
\definecolor{color5}{HTML}{91BFDB}
\definecolor{color6}{HTML}{4575B4}

\definecolor{rouge}{rgb}{1.0,0.0,0.0}
\definecolor{blue}{rgb}{0.0,0.0,1.0}

\newcommand*{\ie}{\textit{i.e.},}

\newcommand*\FinalRevision{false}
\newcommand*{\REMM}[1]{\ifthenelse{\equal{\FinalRevision}{false}}{\fbox{\begin{minipage}{0.4\textwidth}\textsl{\textcolor{rouge}{#1}}\end{minipage}}}{#1}}
\newcommand*{\REM}[1]{\ifthenelse{\equal{\FinalRevision}{false}}{\textbf{\uline{\textcolor{rouge}{#1}}}}{}}
\newcommand*{\TODO}[1]{\ifthenelse{\equal{\FinalRevision}{false}}{\fbox{\begin{minipage}{0.4\textwidth}\textsl{\textcolor{blue}{#1}}\end{minipage}}}{#1}}
\newcommand*{\rev}[1]{\ifthenelse{\equal{\FinalRevision}{false}}{\textbf{\textcolor{blue}{#1}}}{}}
\newcolumntype{M}[1]{>{\centering}m{#1}}

\newenvironment{compactenumerate}
{ \begin{enumerate}
    \setlength{\itemsep}{0pt}
    \setlength{\parskip}{0pt}
    \setlength{\parsep}{0pt}     }
{ \end{enumerate}                  } 

\newcommand*{\Cite}[1]{~\cite{#1}}

\newcommand*{\Eq}[1]{~\eqref{#1}}
\newcommand*{\Fig}[1]{Figure~\ref{#1}}

\providecommand{\RR}{\mathbb{R}}

\newcommand*{\set}[1]{\left\{#1\right\}}

\DeclareMathOperator{\card}{card}



%% file: intro.tex
Ultrametrics provide a natural way to describe a recursive partitioning of data into increasingly finer clusters, also known as hierarchical clustering \cite{Murtagh2012}. Ultrametrics are intuitively represented by dendrograms, \ie{} rooted trees whose leaves correspond to data points, and whose internal nodes represent the clusters of its descendant leaves. In topology, this corresponds to a metric space in which the usual triangle inequality is strengthened by the ultrametric inequality, so that every triple of points forms an isosceles triangle, with the two equal sides at least as long as the third side. The main question investigated in this article is: <<~\emph{How well can we construct an ultrametric to fit the given dissimilarity data?} >> This is what we refer to as ultrametric fitting.

Ultrametric fitting can be traced back to the early work on numerical taxonomy \cite{Sneath1962} in the context of phylogenetics \cite{Felsenstein2003}. Several well-known algorithms originated in this field, such as single linkage \cite{Gower1969}, average linkage \cite{Jardine1968}, and Ward method \cite{Ward1963}. Nowadays, there exists a large literature on ultrametric fitting, which can be roughly divided in four categories: agglomerative and divisive greedy heuristics \cite{deAmorim2015, Ackerman2016, Dasgupta2016, kobren2017, CohenAddadNIPS2017, Chehreghani2018, Bonald2018}, integer linear programming \cite{YarkonyNIPS2015, DISUMMA2015, RoyNIPS2016}, continuous relaxations \cite{DeSoete1984, Ailon2011, Charikar2017, monath2017}, and probabilistic formulations \cite{Hartigan1985, Neal2003, vikram2016}. Our work belongs to the family of continuous relaxations.

The most popular methods for ultrametric fitting probably belong to the family of agglomerative heuristics. They follow a bottom-up approach, in which the given dissimilarity data are sequentially merged through some specific strategy. But since the latter is specified procedurally, it is usually hard to understand the objective function being optimized. In this regard, several recent works \cite{Dasgupta2016, RoyNIPS2016, Charikar2017, CohenAddadNIPS2017, CohenAddad2018} underlined the importance to cast ultrametric fitting as an optimization problem with a well-defined cost function, so as to better understand how the ultrametric is built. 

Recently, Dasgupta \cite{Dasgupta2016} introduced a cost function for evaluating an ultrametric, and proposed an heuristic to approximate its optimal solution. The factor of this approximation was later improved by several works, based on a linear programming relaxation \cite{RoyNIPS2016}, a semidefinite programming relaxation \cite{Charikar2017}, or a recursive $\phi$-sparsest cut algorithm \cite{CohenAddad2018}. Along similar lines, it was shown that average linkage provides a good approximation of the optimal solution to Dasgupta cost function \cite{Moseley2017, Charikar2019}. Closer to our approach, a differentiable relaxation inspired by Dasgupta cost function was also proposed \cite{monath2017}. Moreover, a regularization for Dasgupta cost function was formulated in the context of semi-supervised clustering \cite{vikram2016, Chatziafratis2018}, based on triplet constraints provided by the user.  

More generally, the problem of finding the closest ultrametric to dissimilarity data was extensively studied through linear programming relaxations \cite{Ailon2011} and integer linear programming \cite{DISUMMA2015}. A special case of interest arises when the dissimilarities are specified by a planar graph, which is a natural occurrence in image segmentation \cite{MALIS2009, ArbelaezPAMI2011,ManinisPAMI2018,DEEPMALIS}. By exploiting the planarity of the input graph, a tight linear programming relaxation can be derived from the minimum-weight multi-cut problem \cite{YarkonyNIPS2015}. There exist many other continuous relaxations of discrete problems in the specific context of image segmentation \cite{Ishikawa2003, PockECCV2008, PockCVPR2009, PockECCV2009, MollenhoffCVPR2016, Foare2018}, but they typically aim at a flat representation of data, rather than hierarchical.

\textbf{Contribution.} 
We propose a general optimization framework for ultrametric fitting based on gradient descent.
Our approach consists of optimizing a cost function over the continuous space of ultrametrics, where the ultrametricity constraint is implicitly enforced by a min-max operation. 
We demonstrate the versatility of our approach by investigating several cost functions:
\vspace{-0.75em}
\begin{compactenumerate}
    \item the \emph{closest-ultrametric} fidelity term, which expresses that the fitted ultrametric should be close to the given dissimilarity graph;
    \item the \emph{cluster-size} regularization, which penalizes the presence of small clusters in the upper levels of the associated hierarchical  clustering;
    \item the \emph{triplet} regularization for semi-supervised learning, which aims to minimize the intra-class distance and maximize the inter-class distance;
    \item the \emph{Dasgupta} fidelity term, which is a continuous relaxation of Dasgupta cost function expressing that the fitted ultrametric should associate large dissimilarities to large clusters.
\end{compactenumerate}
\vspace{-0.75em}
We devise efficient algorithms with automatic differentiation in mind, and we show that they scale up to millions of vertices on sparse graphs. Finally, we evaluate the proposed cost functions on synthetic and real datasets, and we show that they perform as good as Ward method and semi-supervised SVM.

%% file: proposed.tex
\begin{figure}
	\centering
	\begin{subfigure}[b]{0.3\textwidth}
		\begin{equation*}
			\bordermatrix{%
				& x_1 & x_2 & x_3 & x_4 \cr
				x_1 &  0  & r_1 & r_3 & r_3 \cr
				x_2 & r_1 &  0  & r_3 & r_3 \cr
				x_3 & r_3 & r_3 &  0  & r_2 \cr
				x_4 & r_3 & r_3 & r_2 &  0  \cr
			}
		\end{equation*}
		\caption{Ultrametric}\label{fig:ultrametric}
	\end{subfigure}
	\hfill
	\begin{subfigure}[b]{0.37\textwidth}
		\centering
		\begin{tikzpicture}[sloped]
		\tikzstyle{dot}=[circle,draw,fill, inner sep=0pt, minimum size=4pt]
		\node (a) at (-1.5,0) {$x_1$};
		\node (b) at (-0.5,0) {$x_2$};
		\node (c) at ( 0.5,0) {$x_3$};
		\node (d) at ( 1.5,0) {$x_4$};
		\node[dot] (ab) at (-1,.65) {};
		\node[dot] (cd) at ( 1,1.25) {};
		\node[dot] (root) at (0,1.8) {};
		\draw  (a) |- (ab.center);
		\draw  (b) |- (ab.center);
		\draw  (c) |- (cd.center);
		\draw  (d) |- (cd.center);
		\draw  (cd.center) |- (root.center);
		\draw  (ab.center) |- (root.center);
		\draw[->,-triangle 60] (-2,0) -- node[above]{} (-2,2.2);
		\node[left] (r0) at (-2,0) {\small$0$};
		\node[left] (r1) at (-2,.65) {\small$r_1$};
		\node[left] (r2) at (-2,1.25) {\small$r_2$};
		\node[left] (r3) at (-2,1.8) {\small$r_3$};
		\draw[dashed,thin] (-1.35,.65) -- (r1);
		\draw[dashed,thin] ( 0.4,1.25) -- (r2);
		\draw[dashed,thin] (-0.9,1.8) -- (r3);
		\node[left] at (-2,2.2) {\scriptsize altitude};
		\node[above right] at (ab) {\small $n_1$}; 
		\node[above right] at (cd) {\small $n_2$}; 
		\node[above] at (root) {\small $n_3$};
		\end{tikzpicture}
		\caption{Dendrogram}\label{fig:dendrogram}
	\end{subfigure}
	\hfill
	\begin{subfigure}[b]{0.3\textwidth}
		\centering
		\tikzstyle{edge}=[thick]
		\tikzstyle{vertex}=[circle, draw, fill=black!5, text depth=0em, inner sep=2.5, minimum width=0.5cm, minimum height=0.5cm]
		\tikzstyle{weight}=[midway, anchor=center, fill=white]
		\begin{tikzpicture}[auto, node distance=1.25cm]
		\node[vertex] (x1) {$x_1$};
		\node[vertex, right=1.8cm of x1] (x2) {$x_2$};
		\node[vertex, below= of x1] (x3) {$x_3$};
		\node[vertex, below= of x2] (x4) {$x_4$};
		\draw[edge] (x1) -- (x2) node[weight] {$r_1$};
		\draw[edge] (x2) -- (x3) node[weight] {$r_3$};
		\draw[edge] (x3) -- (x4) node[weight] {$r_2$};
		\draw[edge] (x2) -- (x4) node[weight] {$r_3$};
		\end{tikzpicture}
		\caption{Ultrametric on a graph}
	\end{subfigure}
	\caption{Ultrametric $d$ on $\{x_1,x_2,x_3,x_4\}$ given by the dissimilarity matrix (a), and represented by the dendrogram (b) and the graph (c). Two elements $x_i$ and $x_j$ merge at the altitude $r_k=d(x_i,x_j)$ in the dendrogram, and the corresponding node is the lowest common ancestor (l.c.a.) of $x_i$ and $x_j$. For example, the l.c.a.\ of $x_1$ and $x_2$ is the node $n_1$ at altitude $r_1$, hence $d(x_1, x_2) = r_1$; the l.c.a.\ of $x_1$ and $x_3$ is the node $n_3$ at altitude $r_3$, hence $d(x_1, x_3) = r_3$. The graph~(c) with edge weights $u$ leads to the ultrametric (a) via the min-max distance $d_u$ defined in \eqref{eq:minmaxd}. For example, all the paths from $x_1$ to $x_3$ contain an edge of weight $r_3$ which is maximal, and thus $d_u(x_1,x_3) = r_3$.
	}
	\label{fig:all}
\end{figure}
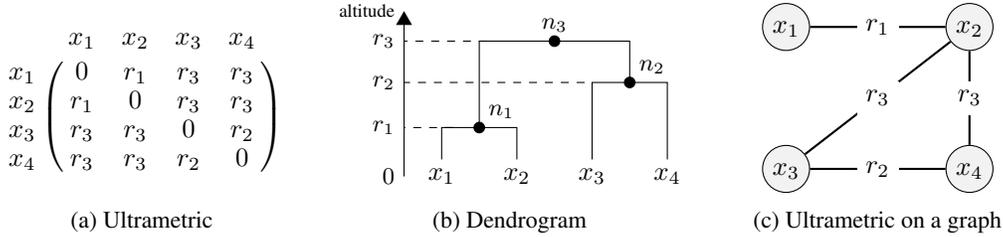

Central to this work is the notion of \emph{ultrametric}, a special kind of metric that is equivalent to hierarchical clustering \cite{CarlssonJMLR2010}. Formally, an ultrametric $d\colon V\times V\to\RR_+$ is a metric on a space $V$ in which the triangle inequality is strengthen by the \emph{ultrametric inequality}, defined as
\begin{equation}\label{eq:ultrametric_inequality}
(\forall (x,y,z)\in V^3)\qquad d(x,y) \leq \max\set{d(x,z), d(z,y)}.
\end{equation}
The notion of ultrametric can also be defined on a connected (non-complete) graph $\G = (\V,\E)$ with non-negative edge weights $\w\in\weightfuns$, where $\weightfuns$ denotes the space of functions from $E$ to $\RR_+$. In this case, the distance is only available between the pairs of vertices in $E$, and the ultrametric constraint must be defined over the set of cycles $\Cycles$ of $\G$ as follows:
\begin{equation}
\label{eq:umconstraint}
\text{$u\in\weightfuns$ is an ultrametric\footnotemark\ on $\G$} \qquad\Leftrightarrow\qquad (\forall C \in \Cycles, \forall e \in C)\quad u(e) \leq \max_{e' \in C \backslash \set{e}} u(e').
\end{equation}
Note that an ultrametric $u$ on $\G$ can be extended to all the pairs of vertices in $V$ through the min-max distance on $u$, which is defined as
\begin{equation}\label{eq:minmaxd}
(\forall (x,y)\in V^2)\qquad  d_u(x,y) = \min_{P\in \Paths_{xy}} \max_{e' \in P} \; u(e'),
\end{equation}
where $\Paths_{xy}$ denotes the set of all paths between the vertices $x$ and $y$ of $\G$. This observation allows us to compactly represent ultrametrics as weight functions $u\in\weightfuns$ on sparse graphs $\G$, instead of more costly pairwise distances. 
Figure \ref{fig:all} shows an example of ultrametric and its possible representations\footnotetext{Some authors use different names, such as \emph{ultrametric contour map}\Cite{ArbelaezPAMI2011} or \emph{saliency map}\Cite{NajmanPAMI1996}.}.

\paragraph{Notation.}
The  dendrogram associated to an ultrametric $u$ on $\G$ is denoted by $T_u$\Cite{CarlssonJMLR2010}.
It is a rooted tree whose leaves are the elements of $V$. Each tree node $n\in T_u$ is the set composed by all the leaves of the sub-tree rooted in $n$.
The altitude of a node $n$, denoted by $\alt_u(n)$, is the maximal distance between any two elements of $n$: \ie{} $\alt_u(n)=\max\set{u(e_{xy}) \mid x, y \in n \textrm{ and } e_{xy}\in E}$.
The size of a node $n$, denoted by $|n|$, is the number of leaves contained in $n$. 
For any two leaves $x$ and $y$, the lowest common ancestor of $x$ and $y$, denoted $\lca_u(x,y)$, is the smallest node of $T_u$ containing both $x$ and $y$.

\subsection{Optimization framework} 
Our goal is to find the ultrametric that "best" represents the given edge-weighted graph. 
We propose to formulate this task as a constrained optimization problem involving an appropriate cost function $J\colon \weightfuns \to \mathbb{R}$ defined on the (continuous) space of distances $\weightfuns$, leading to
\begin{equation}\label{eq:ultrametric_fitting}
\operatorname*{minimize}_{u\in\weightfuns} \quad J(u;\w) \quad\text{s.t.}\quad 
\text{$u$ is an ultrametric on $\G$}.
\end{equation}
The ultrametricity constraint is highly nonconvex and cannot be efficiently tackled with standard optimization algorithms. We circumvent this issue by replacing the constraint with an operation injected directly into the cost function. The idea is that the ultrametricity constraint can be enforced implicitly through the operation that computes the subdominant ultrametric, defined as the largest ultrametric below the given dissimilarity function. One way to compute the subdominant ultrametric is through the min-max operator $\minmax_{\G}\colon \weightfuns \to \weightfuns$ defined by
\begin{equation}\label{eq:minmax}
(\forall \tilde{w}\in\weightfuns, \forall e_{xy}\in \E)\qquad \minmax_{\G}(\tilde{w})(e_{xy}) =
\min_{P\in \Paths_{xy}} \max_{e' \in P} \; \tilde{w}(e'),
\end{equation}
where $\Paths_{xy}$ is defined as in \eqref{eq:minmaxd}. Then, Problem \eqref{eq:ultrametric_fitting} can be rewritten as
\begin{equation}\label{eq:ultrametric_fitting_unconstrained}
\operatorname*{minimize}_{\tilde{w}\in\weightfuns} \; J\big( \minmax_{\G}(\tilde{w});\, w \big).
\end{equation}
Since the min-max operator is sub-differentiable (see \eqref{eq:subdominant_gradient} in Section \ref{sec:algo}), the above problem can be optimized by gradient descent, as long as $J$ is sub-differentiable. This allows us to devise Algorithm~\ref{algo:gradient_descent}.

Note that the mix-max operator already proved useful in image segmentation to define a structured loss function for end-to-end supervised learning \cite{MALIS2009,DEEPMALIS}. The goal was however to find a flat segmentation rather than a hierarchical one. To the best of our knowledge, we are the first to use the mix-max operator within an optimization framework for ultrametric fitting.

\begin{algorithm}[t]
\caption{Solution to the ultrametric fitting problem defined in \eqref{eq:ultrametric_fitting}.}
\label{algo:gradient_descent}
\begin{algorithmic}[1]
	\Require{Graph $\G=(\V,\E)$ with edge weights $\w$} 
	\State $\tilde{w}^{[0]} \leftarrow \w$
	\For{$t=0,1,\dots\Big.$}
	\State $g^{[t]} \leftarrow $ gradient of $J\big( \minmax_{\G}(\cdot);\, \w \big)$ evaluated at $\tilde{w}^{[t]}$
	\State $\tilde{w}^{[t+1]} \leftarrow $ update of $\tilde{w}^{[t]}$ using $g^{[t]}\Big.$
	\EndFor
	\State \textbf{return} $\minmax_{\G}(\tilde{w}^{[\infty]})\big.$
\end{algorithmic}
\end{algorithm}

\subsection{Closest ultrametric}
A natural goal for ultrametric fitting is to find the closest ultrametric to the given dissimilarity graph. This task fits nicely into Problem \eqref{eq:ultrametric_fitting} by setting the cost function to the sum of squared errors between the sought ultrametric and the edge weights of the given graph, namely
\begin{equation}\label{eq:closest}
J_{\rm closest}(u;\w) = \sum_{e \in E} \Big(u(e) - w(e)\Big)^2.
\end{equation}
Although the exact minimization of this cost function is a NP-hard problem\Cite{kvrivanek1988complexity}, the proposed optimization framework allows us to compute an approximate solution. Figure~\ref{fig:examples} shows the ultrametric computed by Algorithm \ref{algo:gradient_descent} with $J_{\rm closest}$ for an illustrative example of hierarchical clustering.

A common issue with the closest ultrametric is that small clusters might branch very high in the dendrogram. This is also true for average linkage and other agglomerative methods. Such kind of ultrametrics are undesirable, because they lead to partitions containing very small clusters at large scales, as clearly shown in Figures~\ref{fig:examples:average}-\ref{fig:examples:closest}. We now present two approaches to tackle this issue.

\subsection{Cluster-size regularization}
To fight against the presence of small clusters at large scales, we need to introduce a mechanism that pushes down the altitude of nodes where such incorrect merging occurs. 
This can be easily translated in our framework, as the altitude of a node corresponds to the ultrametric distance between its children. Specifically, we penalize the ultrametric distance proportionally to some non-negative coefficients that depend on the corresponding nodes in the dendrogram, yielding
\begin{equation}\label{eq:area}
J_{\rm size}(u) = \sum_{e_{xy} \in E}  \frac{u(e_{xy})}{\gamma_u(\lca_u(x,y))}.
\end{equation}
Here above, the $\gamma$ coefficients play an essential role: they must be small for the nodes that need to be pushed down, and large otherwise. We thus rank the nodes by the size of their smallest child, that is
\begin{equation}
(\forall n\in T_u)\qquad \gamma_u(n) = \min\set{|c|, c\in\children_u(n)},
\end{equation}
where $\children_u(n)$ denotes the children of a node $n$ in the dendrogram $T_u$ associated to the ultrametric $u$ on $\G$. Figure~\ref{fig:examples:size} shows the ultrametric computed by Algorithm \ref{algo:gradient_descent} with $J_{\rm closest} + J_{\rm size}$. The positive effect of this regularization can be appreciated by observing that small clusters are no longer branched very high in the dendrogram.

\subsection{Triplet regularization}
Triplet constraints \cite{vikram2016, Chatziafratis2018} provide an alternative way to penalize small clusters at large scales. Like in semi-supervised classification, we assume that the labels $\mathcal{L}_v$ of some data points $v \in V$ are known, and we build a set of triplets according to the classes they belong to: 
\begin{equation}
\mathcal{T} = \big\{({\rm ref}, {\rm pos}, {\rm neg})\in V^3 \;|\; \mathcal{L}_{\rm ref} = \mathcal{L}_{\rm pos} \quad\textrm{and}\quad \mathcal{L}_{\rm ref} \neq \mathcal{L}_{\rm neg} \big\}.
\end{equation}
These triplets provide valuable information on how to build the ultrametric. Intuitively, we need a mechanism that reduces the ultrametric distance within the classes, while increasing the ultrametric distance between different classes. This can be readily expressed in our framework with a regularization acting on the altitude of nodes containing the triplets, leading to
\begin{equation}
\label{eq:triplet}
J_{\rm triplet}(u) = \sum_{({\rm ref}, {\rm pos}, {\rm neg})\in\mathcal{T}} \max\{0, \alpha + d_u({\rm ref},{\rm pos}) - d_u({\rm ref},{\rm neg})\}.
\end{equation}
Here above, the constant $\alpha>0$ represents the minimum prescribed distance between different classes. Figure~\ref{fig:examples:triplet} shows the ultrametric computed by Algorithm \ref{algo:gradient_descent} with $J_{\rm closest} + J_{\rm triplet}$.

\newcommand{\image}[3][]{%
\begin{subfigure}[b]{0.15\textwidth}%
\includegraphics[width=\textwidth]{#2.png}%
#1%
#3%
\end{subfigure}
}

\begin{figure}
\centering
\image{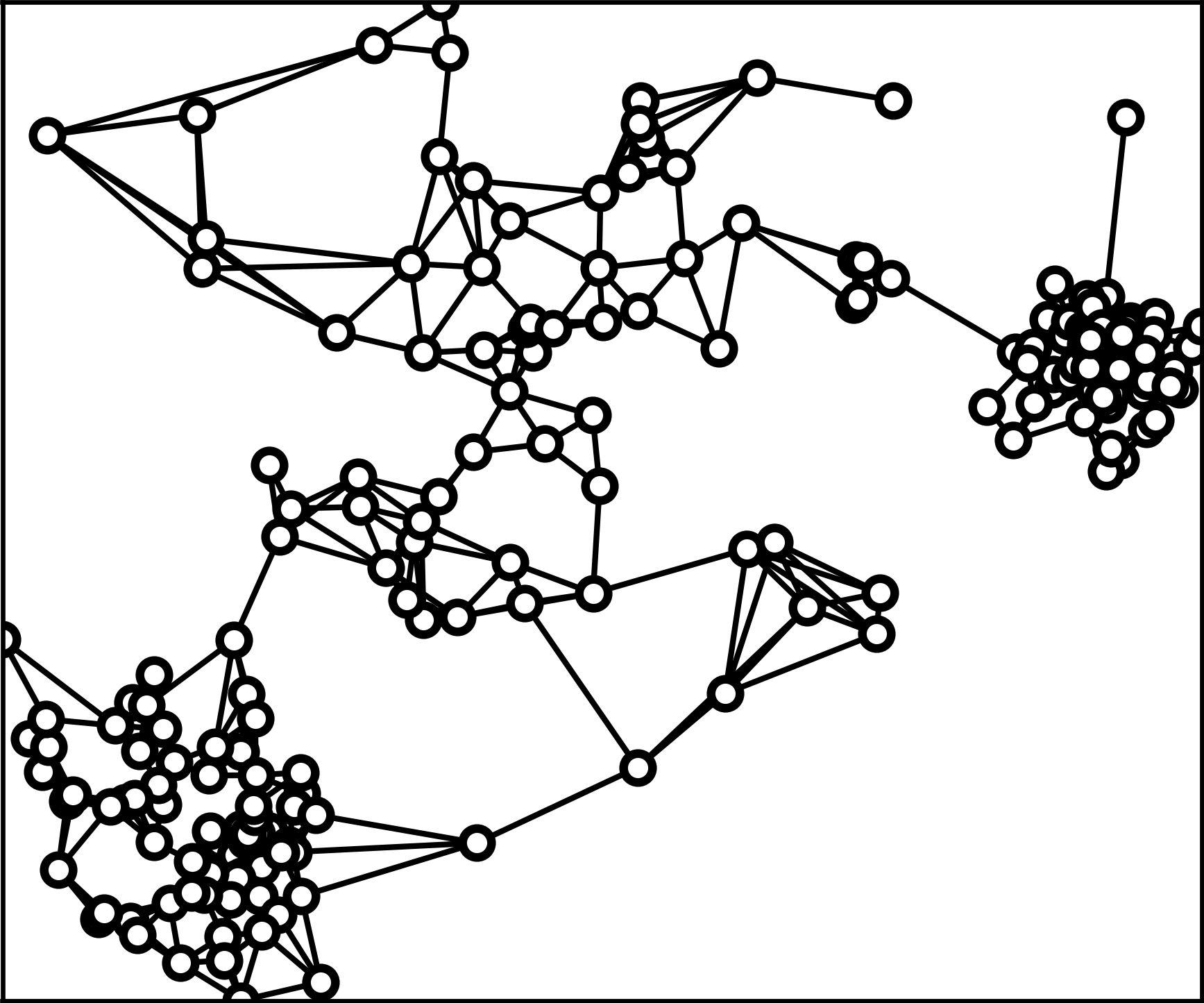}{}
\hfill
\image{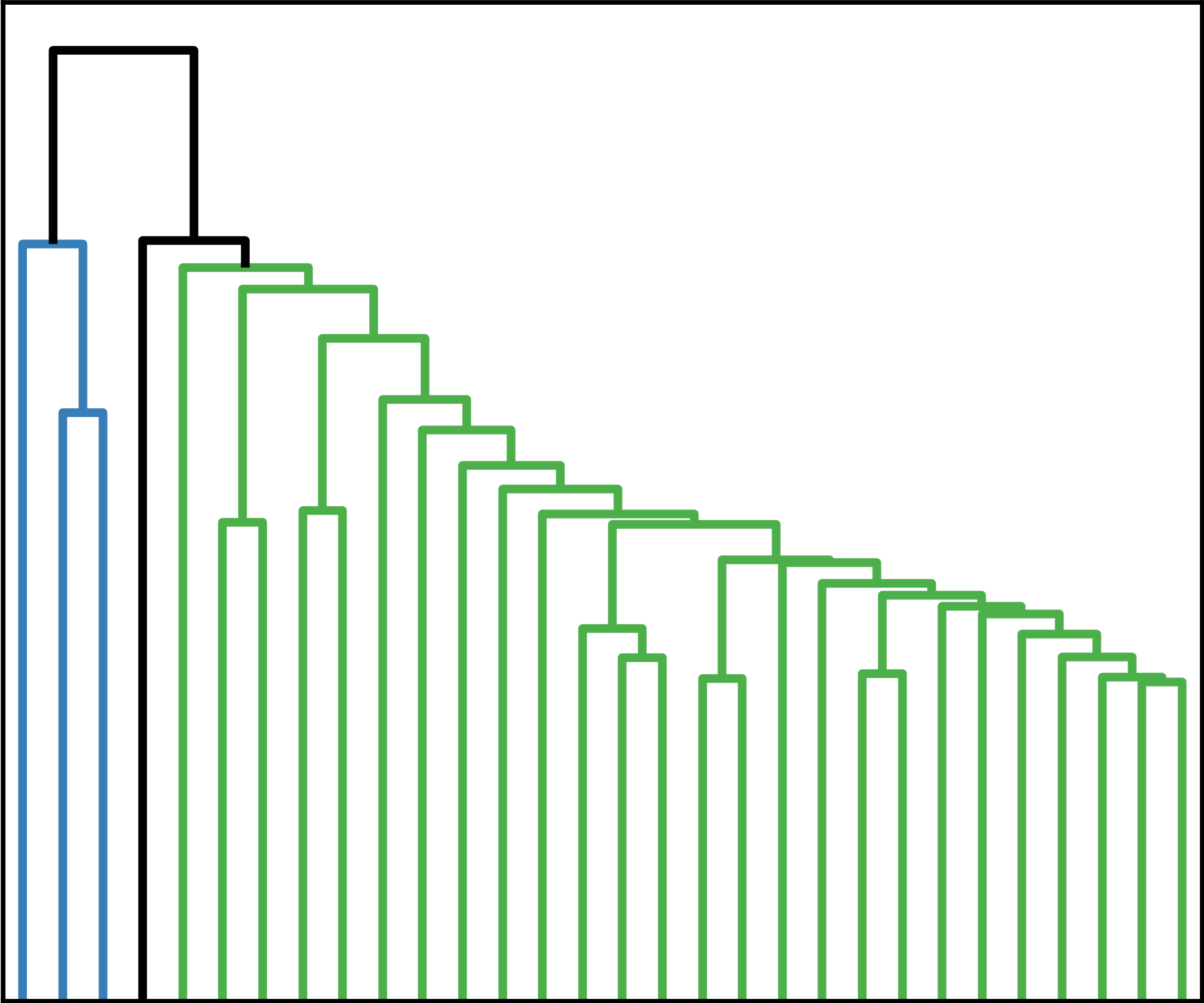}{}
\hfill
\image{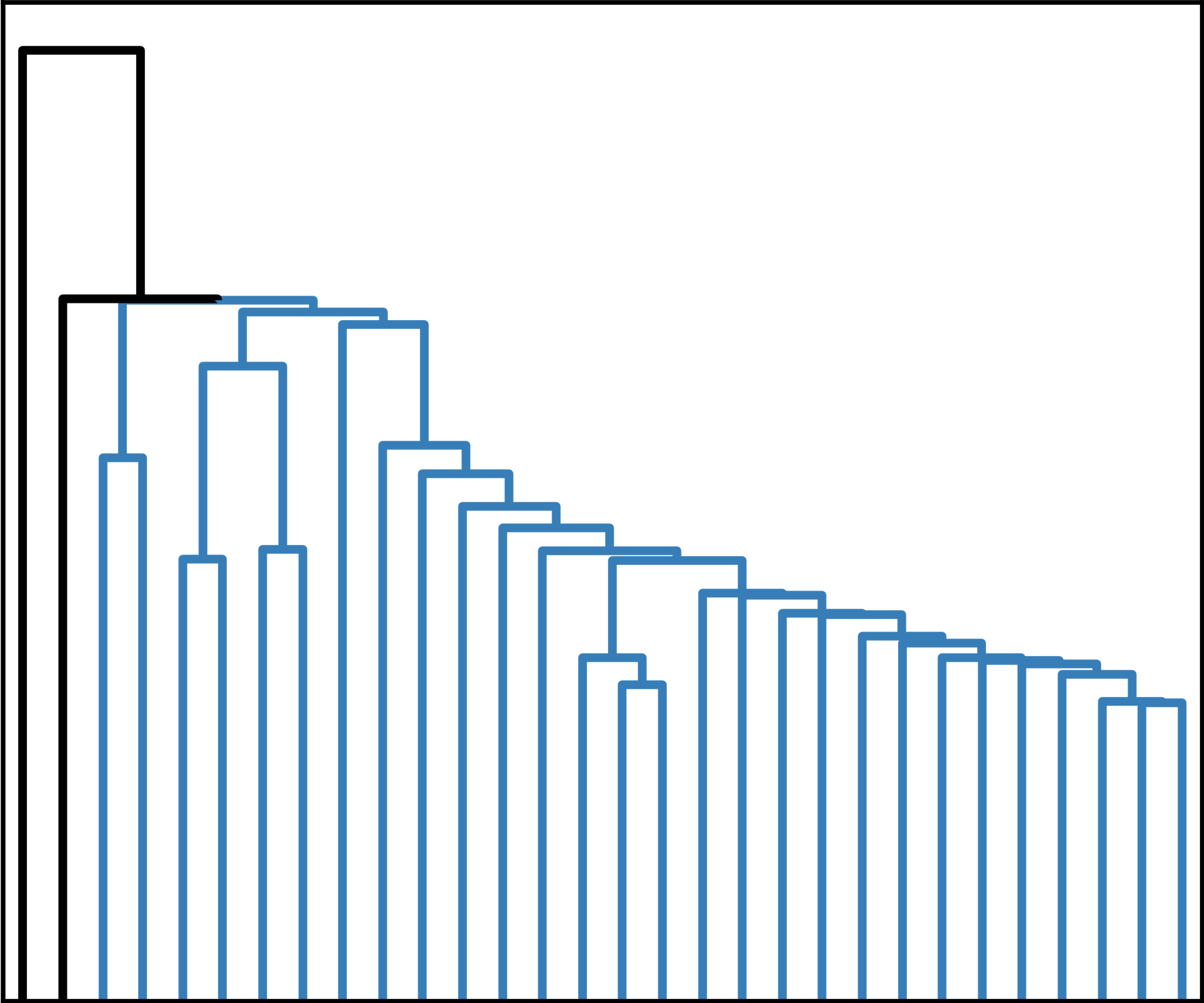}{}
\hfill
\image{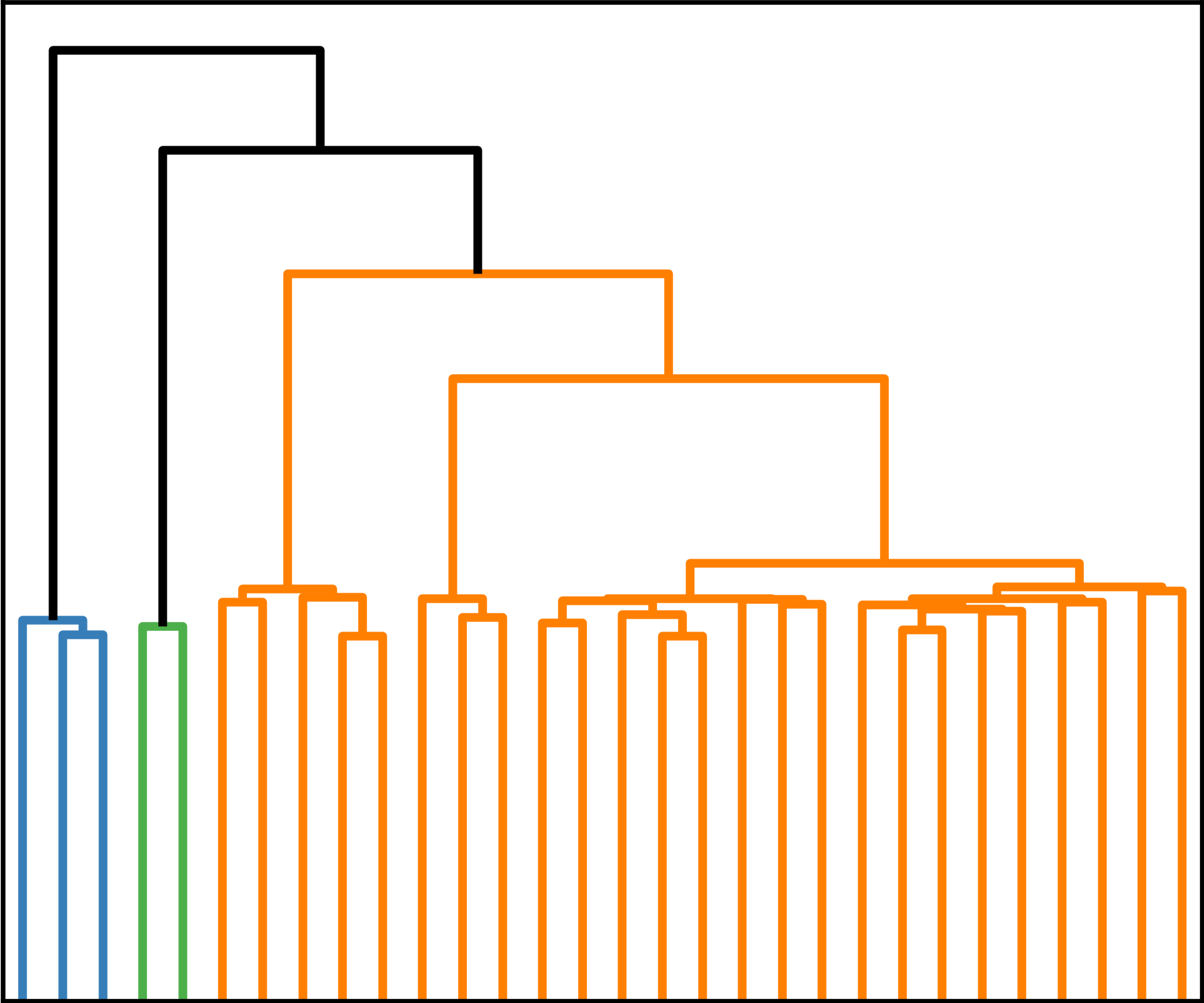}{}
\hfill
\image{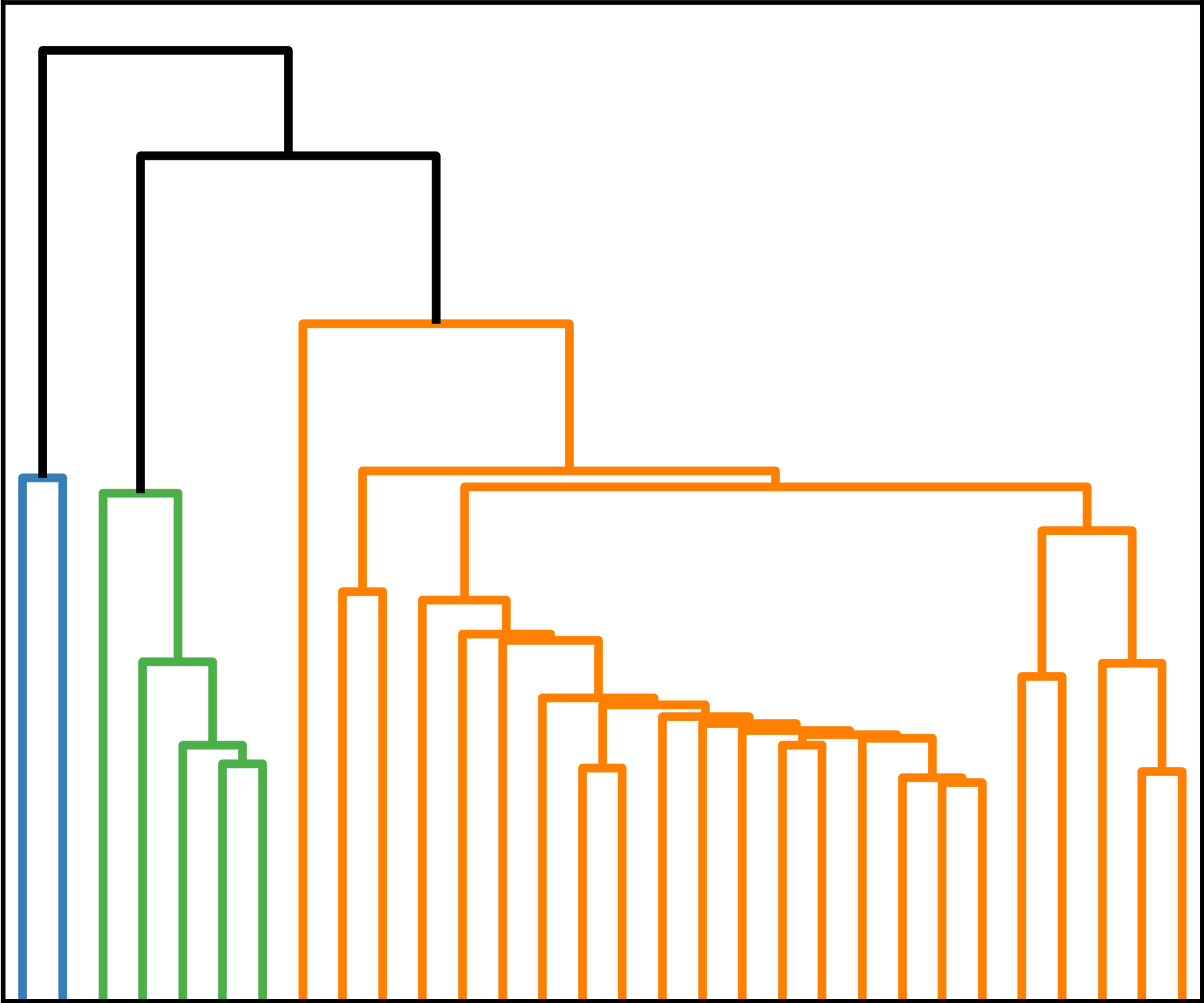}{}
\hfill
\image{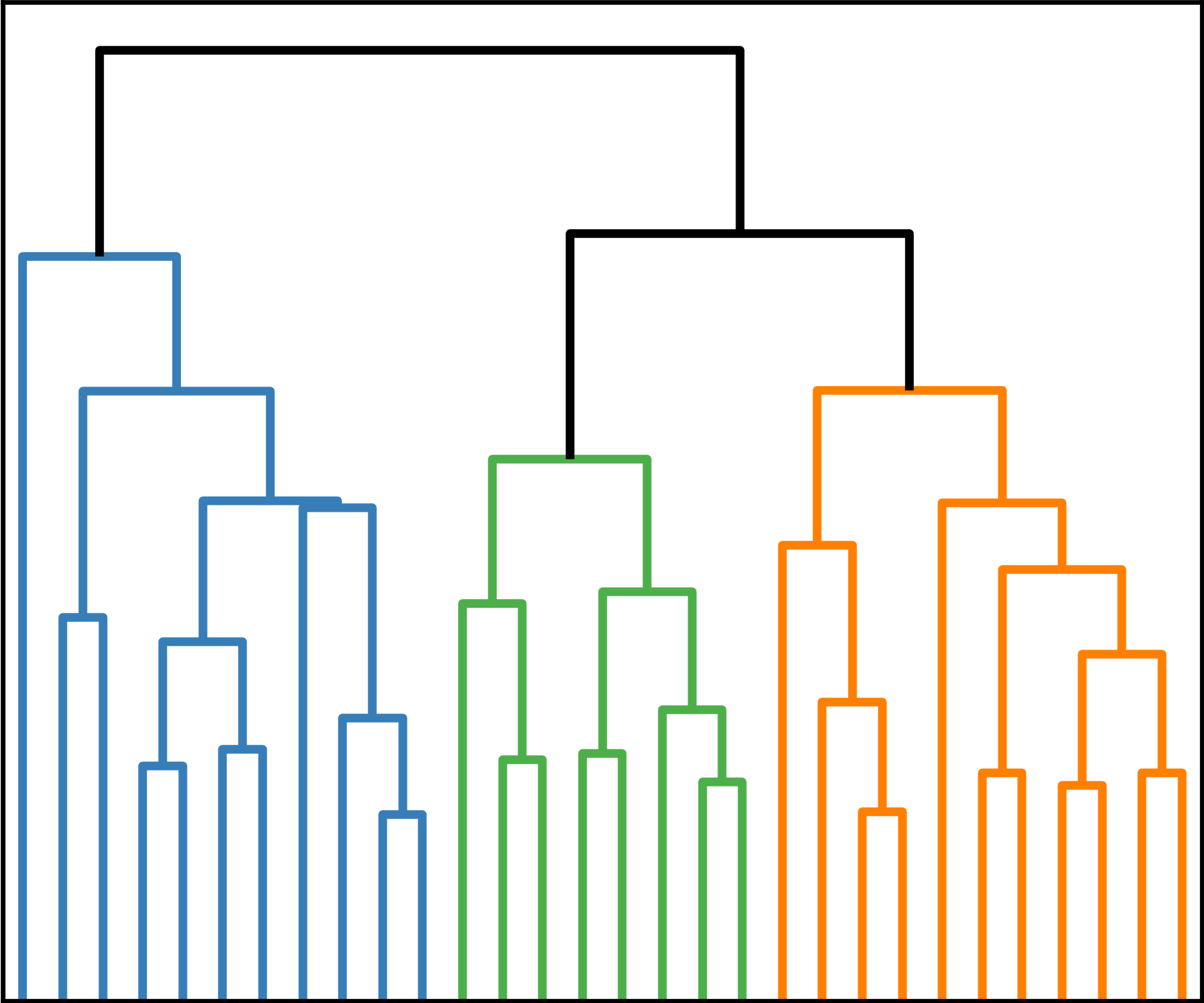}{}

\image[\caption{\footnotesize Graph/Labels}]{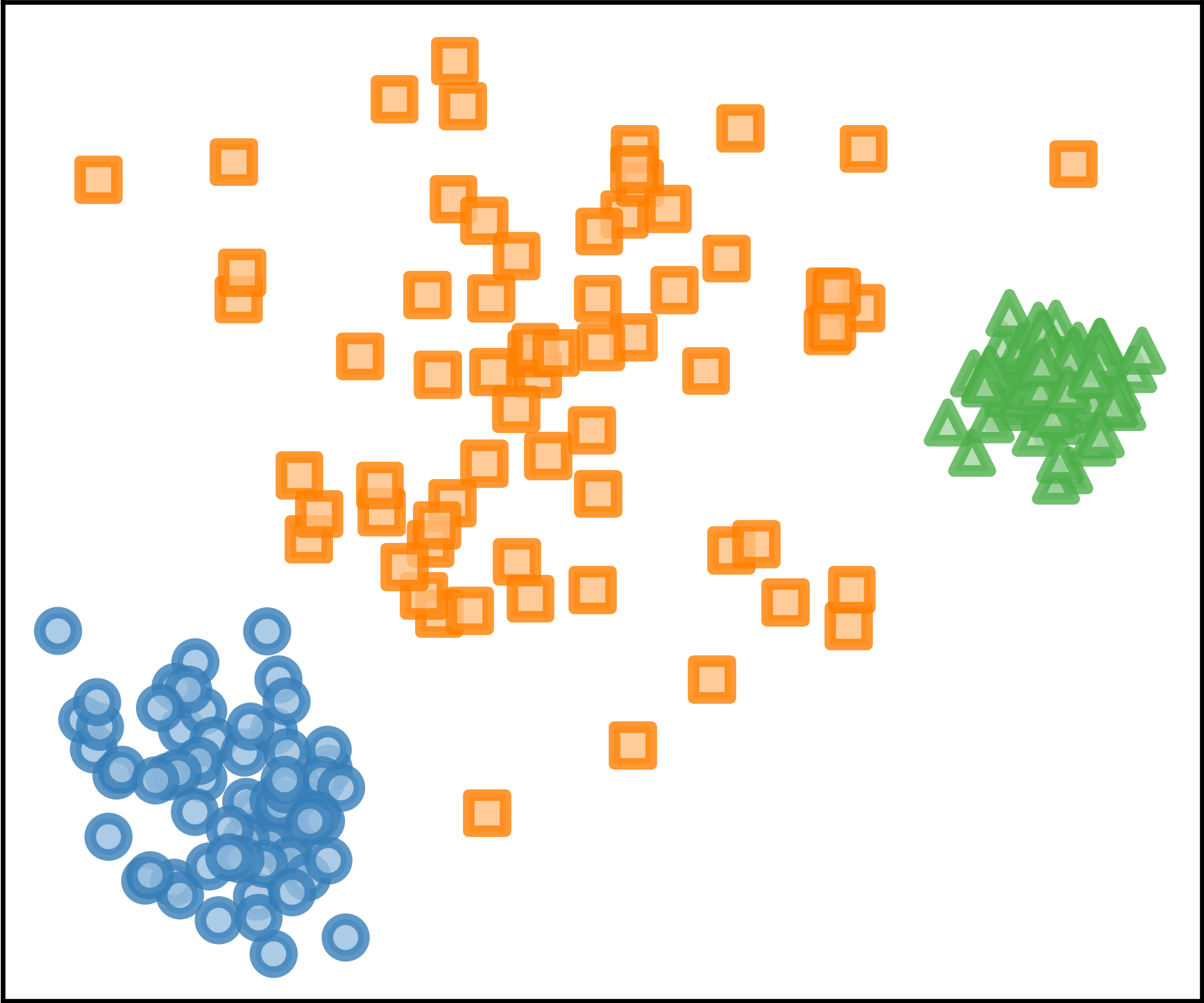}{}
\hfill
\image[\caption{\footnotesize Average link}]{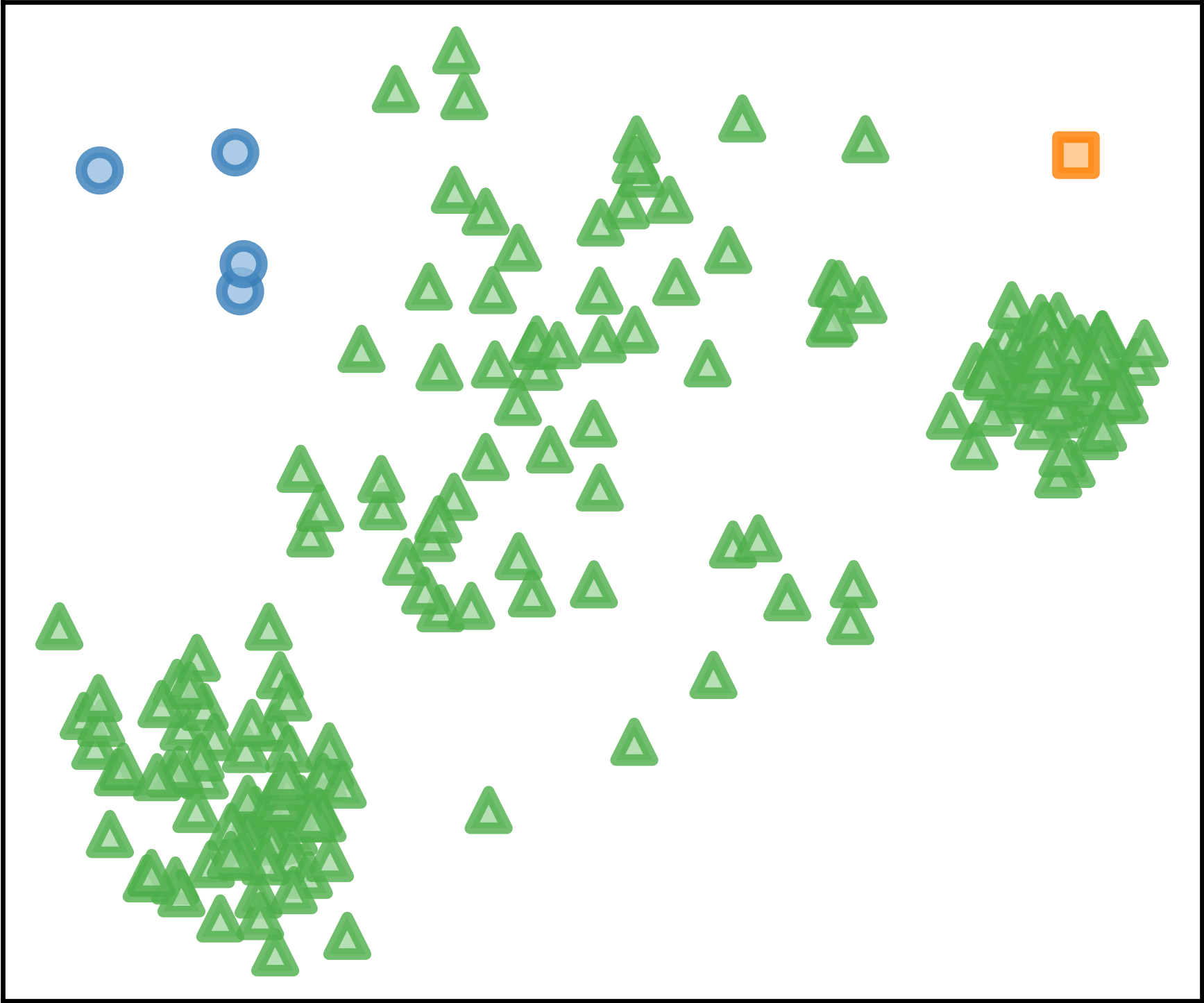}{\label{fig:examples:average}}
\hfill
\image[\caption{\footnotesize Closest}]{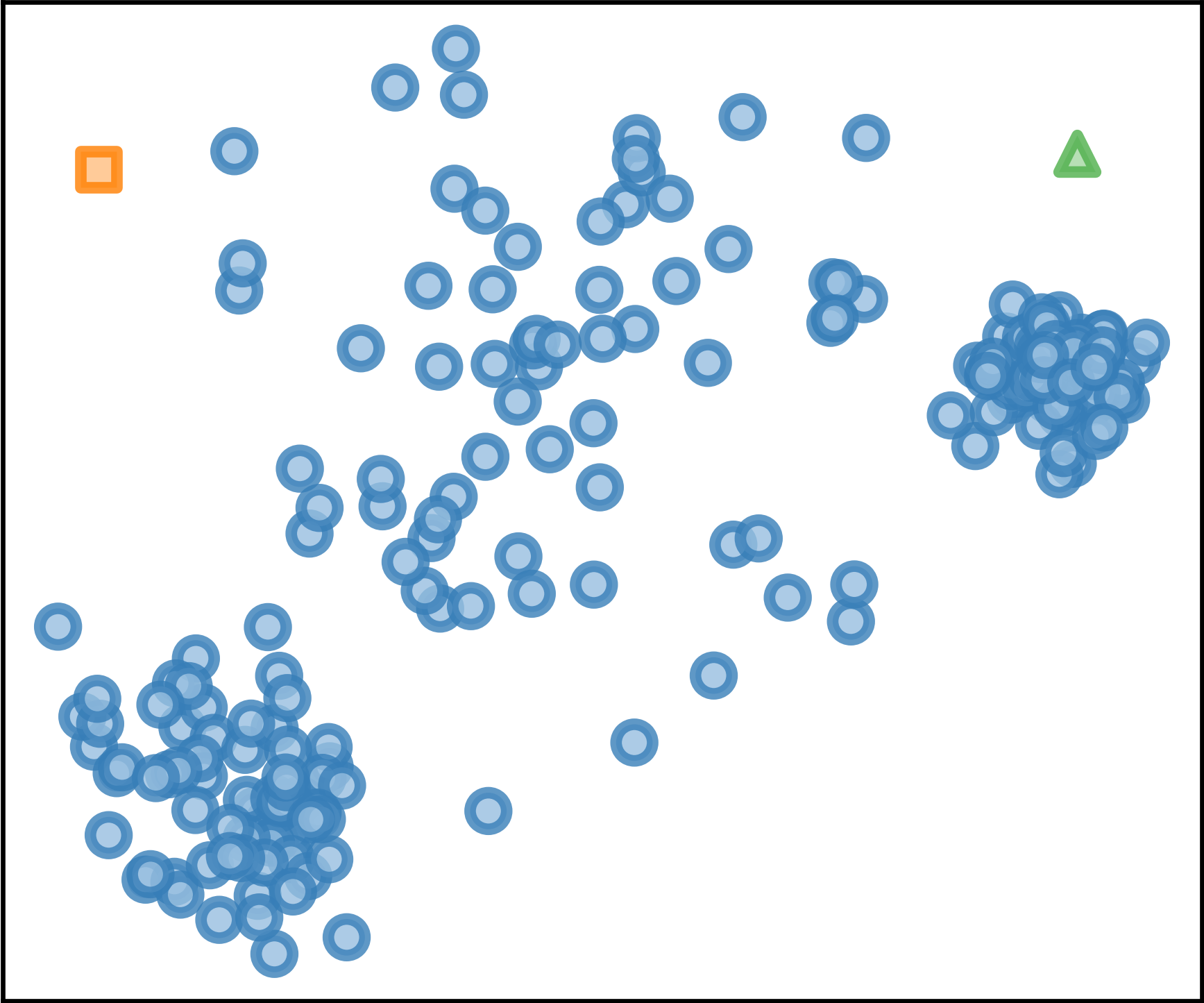}{\label{fig:examples:closest}}
\hfill
\image[\caption{\footnotesize Closest+Size}]{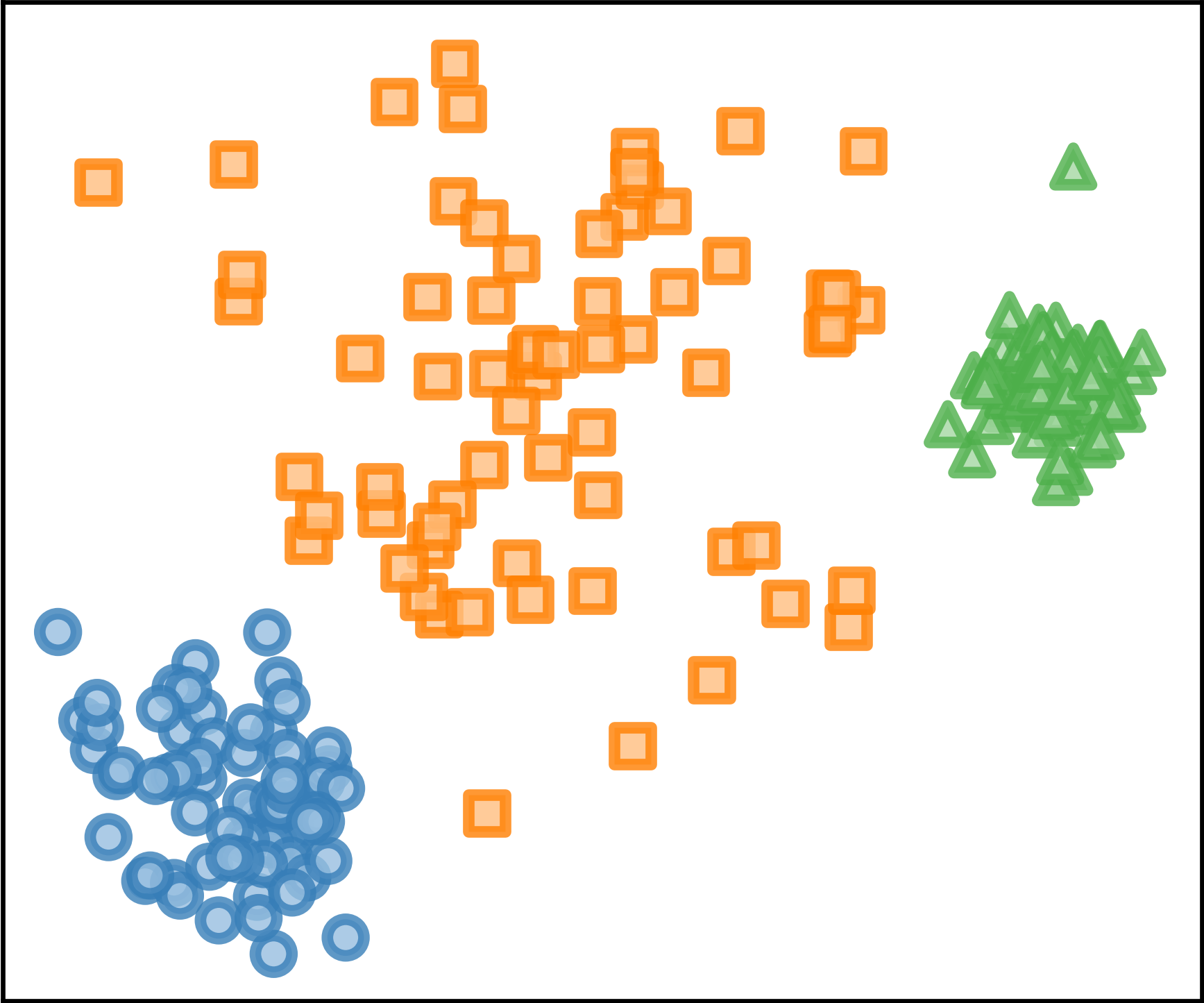}{\label{fig:examples:size}}
\hfill
\image[\caption{\scriptsize Closest+Triplet}]{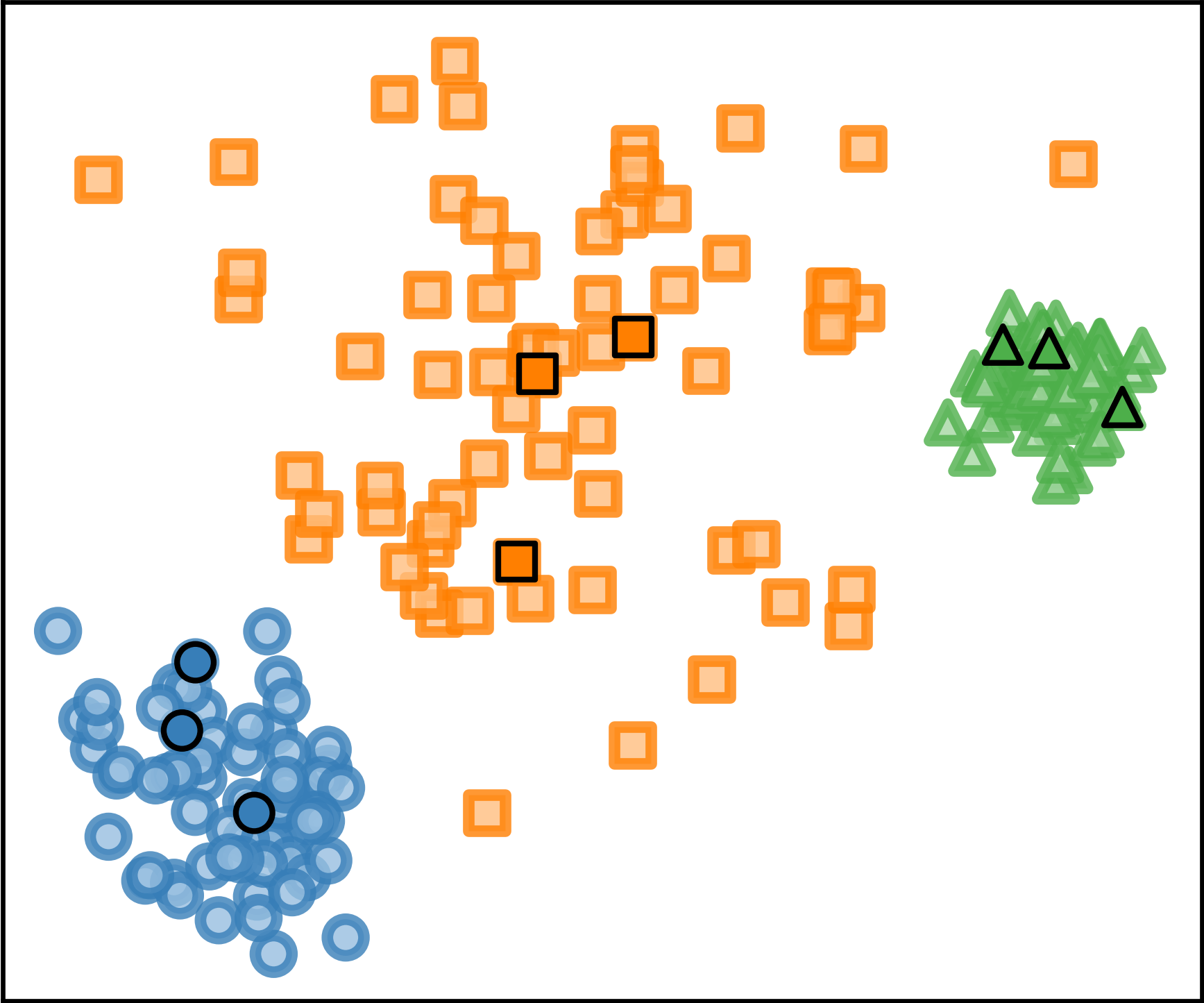}{\label{fig:examples:triplet}}
\hfill
\image[\caption{\footnotesize Dasgupta}]{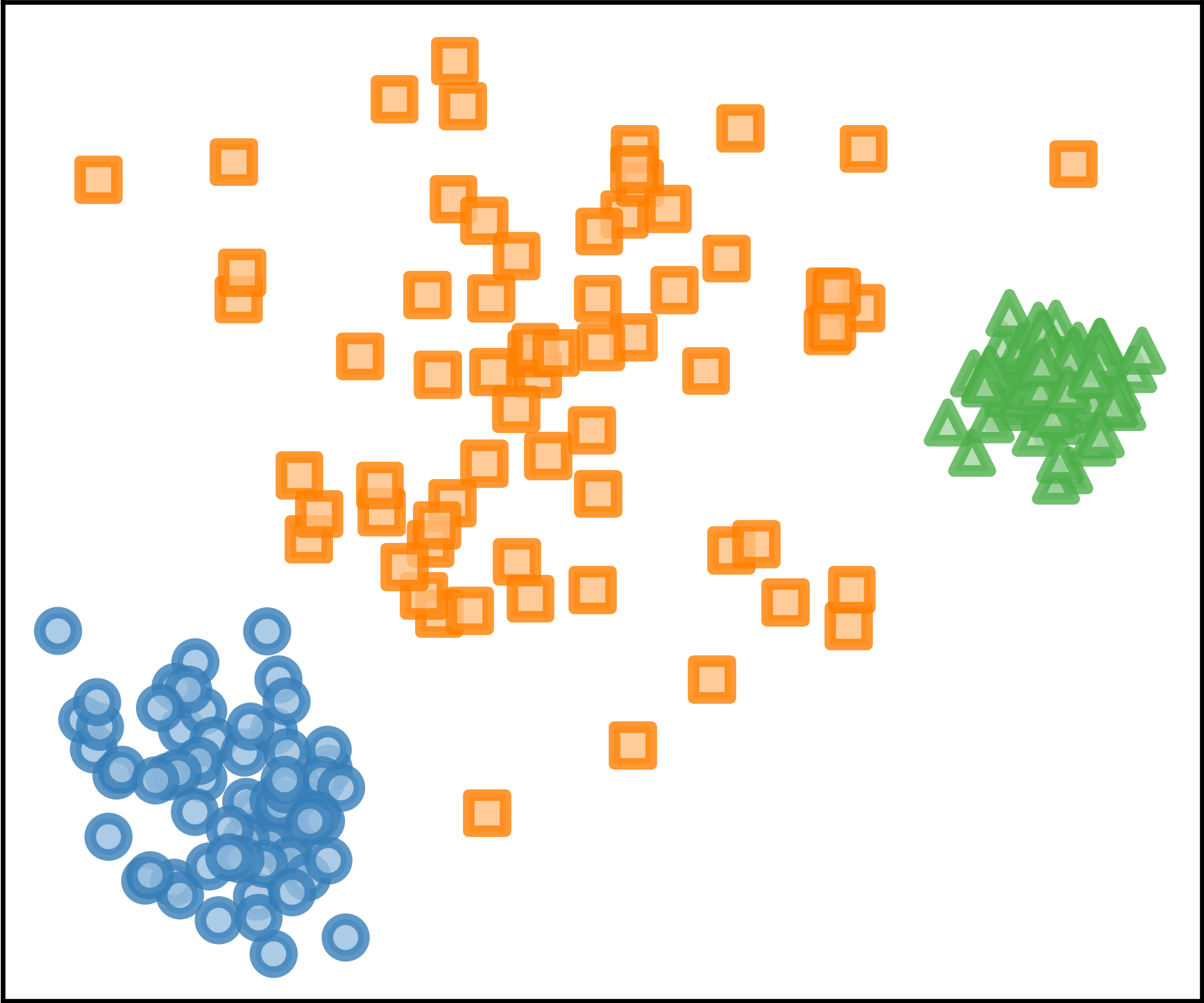}{\label{fig:examples:dasgupta}}
\caption{Illustrative examples of hierarchical clustering. \emph{Top row}: Ultrametrics fitted to the input graph; only the top-30 non-leaf nodes are shown in the dendrograms (all the others are contracted into leaves). \emph{Bottom row}: Assignments obtained by thresholding the ultrametrics at three clusters. The detrimental effect of having "small clusters at large scales" can be observed in (b) and (c).}
\label{fig:examples}
\end{figure}

\subsection{Dasgupta cost function}
Dasgupta cost function\Cite{Dasgupta2016} has recently gained traction in the seek of a theoretically grounded framework for hierarchical clustering\Cite{RoyNIPS2016,CohenAddadNIPS2017,RoyJMLR2017,CohenAddad2018,Chatziafratis2018}. 
However its minimization is known to be a NP-hard problem\Cite{Dasgupta2016}.
The intuition behind this function is that large clusters should be associated to large dissimilarities.
The idea it then to minimize, for each edge $e\in E$, the size of the dendrogram node  associated to $e$ divided by the weight of $e$, yielding
\begin{equation}
\label{eq:dasgupta}
J_{\rm Dasgupta}(u; w) = \sum_{e_{xy} \in E} \frac{|\textrm{lca}_u(x,y)|}{w(e_{x,y})}.
\end{equation}
However, we cannot directly use\Eq{eq:dasgupta} in our optimization framework, as the derivative of $|\lca_u(x,y)|$ with respect to the underlying ultrametric $u$ is equal to 0 almost everywhere. 
To solve this issue, we propose a \emph{soft-cardinal} measure of a node that is  differentiable w.r.t.\ the associated ultrametric $u$. 
Let $n$ be a node of the dendrogram $T_u$, and let $\{x\}\subseteq n$ be a leaf of the sub-tree rooted in $n$. We observe that the cardinal of $n$ is equal to the number of vertices $y\in\V$ such that the ultrametric distance $d_u(x,y)$ between $x$ and $y$ is strictly lower than the altitude of $n$, namely
\begin{equation}
\label{eq:nodecardinal}
|n| = \sum_{y \in \V}H(\alt_u(n) - d_u(x,y)),
\end{equation}
where $H$ is the Heaviside function. By replacing $H$ with a continuous approximation, such as a sigmoid function, we provide a soft-cardinal measure of a node of $T_u$ that is differentiable with respect to the ultrametric $u$.
Figure~\ref{fig:examples:dasgupta} shows the ultrametric computed by Algorithm \ref{algo:gradient_descent} with $J_{\rm Dasgupta}$.

Note that a differentiable cost function inspired by Dasgupta cost function was proposed in~\cite{monath2017}. This function replaces the node size by a parametric probability measure which is optimized over a fixed tree. This is fundamentally different from our approach, where the proposed measure is a continuous relaxation of the node size, and it is directly optimized over the ultrametric distance.

\begin{figure}[t]
    \centering
    \begin{subfigure}[b]{0.4\textwidth}
    \centering
    \includegraphics[width=0.85\textwidth]{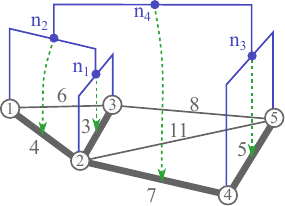}
    \caption{Single linkage clustering}
    \label{fig:singleLinkageClustering}
    \end{subfigure} \hfill
    \begin{subfigure}[b]{0.5\textwidth}
    \begin{equation*}
    \bordermatrix{%
         & e_{13} & e_{12} & e_{23}& e_{35} & e_{25} & e_{24} & e_{45} \cr
  e_{13} &  0 & 0 & 0 & 0 & 0 & 0 & 0 \cr
  e_{12} & 1 & 1 & 0 & 0 & 0 & 0 & 0 \cr
  e_{23} & 0 & 0 & 1 & 0 & 0 & 0 & 0 \cr
  e_{35} & 0 & 0 & 0 & 0 & 0 & 0 & 0 \cr
  e_{25} & 0 & 0 & 0 & 0 & 0 & 0 & 0 \cr
  e_{24} & 0 & 0 & 0 & 1 & 1 & 1 & 0 \cr
  e_{45} & 0 & 0 & 0 & 0 & 0 & 0 & 1
}
    \end{equation*}
    \caption{Jacobian}\label{fig:minmaxjacobian}
    \end{subfigure}
    \caption{Each node of the single linkage clustering (in blue) of the graph (in grey) is canonically associated (green dashed arrows) to an edge of a minimum spanning tree of the graph (thick edges): this edge is the pass edge between the leaves of the two children of the node.  Edges are numbered from 1 to $M$ (number of edges). The $i$-th column of the Jacobian matrix of the $\minmax$ operator is equal to the indicator vector denoting the pass edge holding the maximal value of the min-max path between the two extremities of the  $i$-th edge. 
    The pass edge can be found efficiently in the single linkage clustering using the l.c.a.\ operation and the canonical link between the nodes and the m.s.t.\ edges. 
    For example, the l.c.a.\ of the vertices $3$ and $5$ linked by the 4-th edge $e_{35}$ is the node $n_4$, which is canonically associated to the 6-th edge $e_{24}$ ($\mstEdge{n_4}=e_{24}$): we thus have $J_{6,4} = 1$. }\label{fig:minmaxjacobianall}
\end{figure}

%% file: algo.tex
In this section, we present a general approach to efficiently compute the various terms appearing in the cost functions introduced earlier. 
All the proposed algorithms rely on some properties of the \emph{single linkage (agglomerative) clustering}, which is a dual representation of the subdominant ultrametric. 
We perform a detailed analysis of the algorithm used to compute the subdominant ultrametric. The other algorithms can be found in the supplemental material.

Single-linkage clustering can be computed similarly to a minimum spanning tree (m.s.t.)\ with Kruskal's algorithm, by sequentially processing the edges of the graph in non decreasing order, and merging the clusters located at the extremities of the edge, when a m.s.t.\ edge is found. 
One consequence of this approach is that each node $n$ of the dendrogram representing the single linkage clustering is canonically associated to an edge of the m.s.t. (see \Fig{fig:singleLinkageClustering}), which is denoted by $\mstEdge{n}$.

In the following, we assume that we are working with a sparse graph $\G=(\V,\E)$, where $\mathcal{O}(|E|)=|V|$. The number of vertices (resp. edges) of $\G$ is denoted by $N$ (resp. $M$). 
For the ease of writing, we denote the edge weights as vectors of $\RR^M$.
The dendrogram corresponding to the single-linkage clustering of the graph $\G$ with edge weights $\tilde{w}\in\weightfuns$ is denoted by $\SL(\tilde{w})$.

\subsection{Subdominant ultrametric}
To obtain an efficient and automatically differentiable algorithm for computing the subdominant ultrametric, we observe that the min-max distance between any two vertices $x,y \in V$ is given by the weight of the \emph{pass edge} between $x$ and $y$. This is the edge holding the maximal value of the min-max path from $x$ to $y$, and an arbitrary choice is made if several pass edges exist. Moreover, the pass edge between $x$ and $y$ corresponds to the l.c.a.\ of $x$ and $y$ in the single linkage clustering of $(\G,\tilde{w})$ (see \Fig{fig:singleLinkageClustering}). Equation \eqref{eq:minmax} can be thus rewritten as 
\begin{equation}
    \label{eq:minmax2}
(\forall \tilde{w}\in\weightfuns, \forall e_{xy}\in \E)\qquad \minmax_{\G}(\tilde{w})(e_{xy}) = \tilde{w}(e_{xy}^{\rm mst})
\quad{\rm with}\quad e_{xy}^{\rm mst} = \mstEdge{\lca_{\SL(\tilde{w})}(x,y)}.
\end{equation}
The single-linkage clustering can be computed in time $\mathcal{O}(N\log N)$  with a variant of Kruskal's minimum spanning tree algorithm\Cite{Gower1969, NajmanISMM2013}.
Then, a fast algorithm allows us to compute the l.c.a.\ of two nodes in constant time $\mathcal{O}(1)$, thanks to a linear time $\mathcal{O}(N)$ preprocessing of the tree\Cite{Bender2000}. The subdominant ultrametric can thus be computed in time $\mathcal{O}(N\log N)$ with Algorithm~\ref{algo:subdominant}. Moreover, the dendrogram associated to the ultrametric returned by Algorithm~\ref{algo:subdominant} is the tree computed on line~2. 

Note that Algorithm~\ref{algo:subdominant} can be interpreted as a special max pooling applied to the input tensor $w$, and can be thus automatically differentiated.
Indeed, a sub-gradient of the min-max operator $\minmax$ at a given edge $e_{xy}$ is equal to 1 on the pass edge between $x$ and $y$ and 0 elsewhere.
Then, the Jacobian of the min-max operator $\minmax$ can be written as the matrix composed of the indicator column vectors giving the position of the pass edge associated to the extremities of each edge in $E$:
\begin{equation}
\label{eq:subdominant_gradient}
\frac{\partial \minmax(\tilde{w})}{\partial \tilde{w}} = \left[\mathbbm{1}_{\minmax^*_\G(\tilde{w}_1)}, \ldots, \mathbbm{1}_{\minmax^*_\G(\tilde{w}_{M})} \right],
\end{equation}
where $\minmax^*_\G(\tilde{w}_i)$ denotes the index of the pass edge between the two extremities of the $i$-th edge, and $\mathbbm{1}_{j}$ is the column vector of $\RR^{M}$ equals to 1 in position $j$, and 0 elsewhere (see \Fig{fig:minmaxjacobian}).

\subsection{Regularization terms}
The cluster-size regularization defined in \eqref{eq:area} can be implemented through the same strategy used in Algorithm~\ref{algo:subdominant}, based on the single-linkage clustering and the fast l.c.a.\ algorithm, leading to a time complexity in $\mathcal{O}(N\log N)$. See supplemental material. 

Furthermore, thanks to Equation \eqref{eq:minmax2}, the triplet regularization defined in \eqref{eq:triplet} can be written as
\begin{equation}
\label{eq:triplet2}
J_{\rm triplet}\big({u}\big) = \sum_{({\rm ref}, {\rm pos}, {\rm neg})\in\mathcal{T}} \max\{0, \alpha + {u}\big(\mstEdge{\lca_{u}({\rm ref},{\rm pos})}\big) - {u}\big(\mstEdge{\lca_{u}({\rm ref},{\rm neg})}\big)\}.
\end{equation}
This can be implemented with a time complexity in $\mathcal{O}(|\mathcal{T}| + N\log N)$. See supplemental material.

\begin{algorithm}[t]
\caption{Subdominant ultrametric operator defined in \eqref{eq:minmax} with \eqref{eq:minmax2}.}
\label{algo:subdominant}
\begin{algorithmic}[1]
	\Require{Graph $\G=(\V,\E)$ with edge weights $\w$} 
	\State $u(e_{xy}) \leftarrow 0$ \textbf{for each} $e_{xy}\in\E$ \Comment{$\mathcal{O}(N)$}
	\State $tree \leftarrow \textrm{single-linkage}(\G,\w)$  \Comment{$\mathcal{O}(N \log N)$ with\cite{Gower1969,NajmanISMM2013}}
	\State preprocess l.c.a. on $tree$  \Comment{$\mathcal{O}(N)$ with\Cite{Bender2000}}
	\ForEach{edge $e_{xy}\in\E$} \Comment{$\mathcal{O}(N)$} 
	    \State $pass\_node \leftarrow \lca_{tree}(x,y)$ \Comment{$\mathcal{O}(1)$ with\Cite{Bender2000}} 
	    \State $pass\_edge \leftarrow \mstEdge{pass\_node}$ \Comment{$\mathcal{O}(1)$ see  \Fig{fig:singleLinkageClustering}}
	    \State $u(e_{xy}) \leftarrow w(pass\_edge)$ \Comment{$\mathcal{O}(1)$}
    \EndFor
	\State \textbf{return} $u$
\end{algorithmic}
\end{algorithm}

\subsection{Dasgupta cost function}
The soft cardinal of a node $n$ of the tree $T_u$ as defined in \eqref{eq:nodecardinal} raises two issues: the arbitrary choice of the reference vertex $x$, and the quadratic time complexity $\Theta(N^2)$ of a naive implementation. One way to get rid of the arbitrary choice of $x$ is to use the two extremities of the edge $\mstEdge{n}$ canonically associated to $n$.
To efficiently compute \eqref{eq:nodecardinal}, we can notice that, if $c_1$ and $c_2$ are the children of $n$, then the pass edge between any element $x$ of $c_1$ and any element $y$ of $c_2$ is equal to edge  $\mstEdge{n}$ associated to $n$. 
This allows us to define $\card(n)$ as the relaxation of $|n|$ in \eqref{eq:nodecardinal}, by replacing  the Heaviside function $H$ with the sigmoid function $\ell$, leading to
\begin{align}
\label{eq:nodecardinaloptim}
\textrm{card}(n)& = \frac{1}{2}\sum_{x\in\sigma(n)}\sum_{y\in V}\ell(\textrm{alt}_u(n) - d_u(x,y)) \nonumber\\
& = \frac{1}{2}\sum_{x\in\sigma(n)} \Big( \ell(\textrm{alt}_u(n) - d_u(x,x)) + \sum_{y\in V\backslash{\{x\}}}\ell(\textrm{alt}_u(n) - d_u(x,y))\Big) \nonumber\\
& = \frac{1}{2}\sum_{x\in\sigma(n)} \Big( \ell(\textrm{alt}_u(n)) + \sum_{y\in \ancestors(x)}\sum_{z\in c_{\hat{x}}(y)}\ell(\textrm{alt}_u(n) - d_u(x,z)) \Big) \nonumber\\
& = \frac{1}{2}\sum_{x\in\sigma(n)} \Big( \ell(\textrm{alt}_u(n)) + \sum_{y\in \ancestors(x)}\sum_{z\in c_{\hat{x}}(y)}\ell(\textrm{alt}_u(n) - \textrm{alt}_u(y)) \Big) \nonumber\\
& = \frac{1}{2}\sum_{x\in\sigma(n)} \Big( \ell(\textrm{alt}_u(n)) + \sum_{y\in \ancestors(x)}|c_{\hat{x}}(y)|\ell(\textrm{alt}_u(n) - \textrm{alt}_u(y)) \Big) \nonumber\\
& = \frac{1}{2}\sum_{x \in \mstEdge{n}}\Big(\ell\big({u}(\mstEdge{n})\big) + \sum_{y \in \ancestors(x)}|\otherChild_{\hat{x}}(y)|\ell\big({u}(\mstEdge{n}) - {u}(\mstEdge{y})\big) \Big),
\end{align}
where  $\ancestors(x)$ is the set of ancestors of $x$,  and $\otherChild_{\hat{x}}(y)$ is the child of $y$ that does not contain $x$.
The time complexity to evaluate \eqref{eq:nodecardinaloptim} is dominated by the sum over the ancestors of $n$ which, in the worst case, is in the order of $\mathcal{O}(N)$, leading also to worst case time complexity of $\mathcal{O}(N^2)$. In practice, dendrograms are usually well balanced, and thus the number of ancestors of a node is in the order of $\mathcal{O}(\log N)$, yielding an empirical complexity in $\mathcal{O}(N\log N)$.

%% file: closest.tex
\begin{figure}
\centering
\fbox{\includegraphics[width=0.2\textwidth]{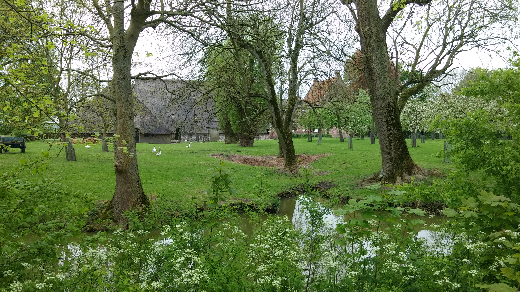}}
\fbox{\includegraphics[width=0.2\textwidth]{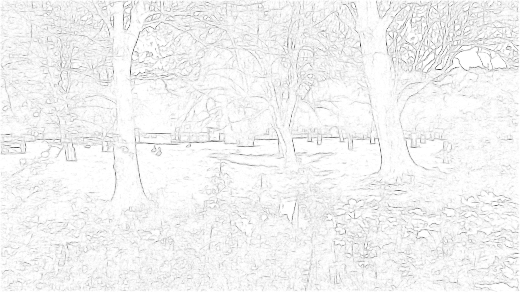}}
\fbox{\includegraphics[width=0.2\textwidth]{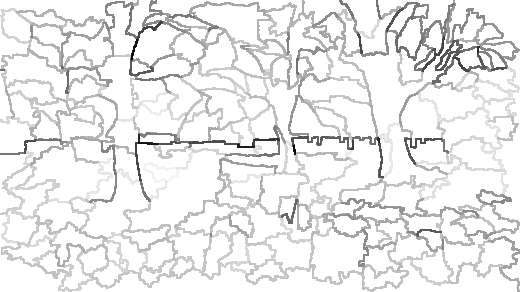}}
\fbox{\includegraphics[width=0.2\textwidth]{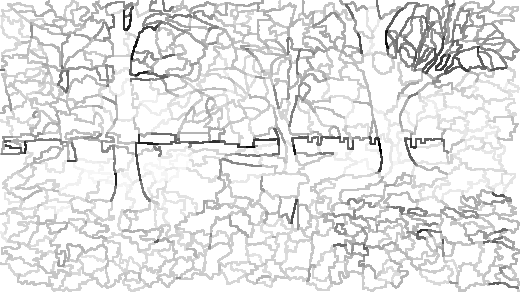}}
\fbox{\includegraphics[width=0.2\textwidth]{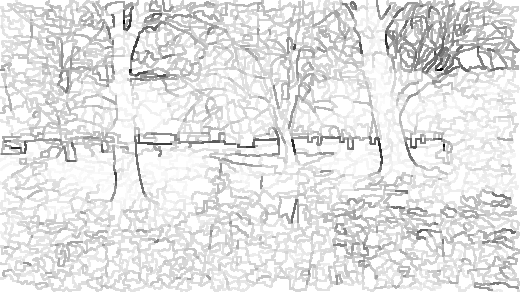}}
\caption{Test image, its gradient, and superpixel contour weights with 525, 1526, and 4434 edges.}
\label{fig:demoRAG}
\end{figure}

\begin{figure}
\centering
\begin{subfigure}[b]{0.32\textwidth}
\includegraphics[width=1\textwidth]{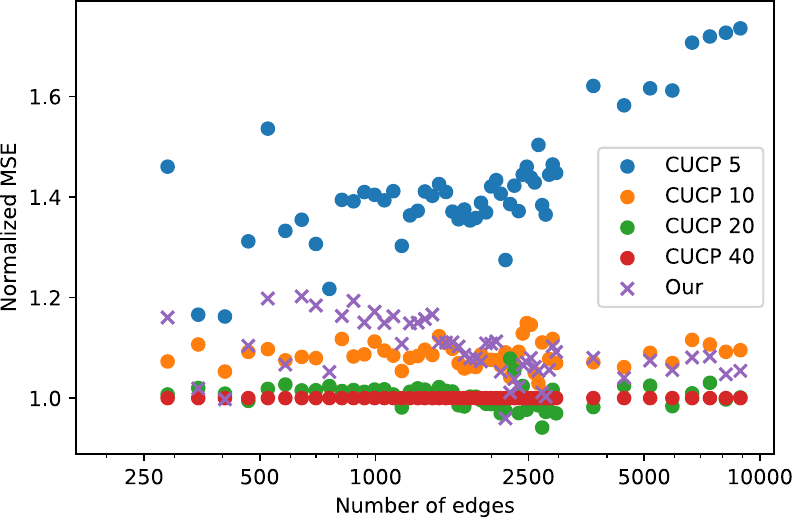}
\caption{Mean square error}
\label{fig:mseCUCP}
\end{subfigure}
\hfill
\begin{subfigure}[b]{0.32\textwidth}
\includegraphics[width=1\textwidth]{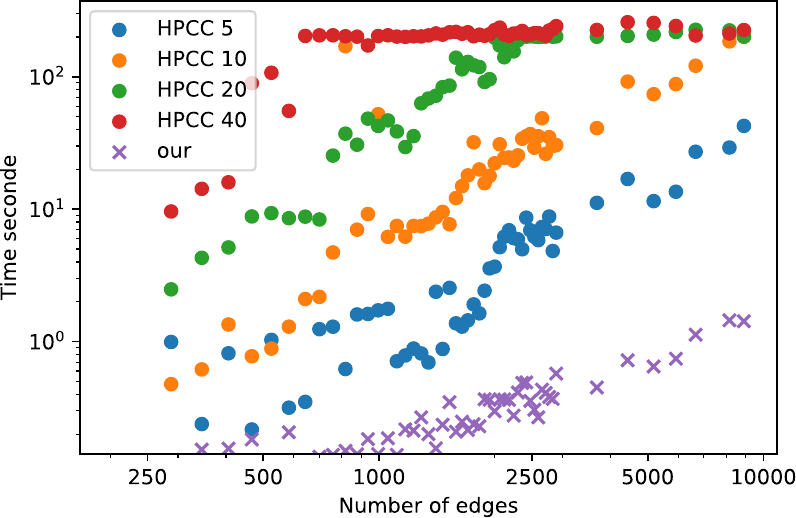}
\caption{Computation time}
\label{fig:runtimeCUCP}
\end{subfigure}
\hfill
\begin{subfigure}[b]{0.32\textwidth}
\includegraphics[width=1\textwidth]{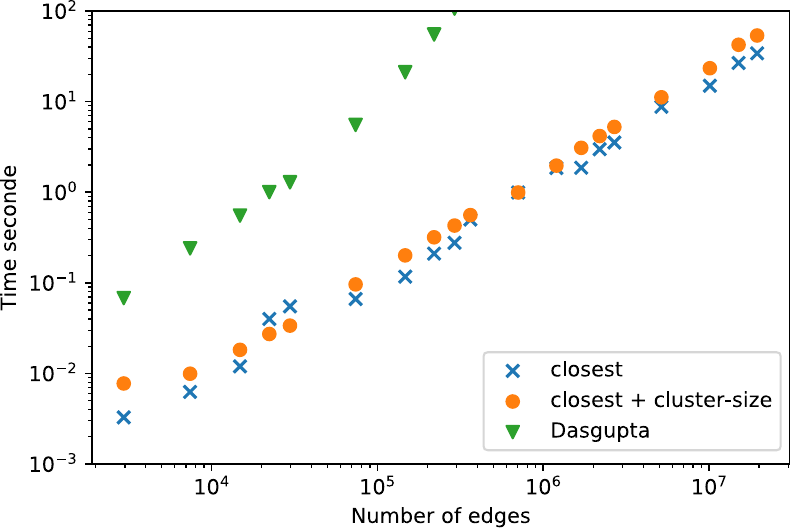}
\caption{Computation time per iteration}
\label{fig:timePerIteration}
\end{subfigure}
\caption{Validation and computation time. Figureas (a) and (b): comparison between the CUCP algorithm\Cite{YarkonyNIPS2015} and the proposed gradient descent approach. For CUCP we tested different numbers of hierarchy levels (5, 10, 20 40) distributed evenly over the range of the input dissimilarity function. Figure (a) shows the final mean square error  (normalized against CUCP 40) w.r.t.\ the number of edges in the tested graph. Figure (b) shows the run-time compared w.r.t.\ the number of edges in the tested graph (CUCP was capped at 200 seconds per instance). Figure~(c) shows the computation time of the tested cost functions (one iteration of Algorithm~\ref{algo:gradient_descent}) with respect to the number of edges in the graph.}
\label{fig:compCUCP}
\end{figure}

In the following, we evaluate the proposed framework on two different setups. The first one aims to assess our continuous relaxation of the closest ultrametric problem with respect to the (almost) exact solution provided by linear programming on planar graphs \cite{YarkonyNIPS2015}. The second one aims to compare the performance of our cost functions to the classical hierarchical and semi-supervised clustering methods. The implementation of our algorithms, based on Higra~\cite{Perret2019softwarex} and PyTorch~\cite{paszke2017automatic} libraries, is available at \url{https://github.com/PerretB/ultrametric-fitting}.

\paragraph{Framework validation}
\label{sec:validation}
As Problem\Eq{eq:ultrametric_fitting_unconstrained} is non-convex, there is no guarantee that the gradient descent method will find the global optimum. 
To assess the performance of the proposed framework, we use the algorithm proposed in\Cite{YarkonyNIPS2015}, denoted by CUCP (Closest Ultrametric via Cutting Plane), as a baseline for the closest ultrametric problem defined in\Eq{eq:closest}.
Indeed, CUCP can provides an (almost) exact solution to the closest ultrametric problem for planar graphs based on a reformulation as a set of correlation clustering/multi-cuts problems with additional hierarchical constraints.
However, CUCP requires to define a priori the set of levels 
which will compose the final hierarchy. 

We generated a set of superpixels adjacency graphs of increasing scale from a high-resolution image (see \Fig{fig:demoRAG}). The weight of the edge linking two superpixels is defined as the mean gradient value, obtained with\Cite{DollarPAMI2015}, along the frontier between the two superpixels.
The results presented in \Fig{fig:compCUCP} shows that the proposed approach is able to provide solutions close to the optimal ones (\Fig{fig:mseCUCP}) using only a fraction of the time needed by the combinatorial algorithm (\Fig{fig:runtimeCUCP}), and without any assumption on the input graph. The complete experimental setup is in the supplemental material.

The computation time of some combinations of cost terms are presented in \Fig{fig:timePerIteration}. 
Note that, Algorithm~\ref{algo:gradient_descent} usually achieves convergence in about one hundred iterations (see supplemental material).
\emph{Closest} and \emph{Closest+Size} can handle graphs with millions of edges. 
\emph{Dasgupta} relaxation is computationally more demanding, which decreases the limit to a few hundred thousands of edges.

%% file: clustering.tex
\paragraph{Hierarchical clustering}
We evaluate the proposed optimization framework on five datasets downloaded from the LIBSVM webpage,\footnote{\texttt{https://www.csie.ntu.edu.tw/$\sim$cjlin/libsvmtools/datasets/}} whose size ranges from $270$ to $1500$ samples. For each dataset, we build a 5-nearest-neighbor graph, to which we add the edges of a minimum spanning tree to ensure the connectivity. Then, we perform hierarchical clustering on this graph, and we threshold the resulting ultrametric at the prescribed number of clusters. We divide our analysis in two sets of comparisons: hierarchical clustering (unsupervised), and semi-supervised clustering. To be consistent among the two types of comparisons, we use the classification accuracy as a measure of performance. 

\begin{figure}
\centering
\begin{subfigure}[b]{0.49\textwidth}
\includegraphics[width=1\textwidth]{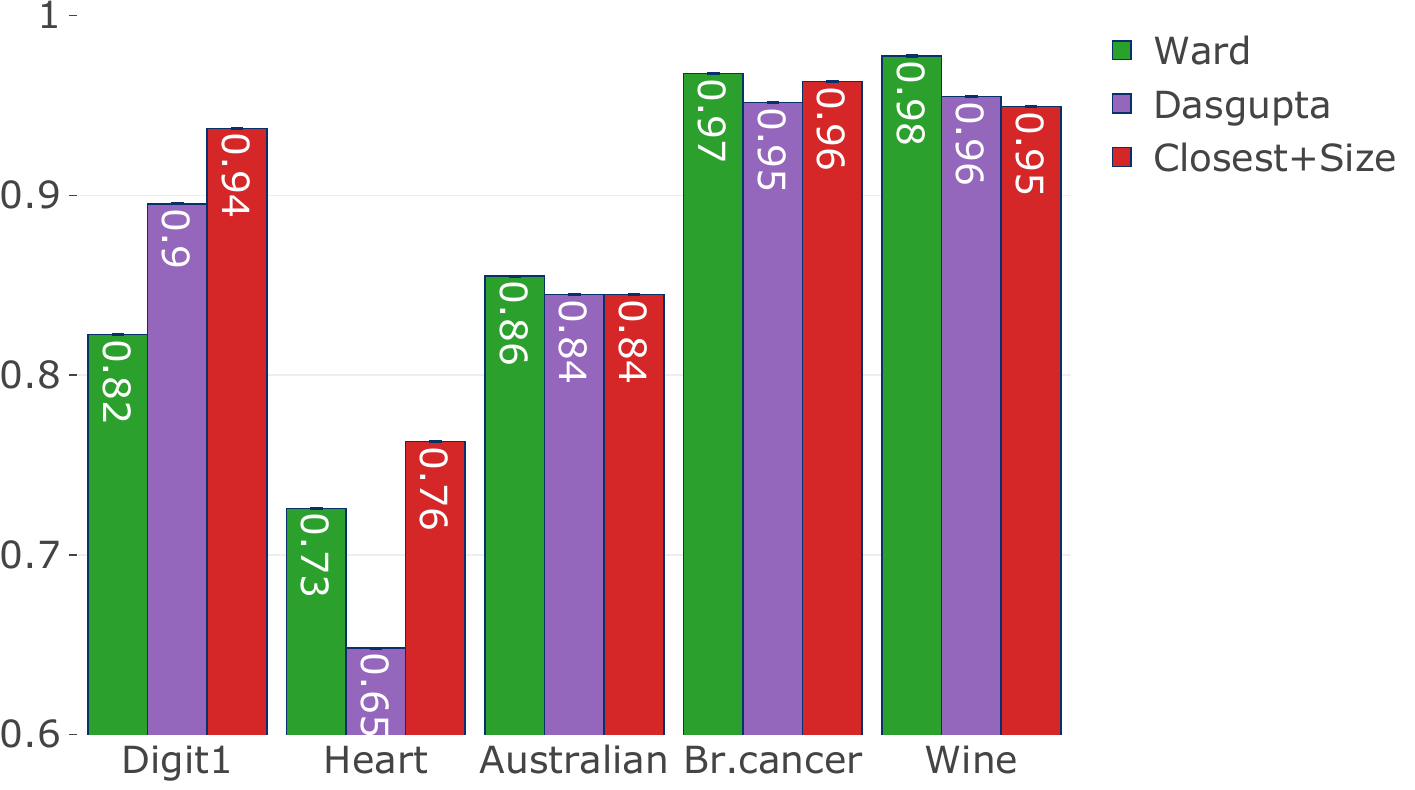}
\caption{Hierarchical clustering (unsupervised)}
\label{fig:performance:clustering}
\end{subfigure}
\begin{subfigure}[b]{0.49\textwidth}
\includegraphics[width=1\textwidth]{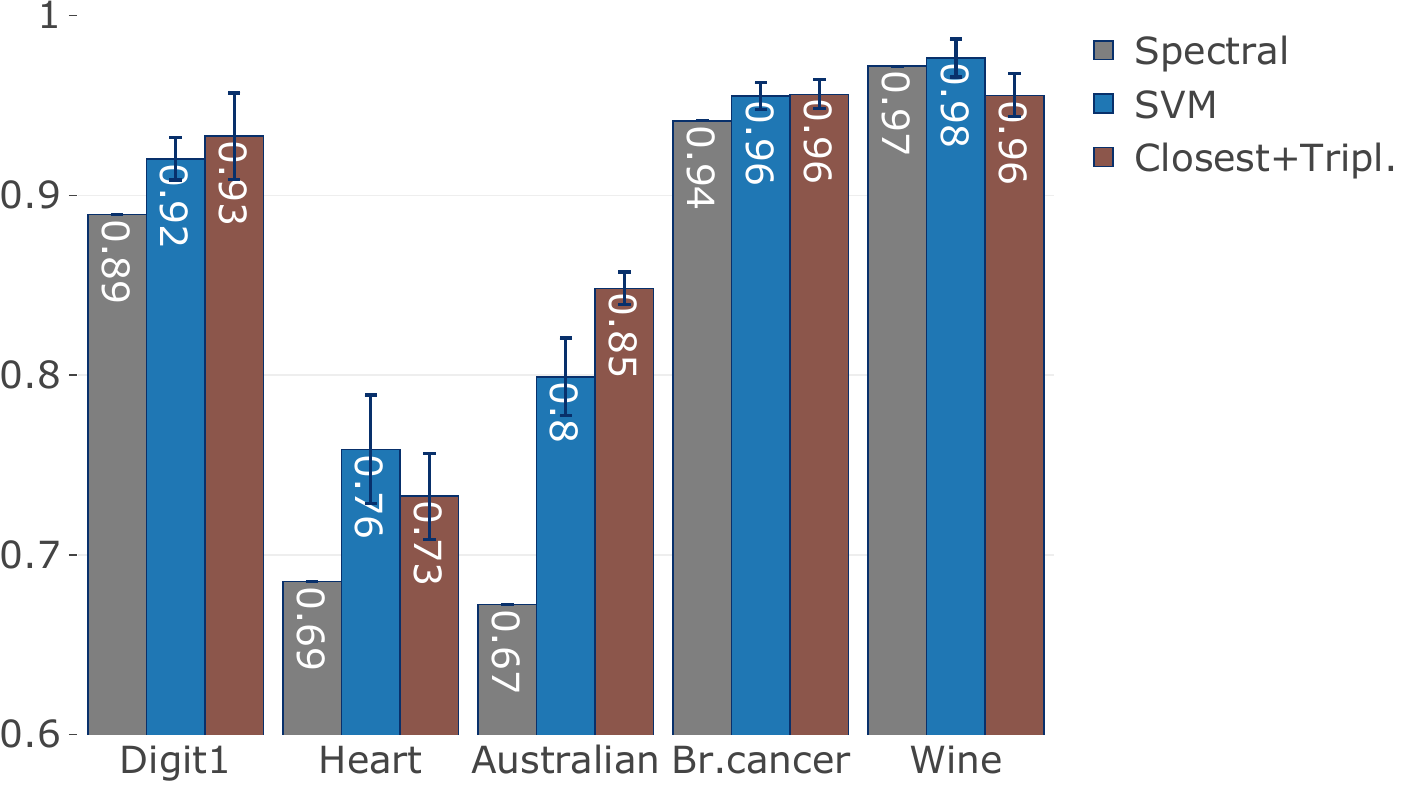}
\caption{Semi-supervised clustering}
\label{fig:performance:semi_supervised}
\end{subfigure}
\caption{Performance on real datasets.}
\label{fig:performance}
\end{figure}

Figure \ref{fig:performance:clustering} compares the performance of three hierarchical clustering methods. The baseline is "\emph{Ward}" agglomerative method, applied to the pairwise distance matrix of each dataset. Average linkage and closest ultrametric are not reported, as their performance is consistently worst. The "\emph{Dasgupta}" method refers to Algorithm \ref{algo:gradient_descent} with $J_{\rm Dasgupta} + \lambda J_{\rm size}$ and $\lambda=1$. The "\emph{Closest+Size}" method refers to  Algorithm \ref{algo:gradient_descent} with the cost function $J_{\rm closest} + \lambda J_{\rm size}$ and $\lambda=10$. In both cases, the regularization is only applied to the top-10 dendogram nodes (see supplemental material). The results show that the proposed approach is competitive with Ward method (one of the best agglomerative heuristics). On the datasets Digit1 and Heart, "\emph{Dasgupta}" performs slightly worse than "\emph{Closest+Size}": this is partly due to the fact that our relaxation of the Dasgupta cost function is sensible to data scaling.

Figure \ref{fig:performance:semi_supervised} compares the performance of two semi-supervised clustering methods, and an additional unsupervised method. The first baseline is "\emph{Spectral}" clustering applied to the Gaussian kernel matrix of each dataset. The second baseline is "\emph{SVM}" classifier trained on the fraction of labeled samples, and tested on the remaining unlabeled samples. Between $10\%$ and $40\%$ of training samples are drawn from each dataset using a 10-fold scheme, and the cross-validated performance is reported in terms of mean and standard deviation. The "\emph{Closest+Triplet}" method refers to Algorithm \ref{algo:gradient_descent} with $J_{\rm closest} + \lambda J_{\rm triplet}$, $\lambda=1$ and $\alpha=10$. The results show that the triplet regularization performs comparably to semi-supervised SVM, which in turn performs better than spectral clustering.

%% file: conclusion.tex
We have presented a general optimization framework for fitting ultrametrics to sparse edge-weighted graphs in the context of hierarchical clustering. We have demonstrated that our framework can accommodate various cost functions, thanks to efficient algorithms that we have carefully designed with automatic differentiation in mind. Experiments carried on simulated and real data allowed us to show that the proposed approach provides good approximate solutions to well-studied problems.

The theoretical analysis of our optimization framework is beyond the scope of this paper. Nonetheless, we believe that statistical physics modelling \cite{carleo2019} may be a promising direction for future work, based on the observation that ultrametricity is a physical property of spin-glasses \cite{Grossman1989, leuzzi1999, katzgraber2009, katzgraber2012, baviera2015, jagannath2017}. Other possible extensions include the end-to-end learning of neural networks for hierarchical clustering, possibly in the context of image segmentation.

%% file: supplemental.tex
\begin{figure}
	\centering
	\begin{subfigure}[b]{0.37\textwidth}%
		\includegraphics[width=\textwidth]{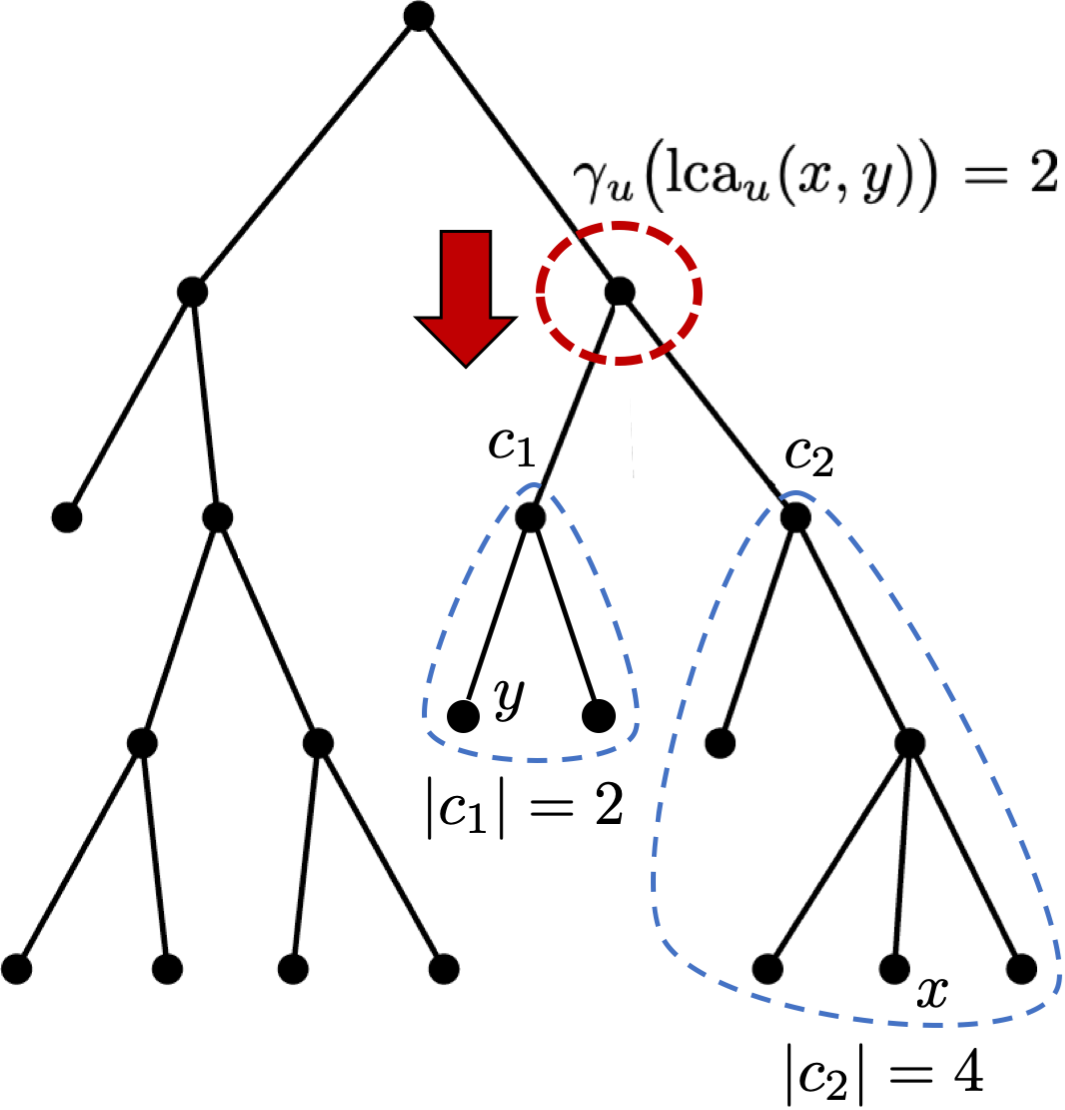}%
		\caption{The cluster-size regularization $J_{\rm size}$ pushes down the nodes of the hierarchical clustering having at least one small child. This corresponds to reducing the distance between the elements in the cluster, effectively preventing small clusters to appear at high altitudes. }
		\label{fig:regularization:size}
	\end{subfigure}
	\hfill
	\begin{subfigure}[b]{0.45\textwidth}%
		\includegraphics[width=\textwidth]{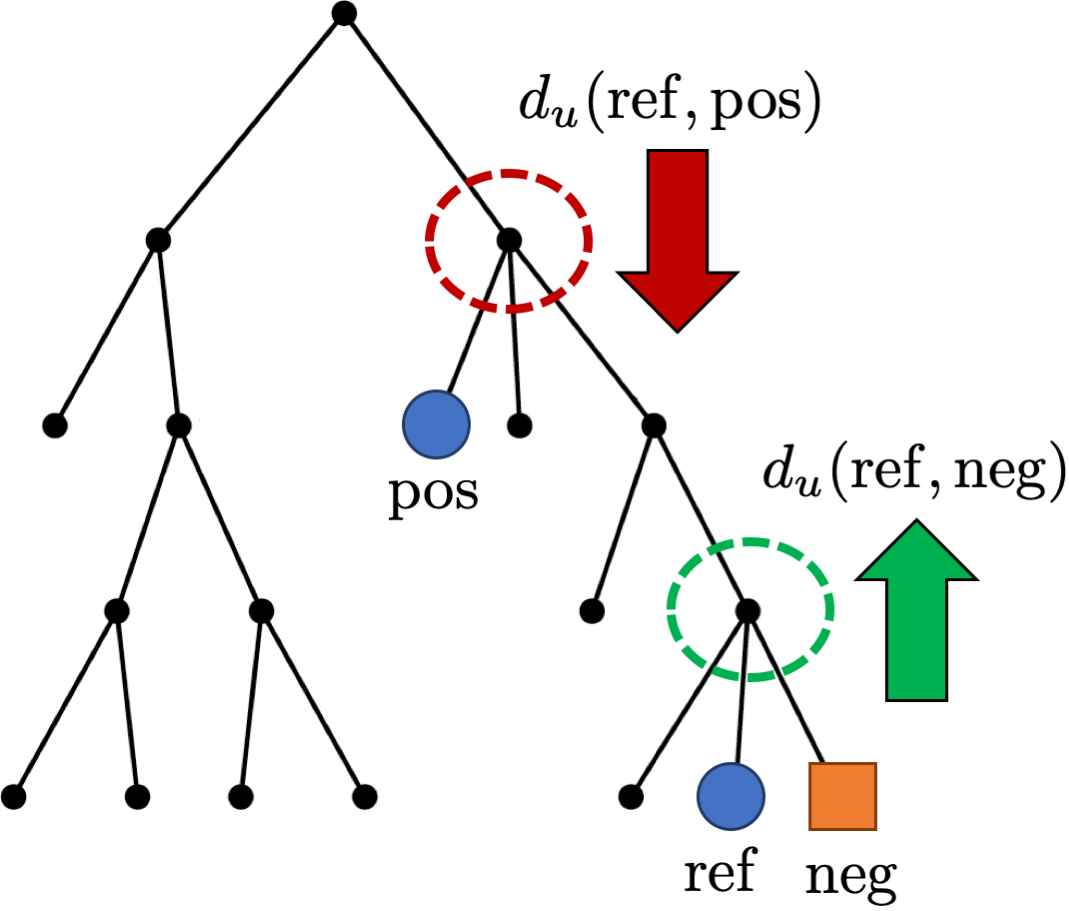}%
		\caption{The triplet regularization $J_{\rm triplet}$ pushes down the lowest common ancestor between elements of the same class (\ie{} it reduces the intra-class distance) and pushes up the lowest common ancestor between elements of different classes (\ie{} it increases the inter-class distance). }
		\label{fig:regularization:triplet}
	\end{subfigure}
	\caption{Intuitive interpretation of the proposed regularization schemes.}
	\label{fig:regularization}
\end{figure}

\subsection{Algorithms}
This section describes in detail the algorithms proposed to compute the cost terms associated to cluster-size regularization, triplet regularization, and Dapgusta cost function.

\paragraph{Cluster-size regularization} This regularization penalizes small clusters at large scales (see Figure \ref{fig:regularization:size}), and the associated cost is computed by Algorithm~\ref{algo:clustersize}.
It proceeds by first computing the size of the smallest child of each node of the tree. 
The size of each node can be trivially computed recursively from the leaves to the root in linear time. 
Then, it identifies the pass node associated to each edge thanks to he fast l.c.a.\ algorithm, and it deduces the individual cost for this edge. Note that we assume the weights $\gamma_u$ are constants, even though they depend on the variable $u$ being optimized. This allows us to simplify the gradient evaluation. Moreover, the algorithm presents an additional hyper-parameter $k$ for applying the regularization only to the top-$k$ dendrogram nodes. In the algorithm we denote by $\textrm{rank}(n)$, the rank of a node $n$ according to the ordering given by their altitudes (from highest to lowest values). The root of the tree is thus ranked 1, the second highest node is ranked 2 and so on. Note that in practice, the single linkage algorithm (as any agglomerative clustering method) naturally orders nodes according to this ordering and no extra computation is required. We set $k=10$ in all our numerical experiments. 

\begin{algorithm}[t]
\caption{Cluster-size regularization \eqref{eq:area} }
\label{algo:clustersize}
\begin{algorithmic}[1]
	\Require{Graph $\G=(\V,\E)$ with ultrametric $u$} 
	\Require{Parameter $k$ to apply regularization only on the top-$k$ nodes}
	\Require{Output the cluster-size regularization value}
	\State $tree \leftarrow \textrm{single-linkage}(\G,\w)$  \Comment{$\mathcal{O}(N \log N)$ with\cite{Gower1969,NajmanISMM2013}}
	\State $area \leftarrow $ cardinal of each node of $tree$ \Comment{$\mathcal{O}(N)$} 
	\ForEach{node $n$ of $tree$ from the leaves to the root (excluded)} \Comment{$\mathcal{O}(N)$} 
        \State $min\_area\_children(n) \leftarrow \infty$ \Comment{$\mathcal{O}(1)$}
        \ForEach{child $c$ of $n$}  \Comment{$\mathcal{O}(1)$}
           
            \State $min\_area\_children(n) \leftarrow \min(min\_area\_children(n), area(c))$ \Comment{$\mathcal{O}(1)$}
        \EndFor
	\EndFor
	\State preprocess l.c.a on $tree$  \Comment{$\mathcal{O}(N)$ with\Cite{Bender2000}}
	\State $reg \leftarrow 0$ \Comment{$\mathcal{O}(1)$}
	\ForEach{edge $e_{xy}\in\E$} \Comment{$\mathcal{O}(N)$} 
	    \State $pass\_node \leftarrow lca_{tree}(x,y)$ \Comment{$\mathcal{O}(1)$ with\Cite{Bender2000}} 
			\If{$\textrm{rank}(pass\_node) \leq k$} \Comment{$\mathcal{O}(1)$}
				\State $reg \leftarrow reg + u(e_{xy}) / min\_area\_children(pass\_node) $ \Comment{$\mathcal{O}(1)$}
			\EndIf
    \EndFor
	\State \textbf{return} $reg\Big.$
\end{algorithmic}
\end{algorithm}

\paragraph{Triplet regularization} This regularization for semi-supervised learning enforces triplet constraints (see Figure \ref{fig:regularization:triplet}), and the associated cost is computed by 
Algorithm~\ref{algo:triplet}, which is very similar to the subdominant ultrametric algorithm. For every triplet $({\rm ref}, {\rm pos}, {\rm neg})$, it searches for $n_1$ and $n_2$, the smallest clusters containing the pairs $({\rm ref},{\rm pos})$ and $({\rm ref}, {\rm neg})$, with the fast l.c.a.\ algorithm. It then introduces a penalization if the altitude of $n_1$ (\ie{} the distance between ${\rm ref}$ and ${\rm pos}$) is not small enough compared to the altitude of $n_2$ (\ie{} the distance between ${\rm ref}$ and ${\rm neg}$).

\begin{algorithm}[t]
\caption{Triplet regularization \eqref{eq:triplet2}}
\label{algo:triplet}
\begin{algorithmic}[1]
	\Require{Graph $\G=(\V,\E)$ with ultrametric $u$} 
	\Require{Triplets $\mathcal{T \subset V^3}$} 
	\Require{Margin $\alpha\in\RR_+$} 
	\Require{Output the triplet regularization value}
	\State $tree \leftarrow \textrm{single-linkage}(\G,\w)$  \Comment{$\mathcal{O}(N \log N)$ with\cite{Gower1969,NajmanISMM2013}}
	\State preprocess l.c.a on $tree$  \Comment{$\mathcal{O}(N)$ with\Cite{Bender2000}}
	\State $reg \leftarrow 0$  \Comment{$\mathcal{O}(1)$}
	\ForEach{$({\rm ref}, {\rm pos}, {\rm neg}) \in \mathcal{T}$} \Comment{$\mathcal{O}(|\mathcal{T}|)$} 
	    \State $pass\_node_1 \leftarrow lca_{tree}(ref, pos)$ \Comment{$\mathcal{O}(1)$ with\Cite{Bender2000}} 
	    \State $pass\_node_2 \leftarrow lca_{tree}(ref, neg)$ \Comment{$\mathcal{O}(1)$ with\Cite{Bender2000}} 
	    \State $reg \leftarrow reg + \max(0, \alpha + u(\mstEdge{pass\_node_1}) - u(\mstEdge{pass\_node_2}))$ \Comment{$\mathcal{O}(1)$}
    \EndFor
	\State \textbf{return} $reg\Big.$
\end{algorithmic}
\end{algorithm}

\paragraph{Dasgupta cost function}
The difficulty in implementing the proposed relaxation of Dasgupta cost term lies in the soft-cardinal function defined in \eqref{eq:nodecardinaloptim}. 
The main function described in Algorithm~\ref{algo:dapgusta} is similar to previously presented algorithm. 
The soft-cardinal function is computed by algorithm~\ref{algo:softcardinal}.
As with Algorithm~\ref{algo:clustersize},  the size of the nodes of the tree can be computed recursively from leaves to root in linear time. 
Note that, on line 9, the child of $y$ that does not contain $x$ can easily be determined by remembering the previous node of the "for each" loop: it is the sibling of the latter. 

\begin{algorithm}[t]
\caption{Dasgupta cost function \eqref{eq:dasgupta}}
\label{algo:dapgusta}
\begin{algorithmic}[1]
	\Require{Graph $\G=(\V,\E)$ with ultrametric $u$} 
	\Require{Output Dapgusta cost function value}
	\State $soft\_cardinal \leftarrow \textrm{soft-cardinal}((\G,u), tree)$ \Comment{Algorithm~\ref{algo:softcardinal} $\mathcal{O}(N^2)$} 
	\State preprocess l.c.a on $tree$  \Comment{$\mathcal{O}(N)$ with\Cite{Bender2000}}
	\State $cost \leftarrow 0$  \Comment{$\mathcal{O}(1)$}
	\ForEach{edge $e_{xy}\in\E$} \Comment{$\mathcal{O}(N)$} 
	    \State $pass\_node \leftarrow lca_{tree}(x,y)$ \Comment{$\mathcal{O}(1)$ with\Cite{Bender2000}} 
	    \State $cost \leftarrow cost + soft\_cardinal(pass\_node)/u(e_{xy})$  \Comment{$\mathcal{O}(1)$}
    \EndFor
	\State \textbf{return} $cost\Big.$
\end{algorithmic}
\end{algorithm}

\begin{algorithm}[t]
\caption{Soft-cardinal function \eqref{eq:nodecardinaloptim}}
\label{algo:softcardinal}
\begin{algorithmic}[1]
	\Require{Graph $\G=(\V,\E)$ with ultrametric $u$} 
	\Require{Single linkage clustering $tree$ on $(\G, u)$} 
	\Require{Output the soft-cardinal of non leaves node of $tree$}
	\State $area \leftarrow $ cardinal of each node of $tree$ \Comment{$\mathcal{O}(N)$} 
	\State preprocess l.c.a on $tree$  \Comment{$\mathcal{O}(N)$ with\Cite{Bender2000}}
	
	\ForEach{non leaf node $n$ of $tree$} \Comment{$\mathcal{O}(N)$}
	    \State $pass\_edge \leftarrow \mstEdge{n}$  \Comment{$\mathcal{O}(1)$}
	    \State $alt\_n \leftarrow u(pass\_edge)$ \Comment{$\mathcal{O}(1)$}
	    \State $soft\_cardinal(n) \leftarrow 2\times\textrm{sigmoid}(alt\_n)$ \Comment{$\mathcal{O}(1)$}
	    \ForEach{vertex $x$ of $pass\_edge$} \Comment{$\mathcal{O}(1)$}
	        \ForEach{ancestor $y$ of $x$} \Comment{$\mathcal{O}(N)$}
	                \State $c\_other \leftarrow $ child of the $y$ that does not contain $x$ \Comment{$\mathcal{O}(1)$}
	                \State $contrib\_y \leftarrow  area(c\_other)\times \textrm{sigmoid}(alt\_n - u(\mstEdge(y)))$ \Comment{$\mathcal{O}(1)$}
	                \State $soft\_cardinal(n) \leftarrow soft\_cardinal(n) + contrib\_y$ \Comment{$\mathcal{O}(1)$}
	        \EndFor
	    \EndFor
    \EndFor
	\State \textbf{return} $soft\_cardinal\Big.$
\end{algorithmic}
\end{algorithm}

\subsection{Framework validation}
All the tests in the comparison with the CUCP algorithm (Section~\ref{sec:validation}) were conducted on a computer with an Intel I7 4 cores processor and 16\,GB of memory.
For the optimization, we use the AMSGrad variation\Cite{j.2018on} of the ADAM method with step-size $0.01$. 
Our implementation of the proposed algorithms are all single threaded.

\begin{figure}[t]
\centering
\includegraphics[width=0.5\textwidth]{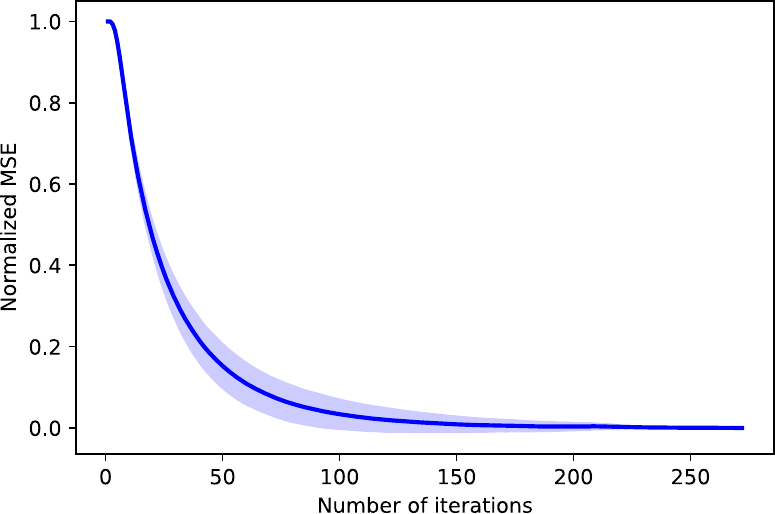}%
\caption{Mean-normalized convergence curve of the proposed approach and standard deviation.}
\label{fig:convergence}
\end{figure}

For each of the 52 test instances, the values $J_{\rm closest}$ obtained at each iteration of Algorithm~\ref{algo:gradient_descent} were normalized between 0 (lowest achieved cost for this instance) and 1 (highest cost reached for this instance). 
Figure \ref{fig:convergence} shows the mean-normalized convergence curve (with its standard deviation). 
We can see, that the convergence rate appears to be very good and smooth in practice. 
The convergence is usually reached within a bit more than a hundred iterations.

\subsection{Illustrative examples}
\begin{figure}[t]
\centering
\image{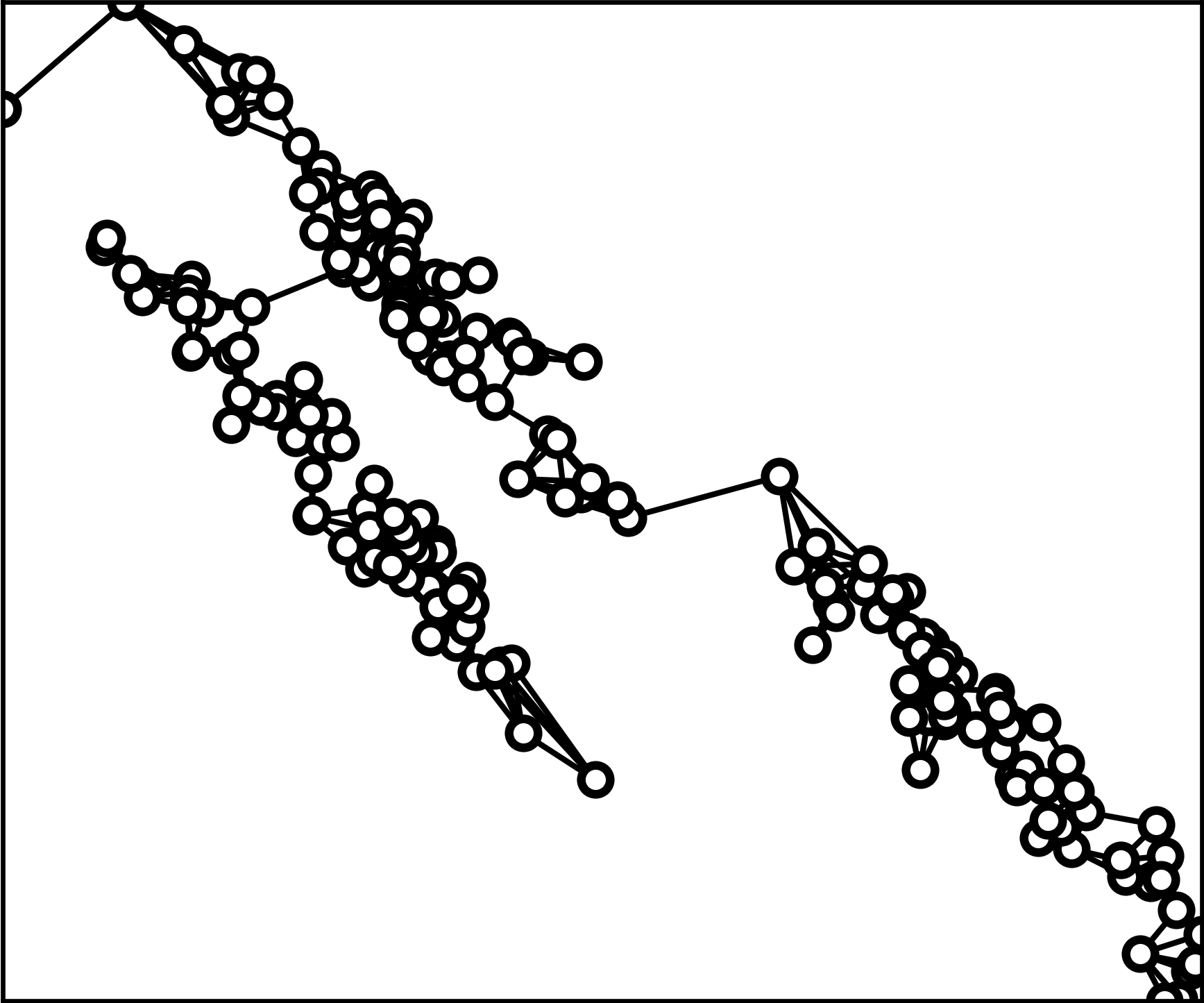}{}
\hfill
\image{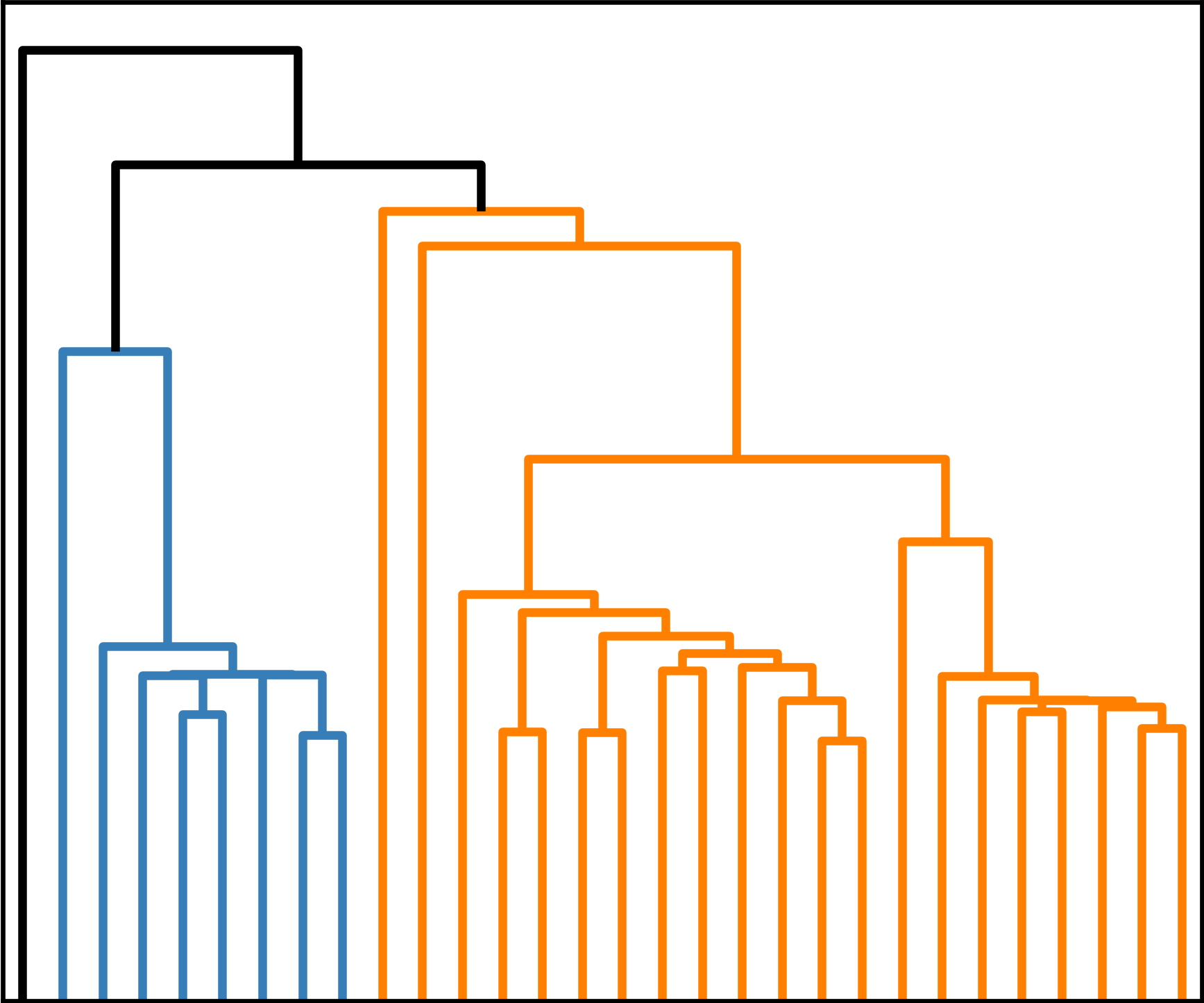}{}
\hfill
\image{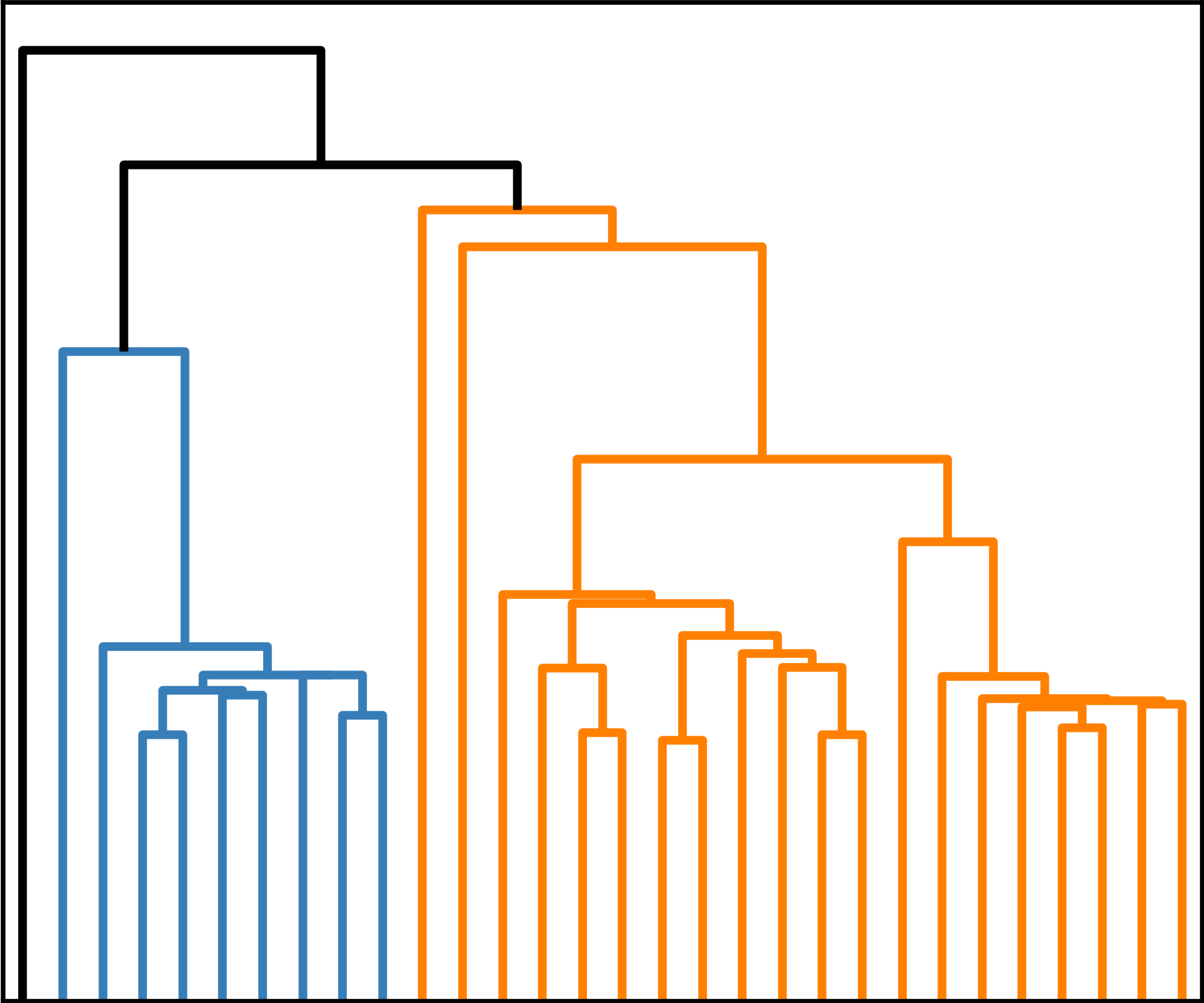}{}
\hfill
\image{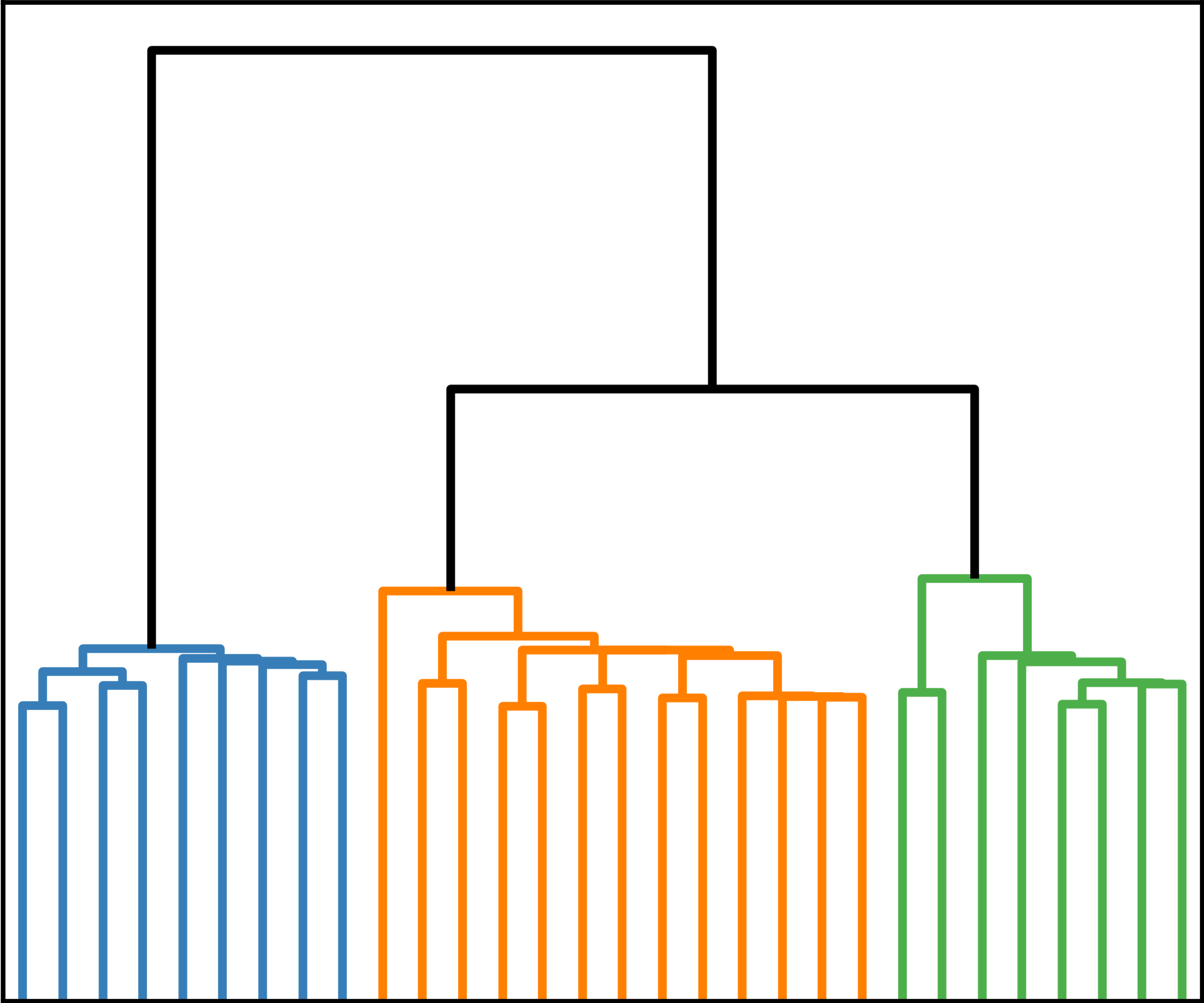}{}
\hfill
\image{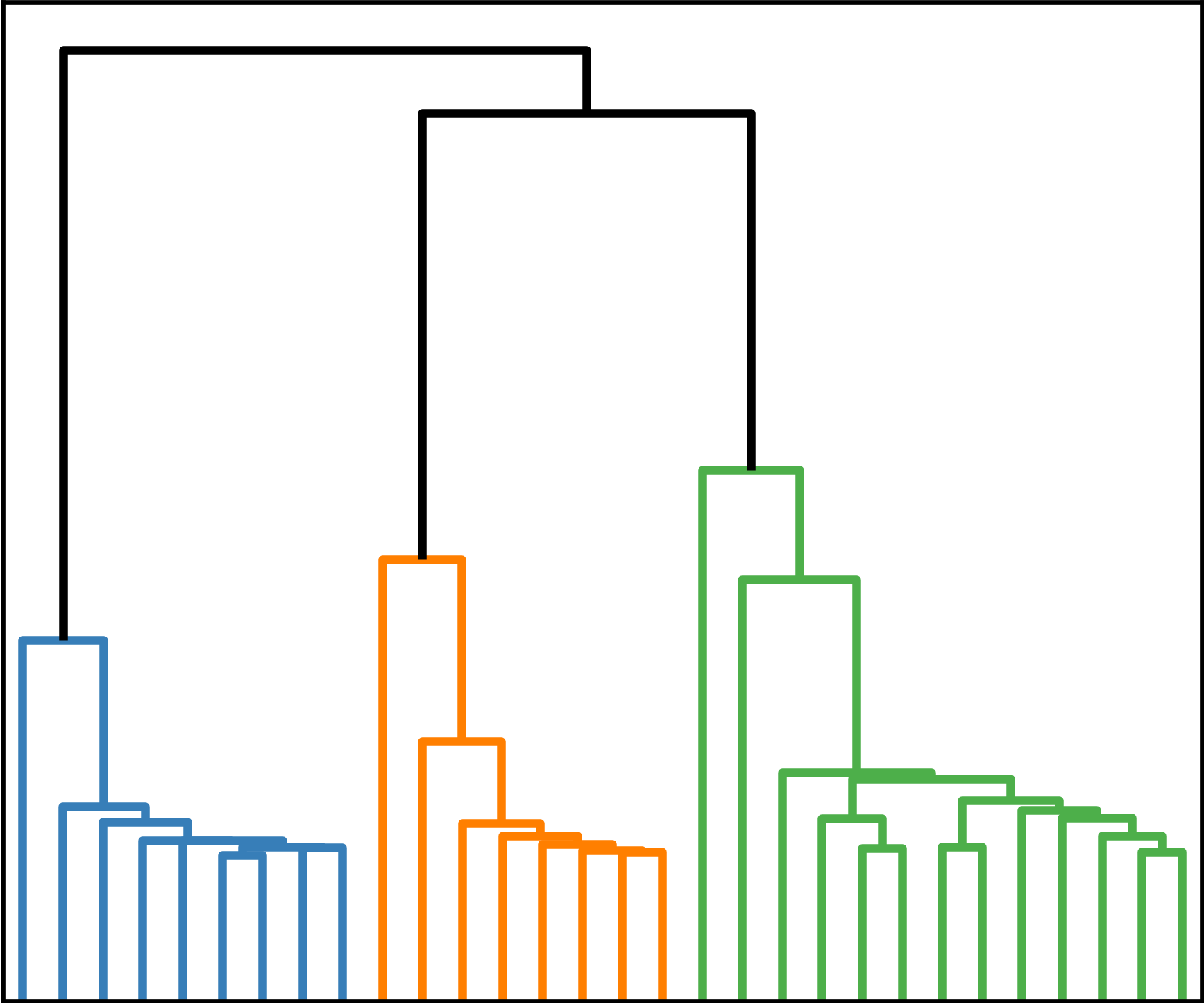}{}
\hfill
\image{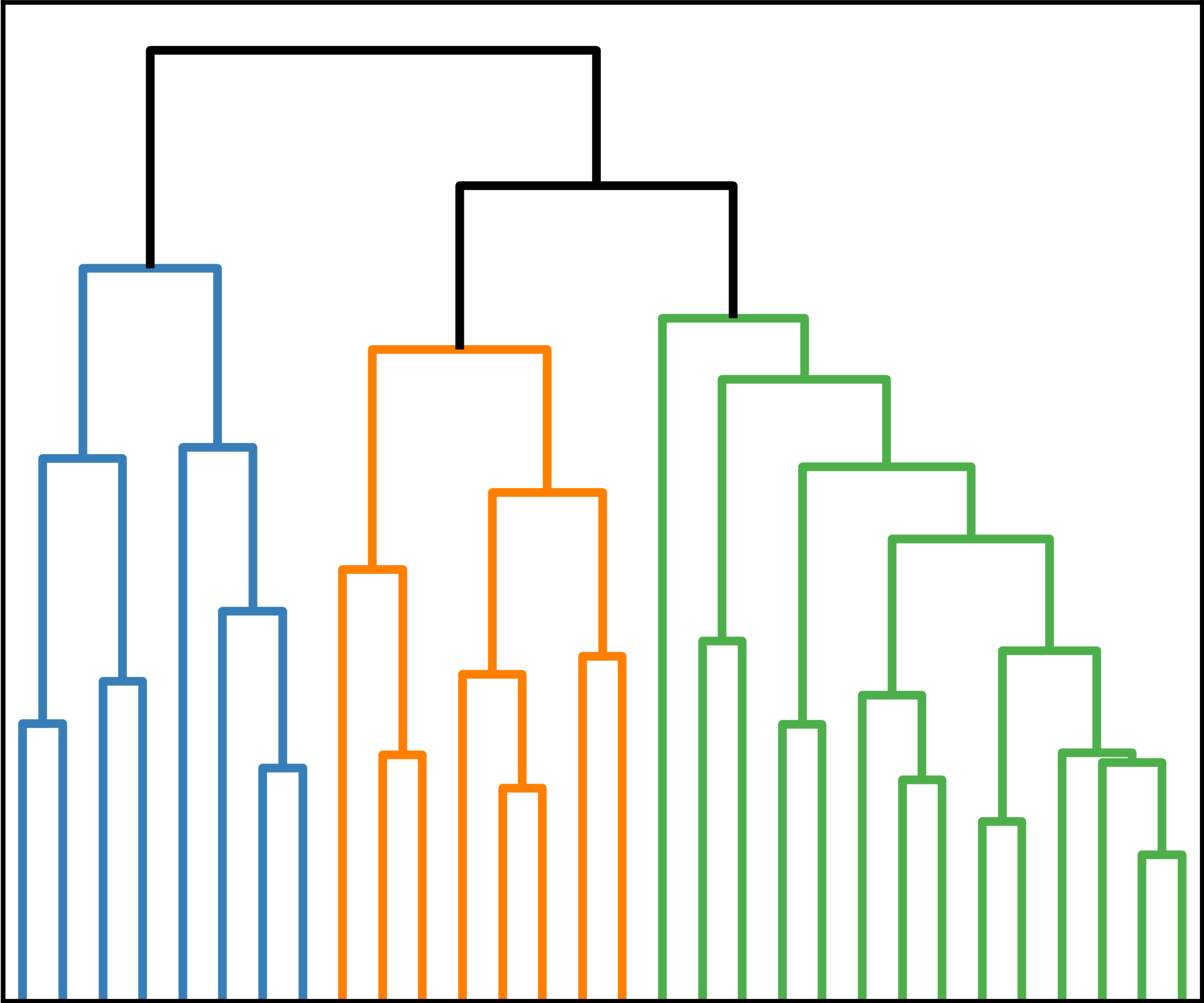}{}

\image{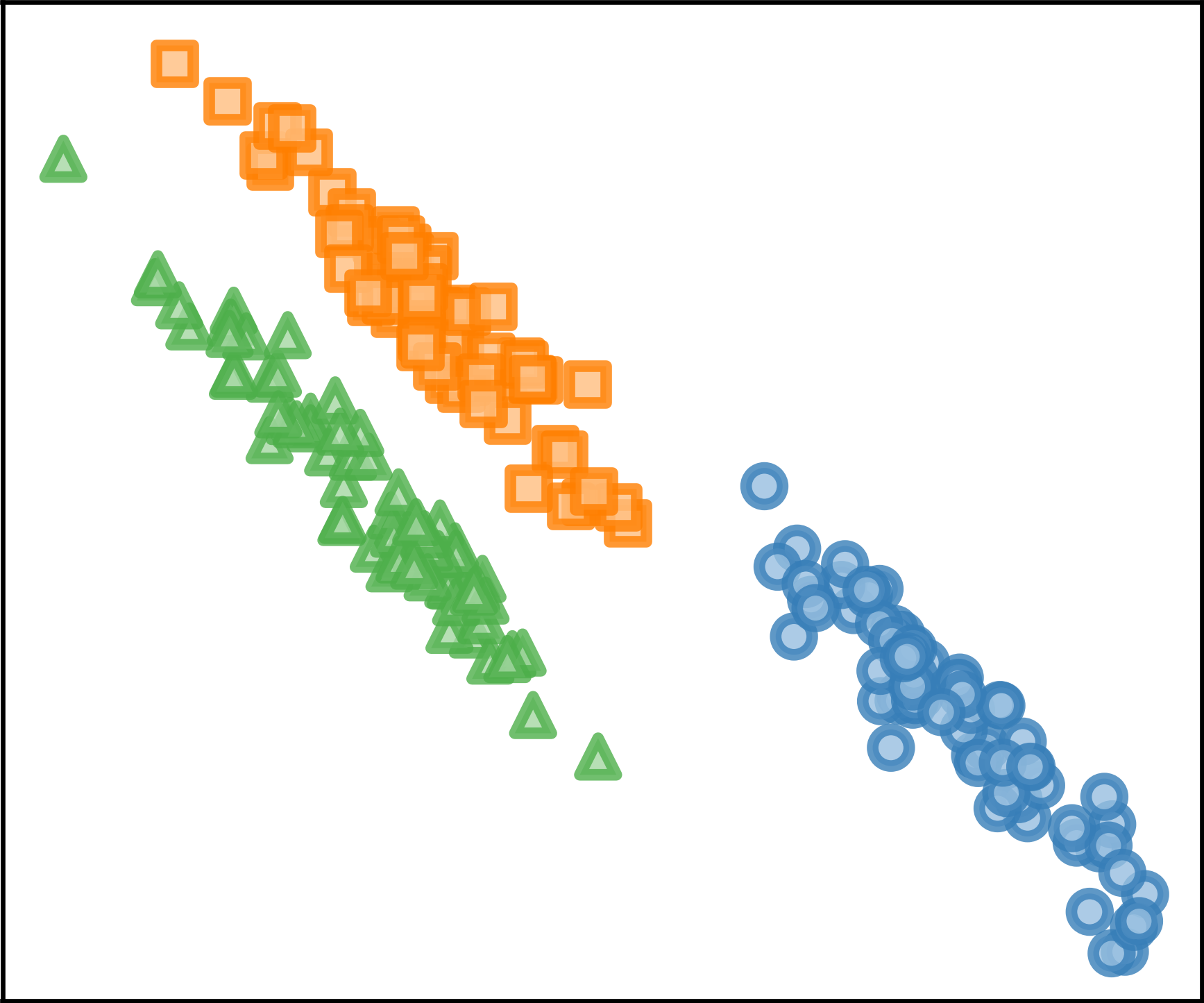}{}
\hfill
\image{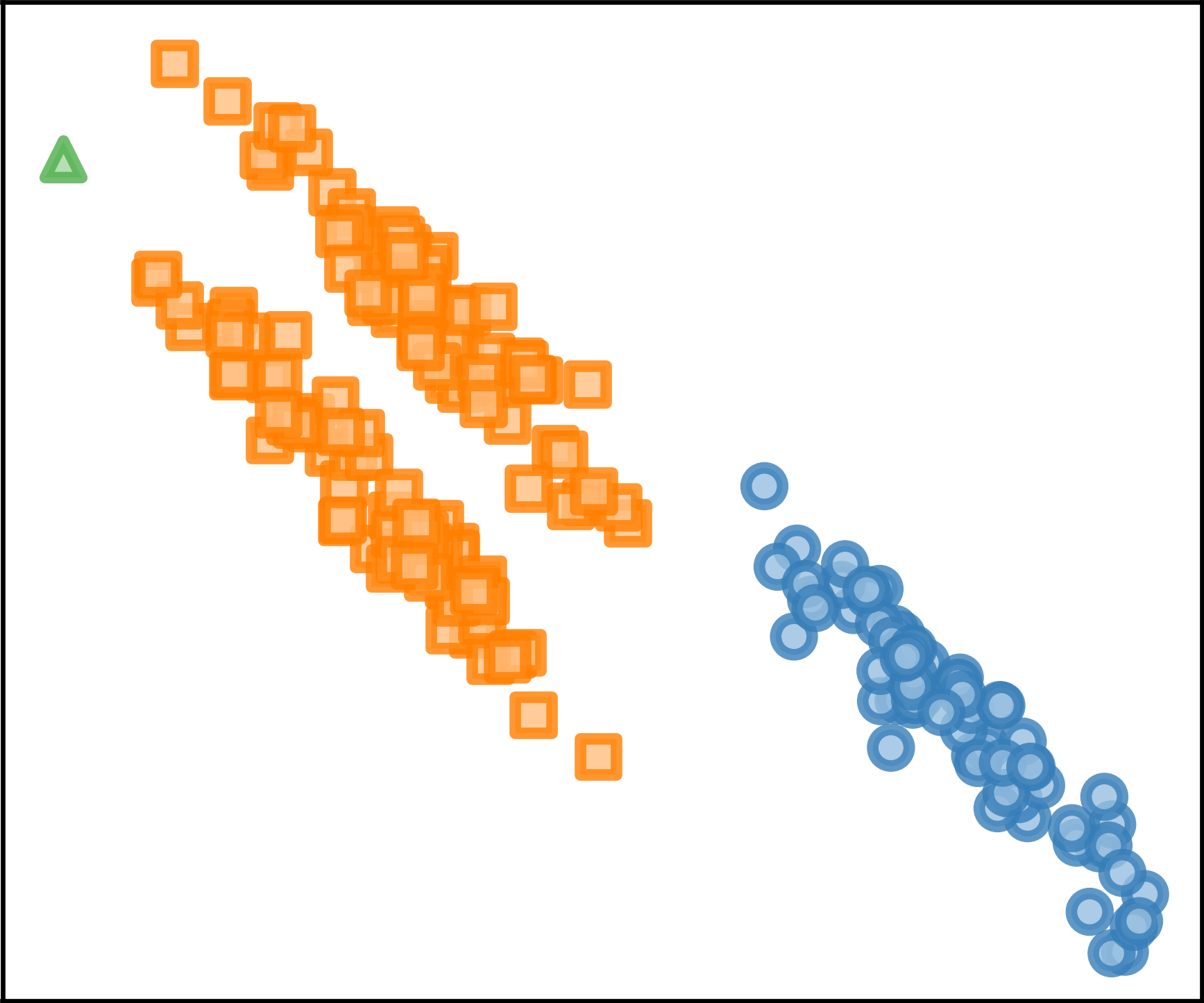}{}
\hfill
\image{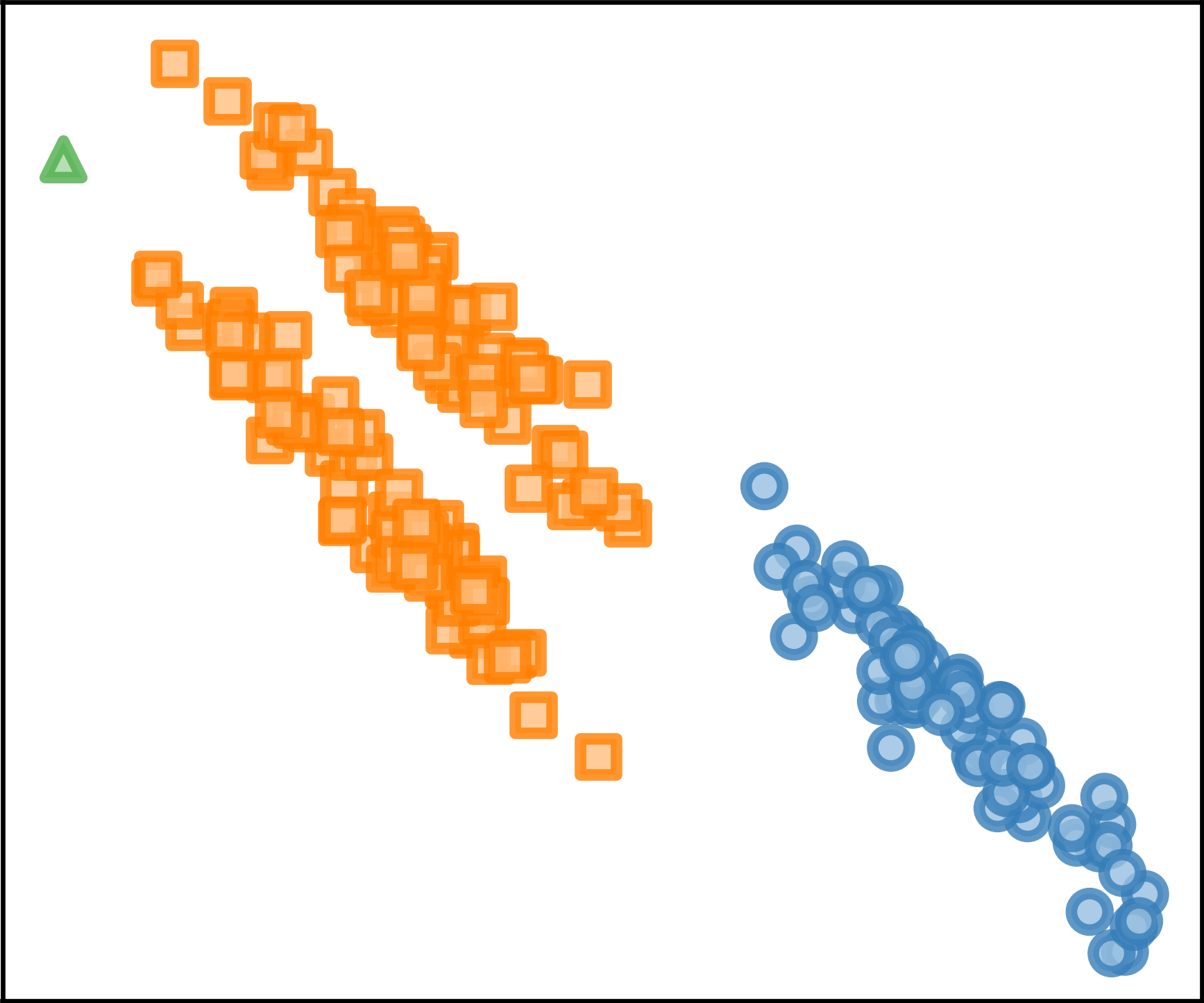}{}
\hfill
\image{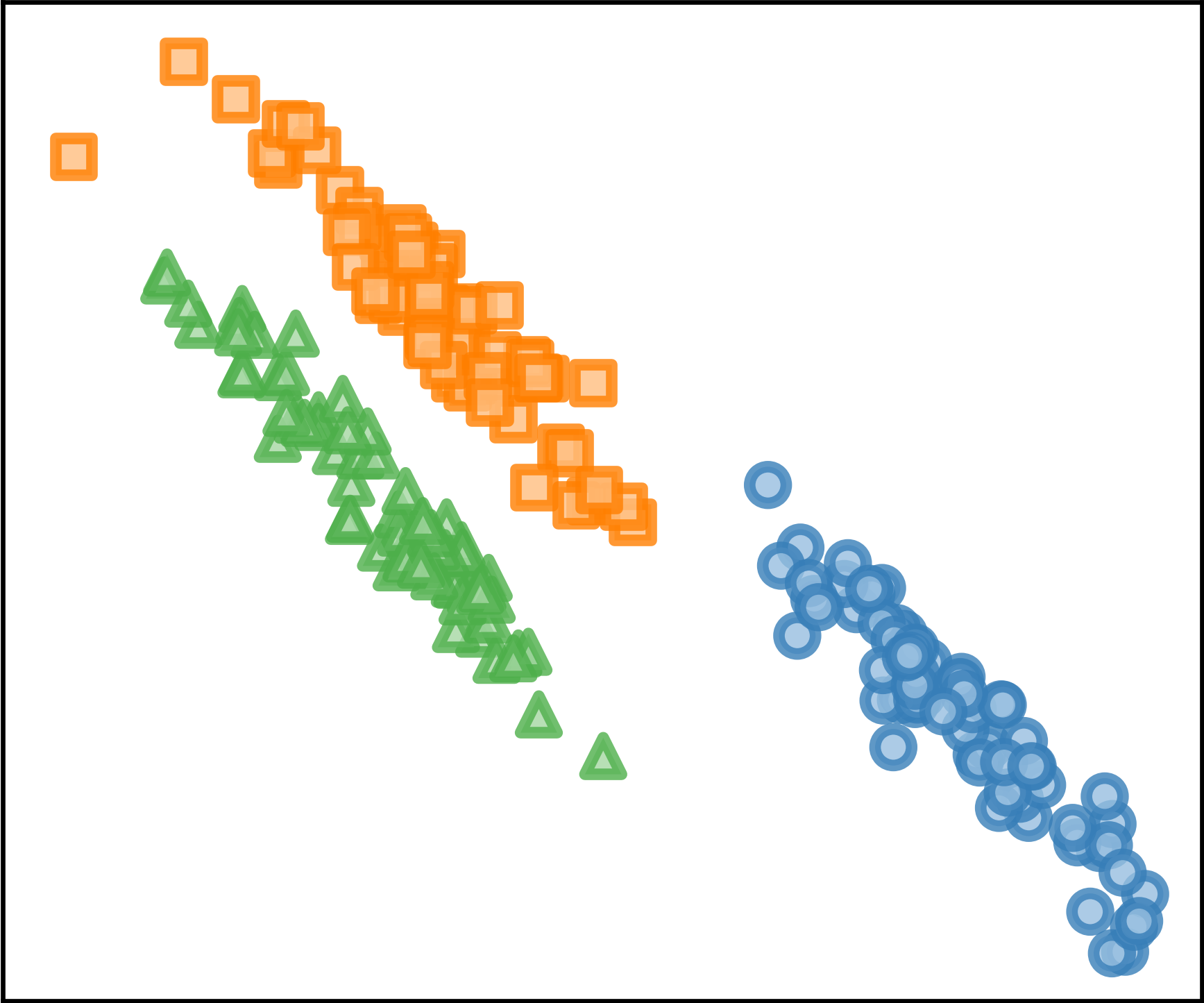}{}
\hfill
\image{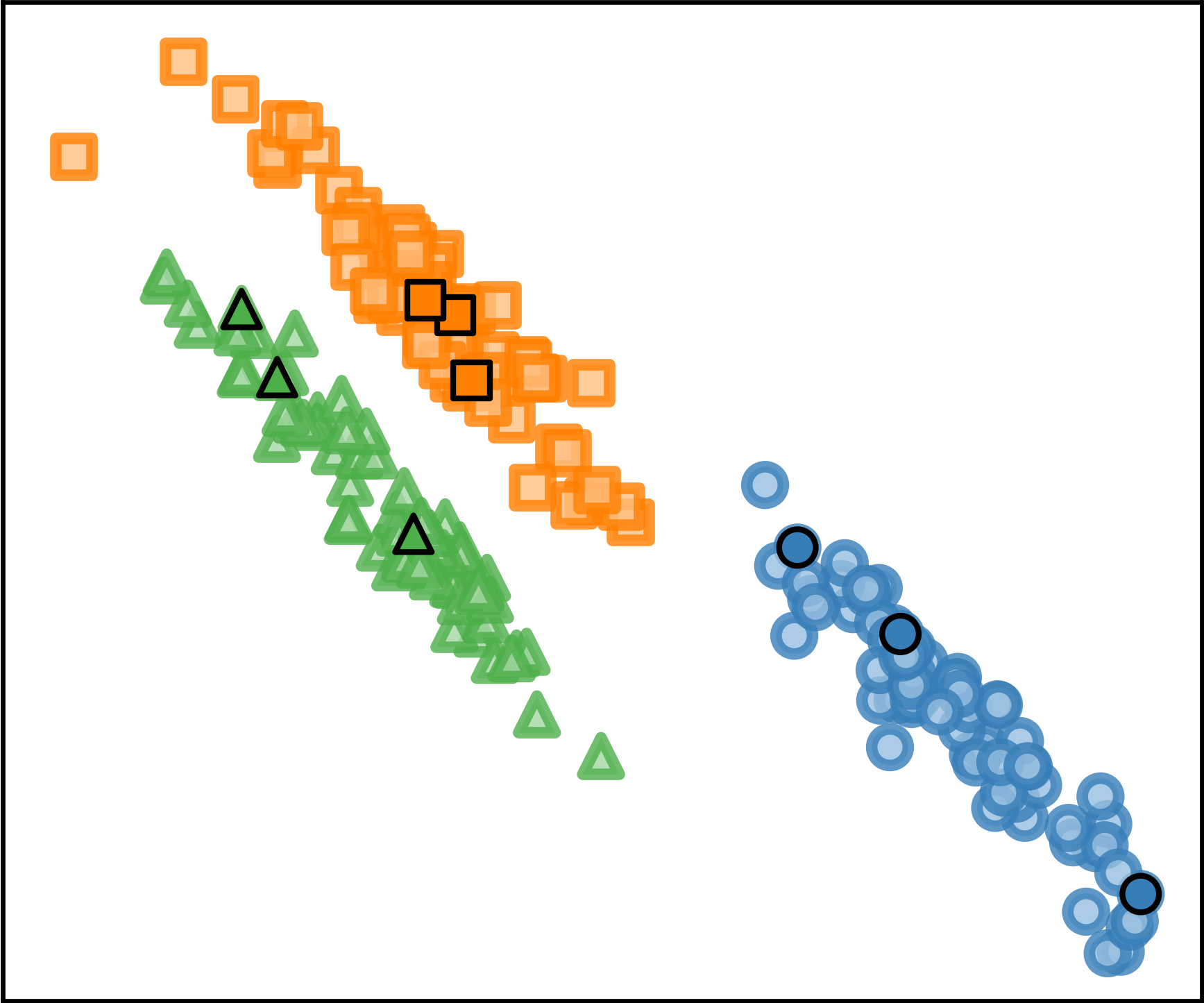}{}
\hfill
\image{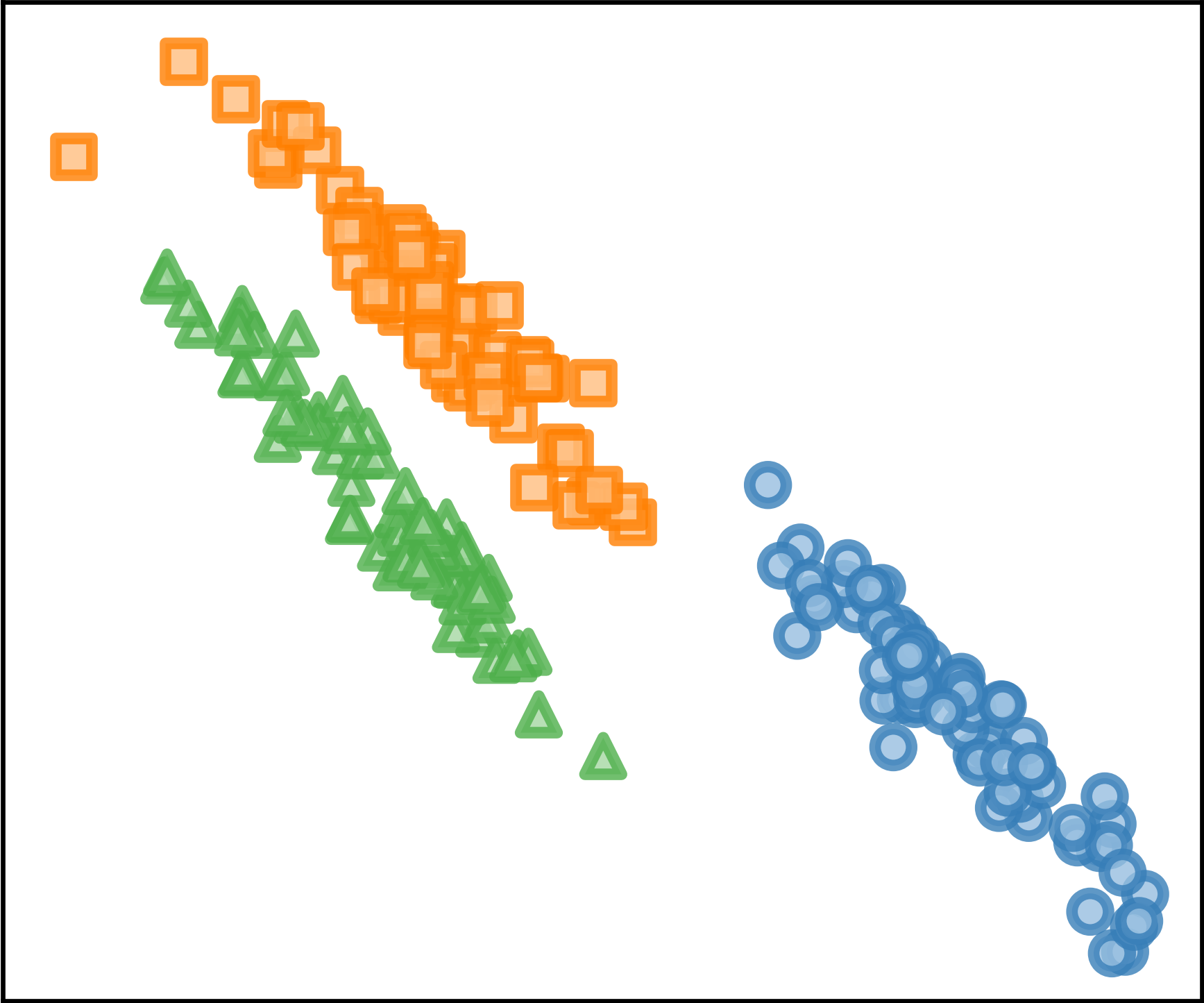}{}

\vspace{1em}

\image{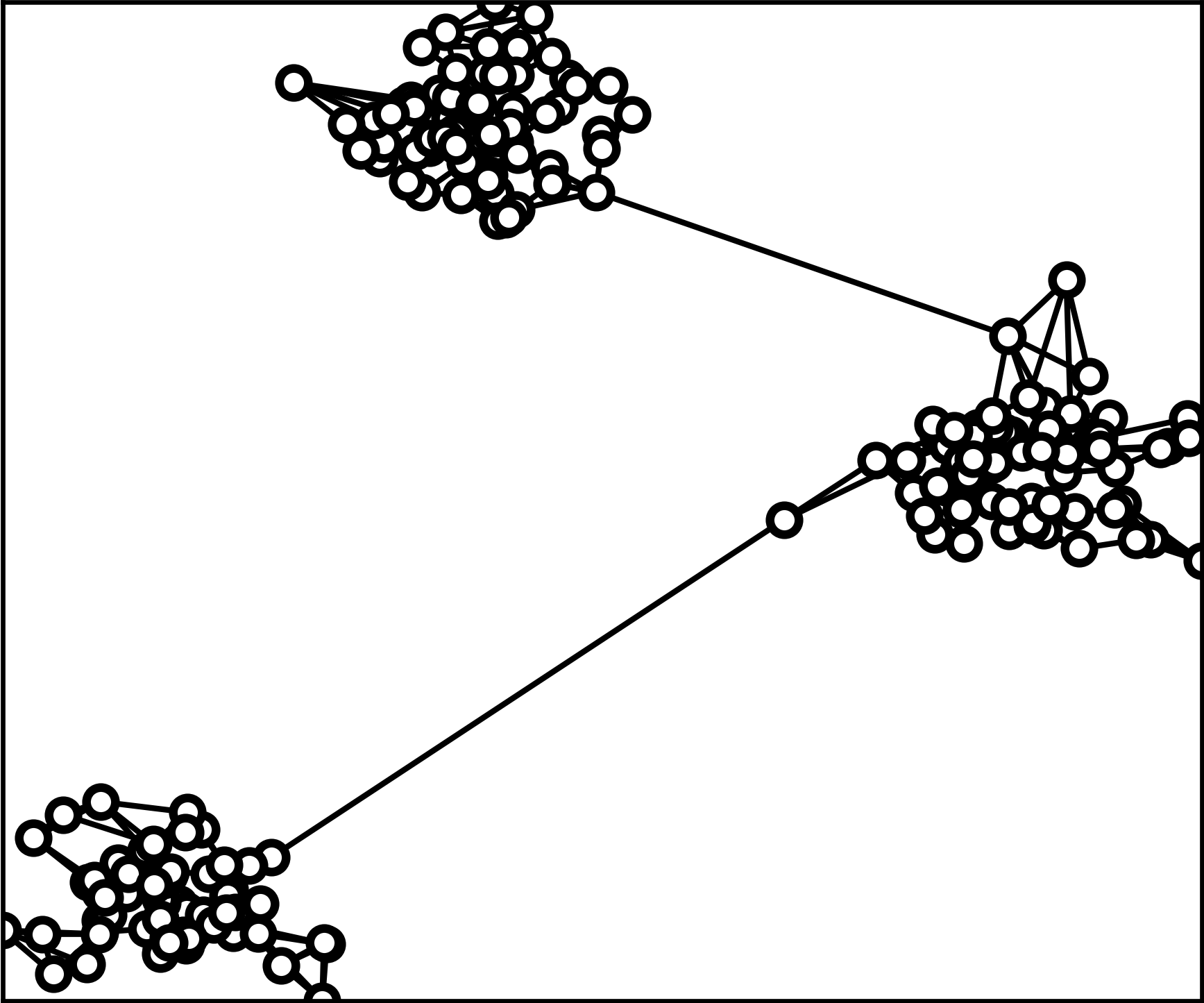}{}
\hfill
\image{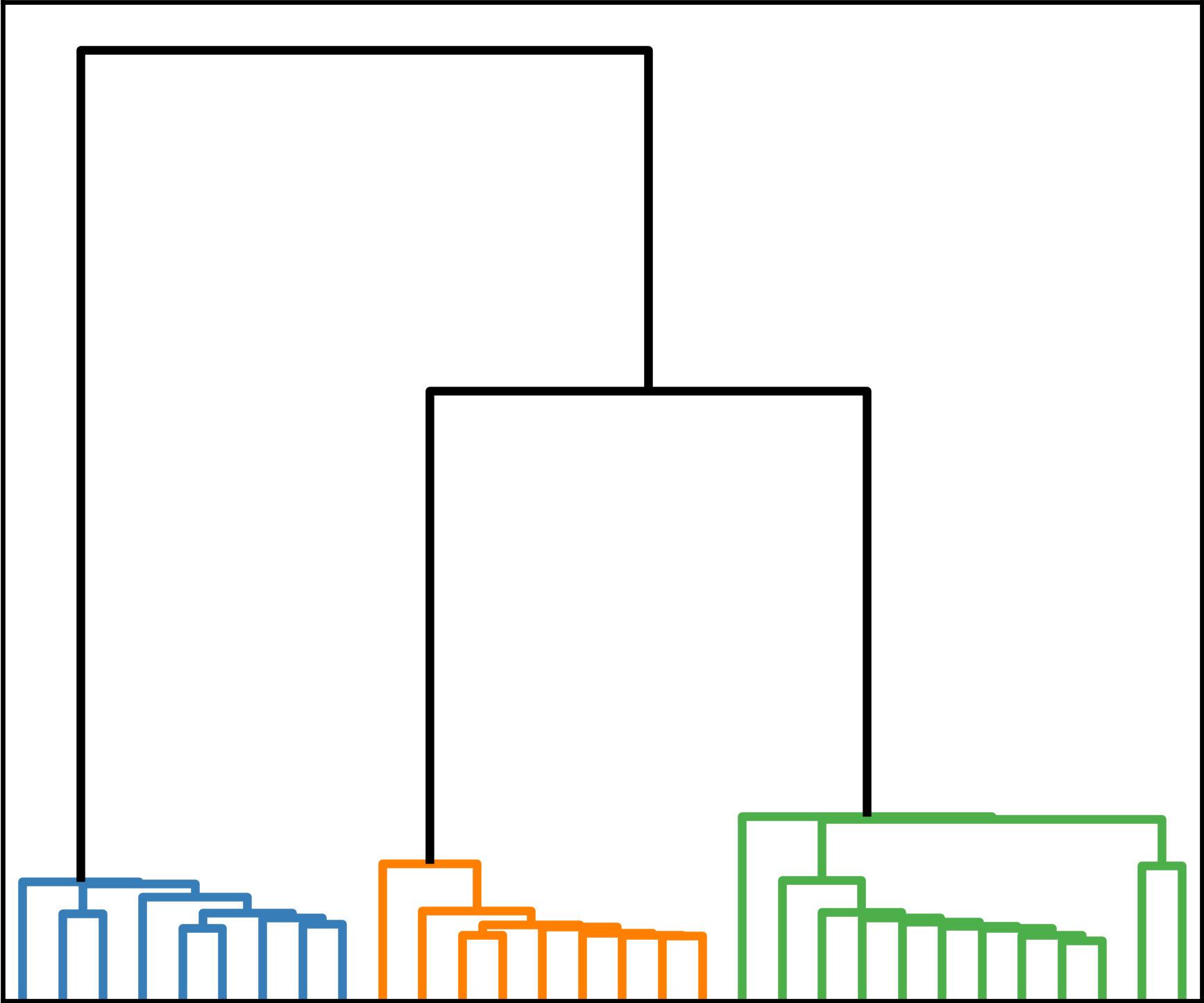}{}
\hfill
\image{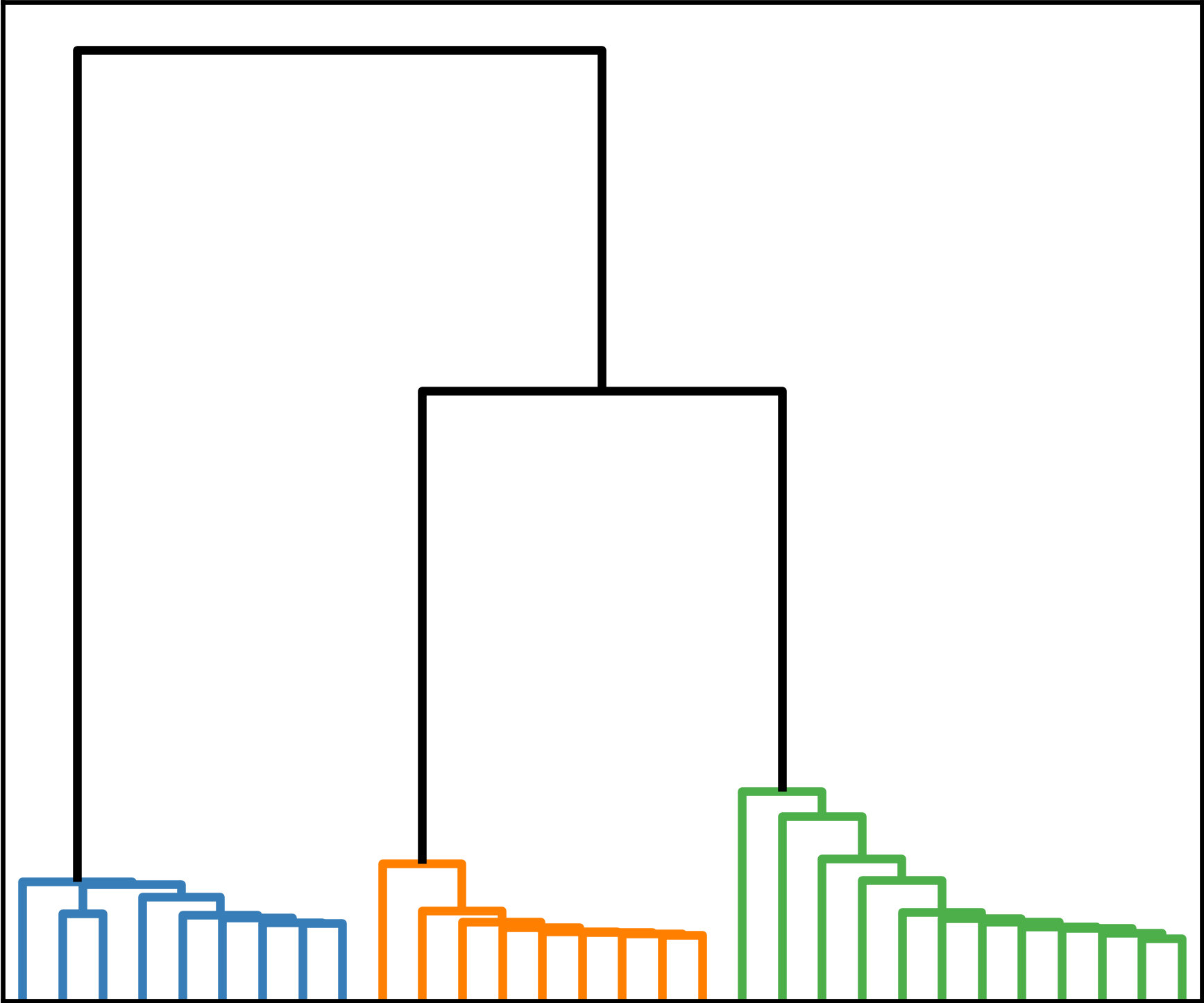}{}
\hfill
\image{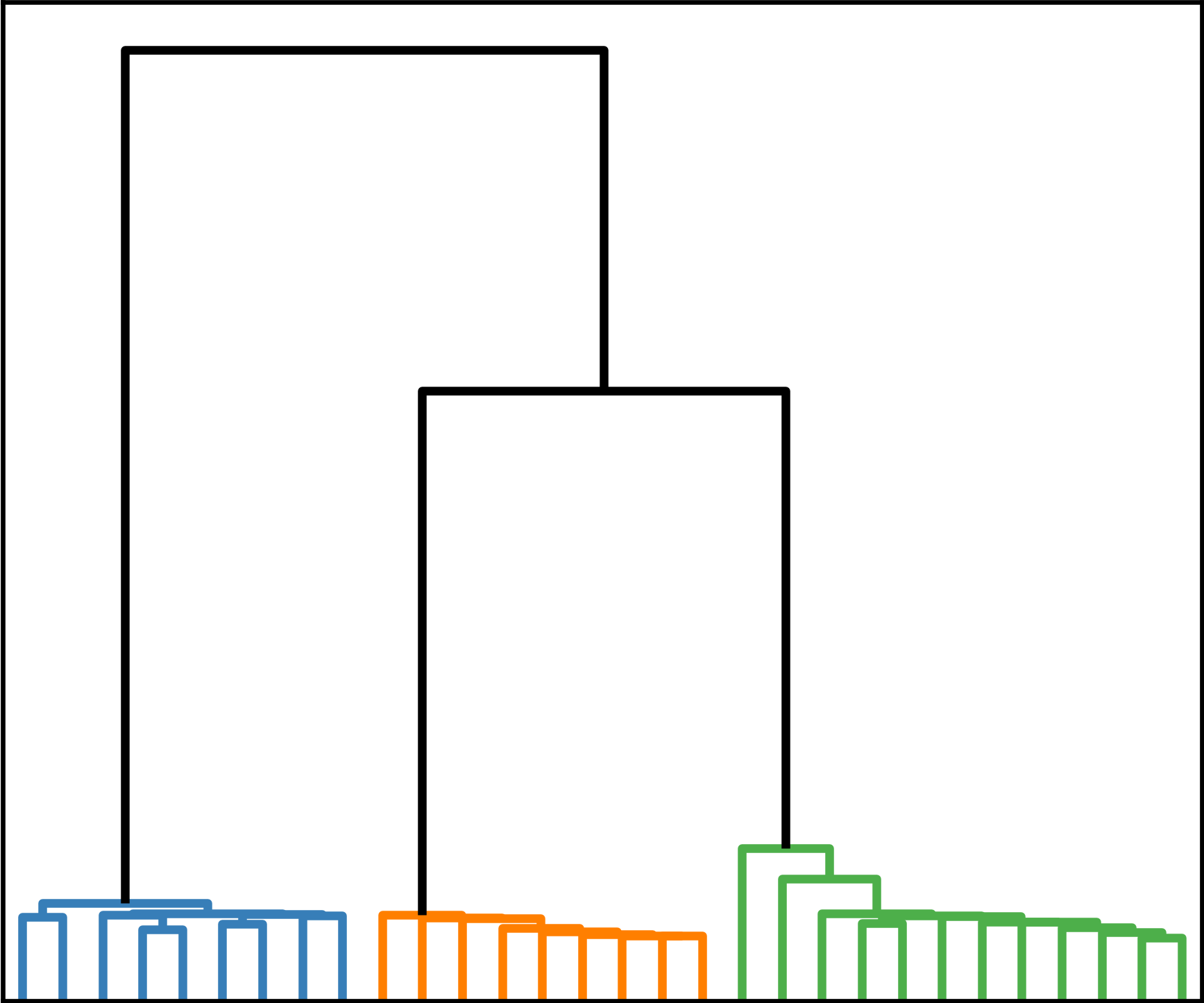}{}
\hfill
\image{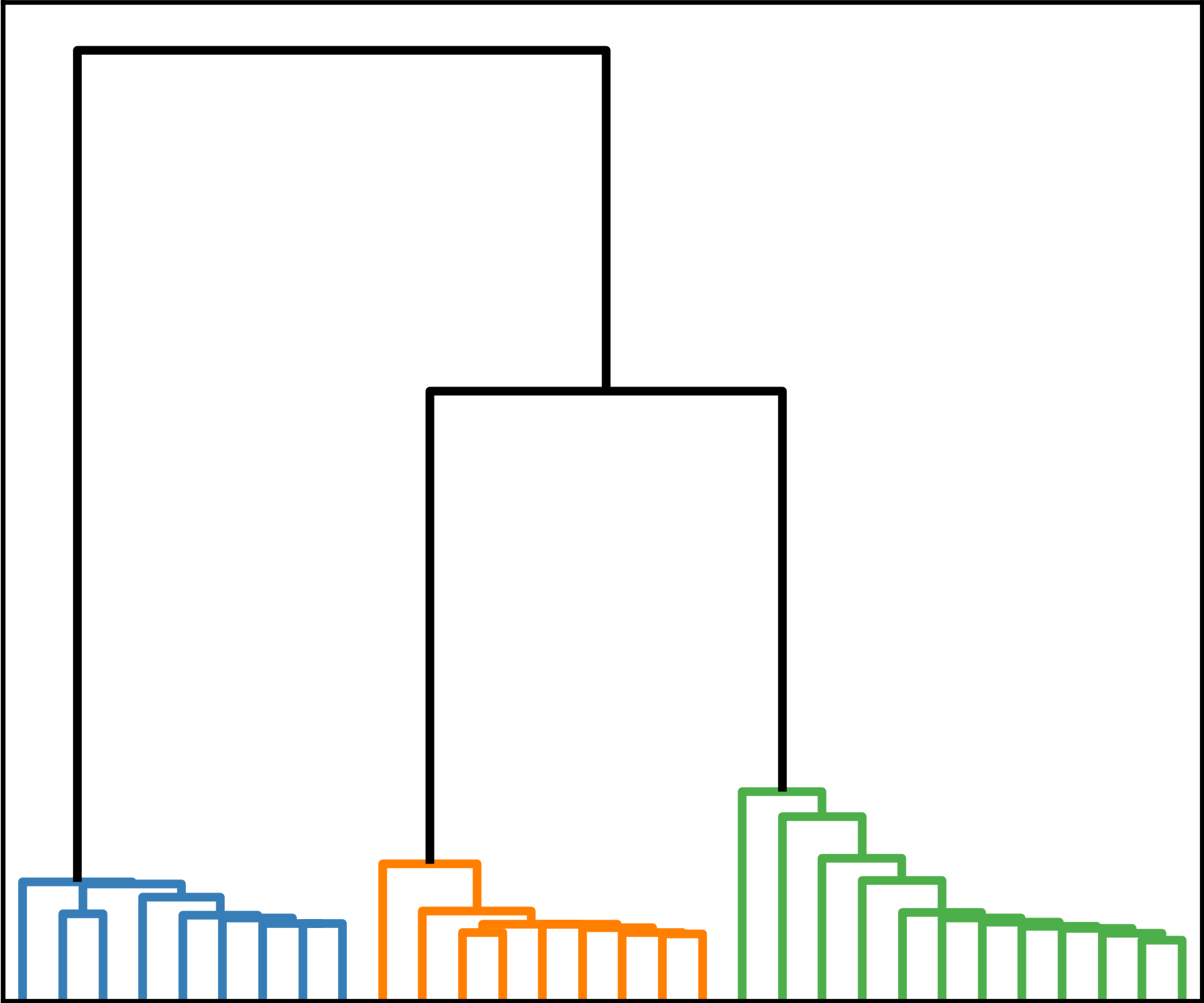}{}
\hfill
\image{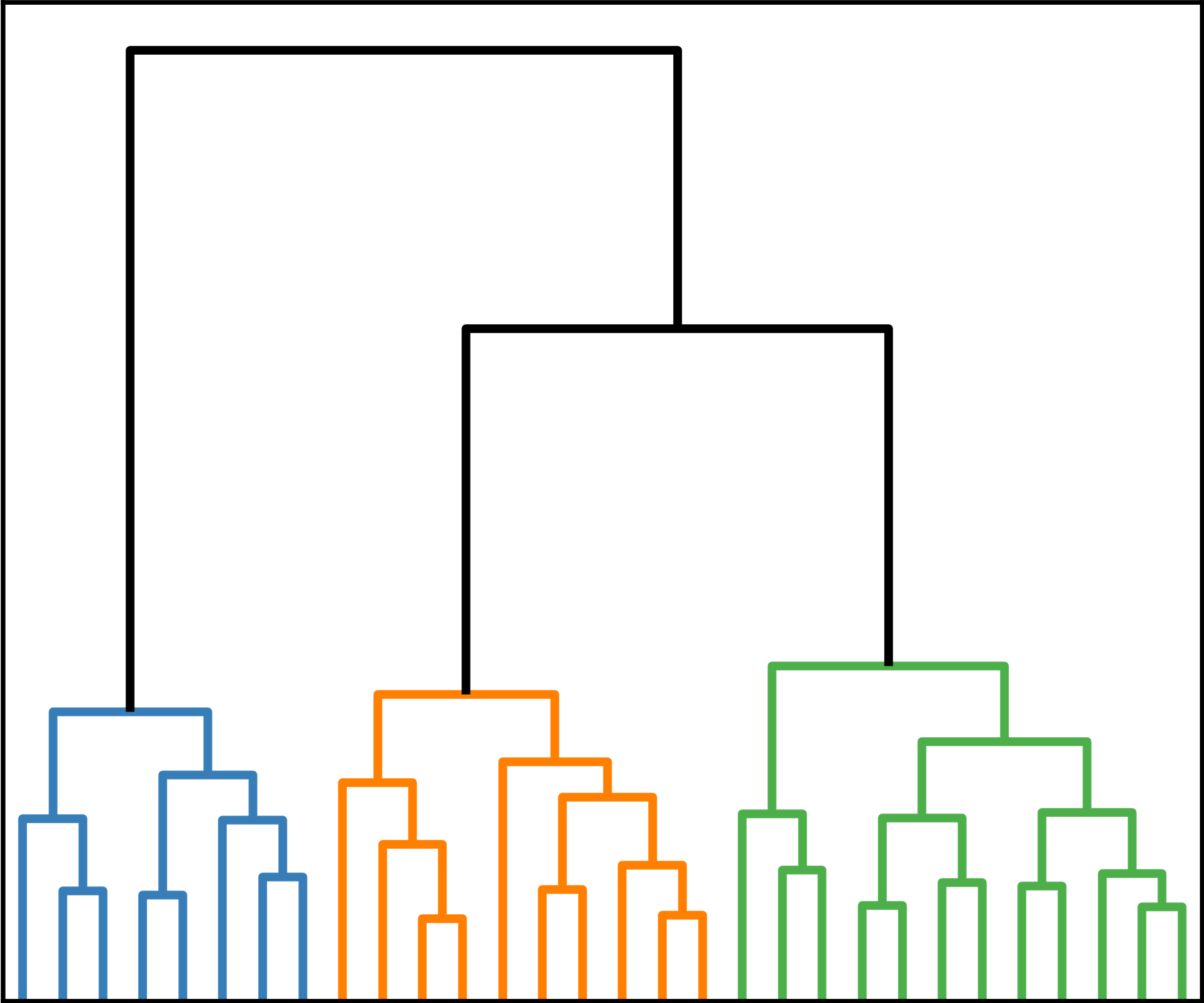}{}

\image{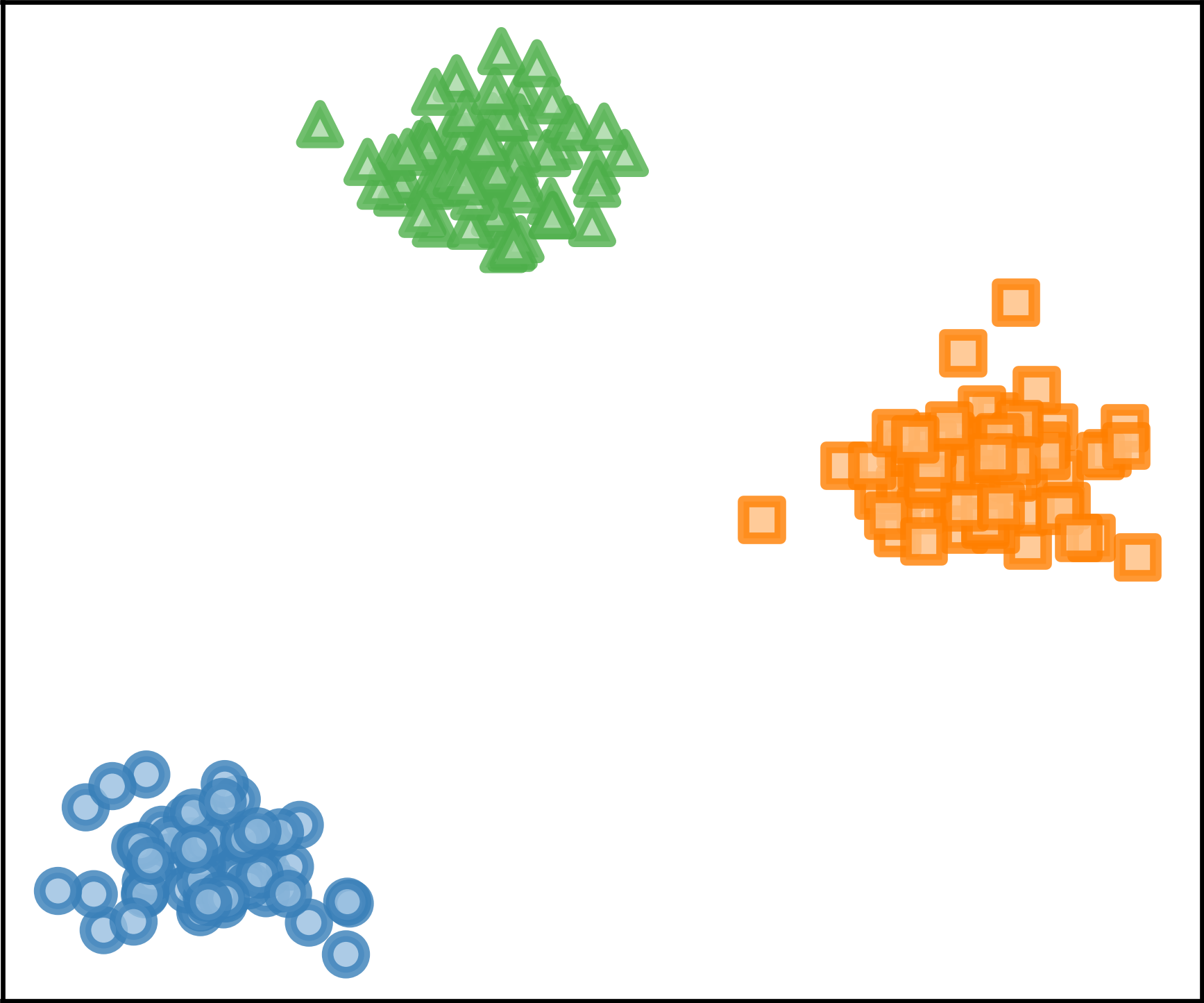}{}
\hfill
\image{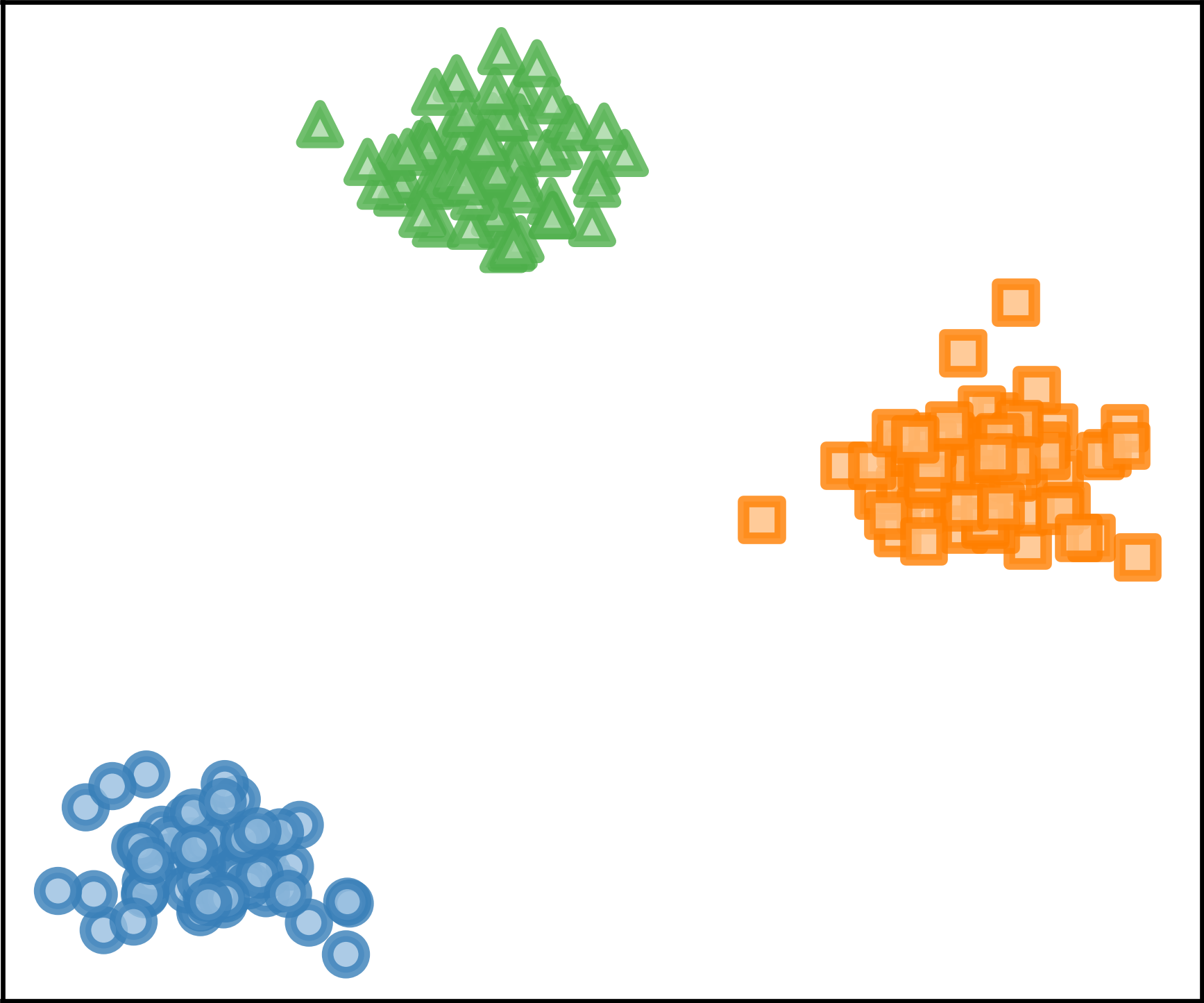}{}
\hfill
\image{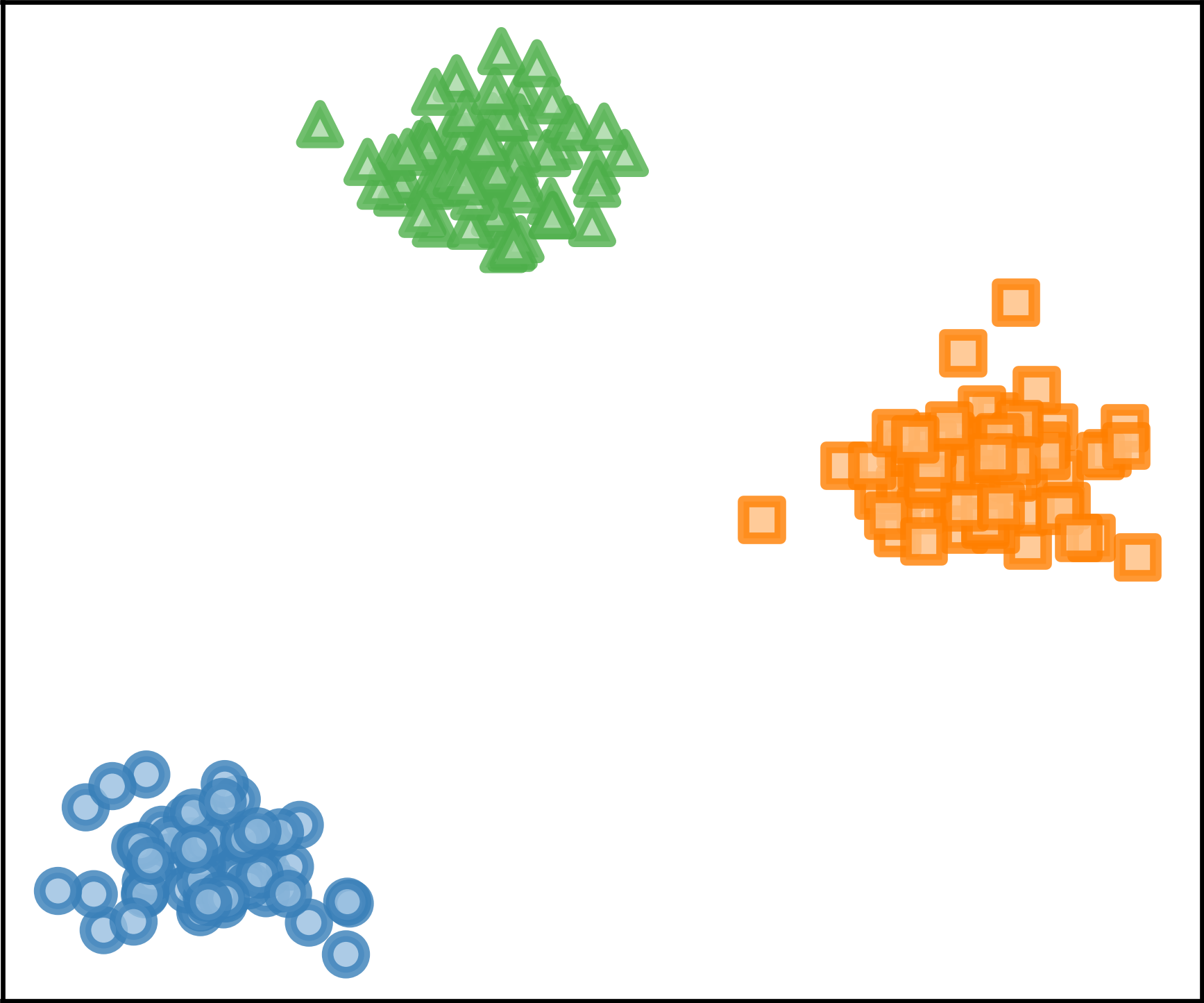}{}
\hfill
\image{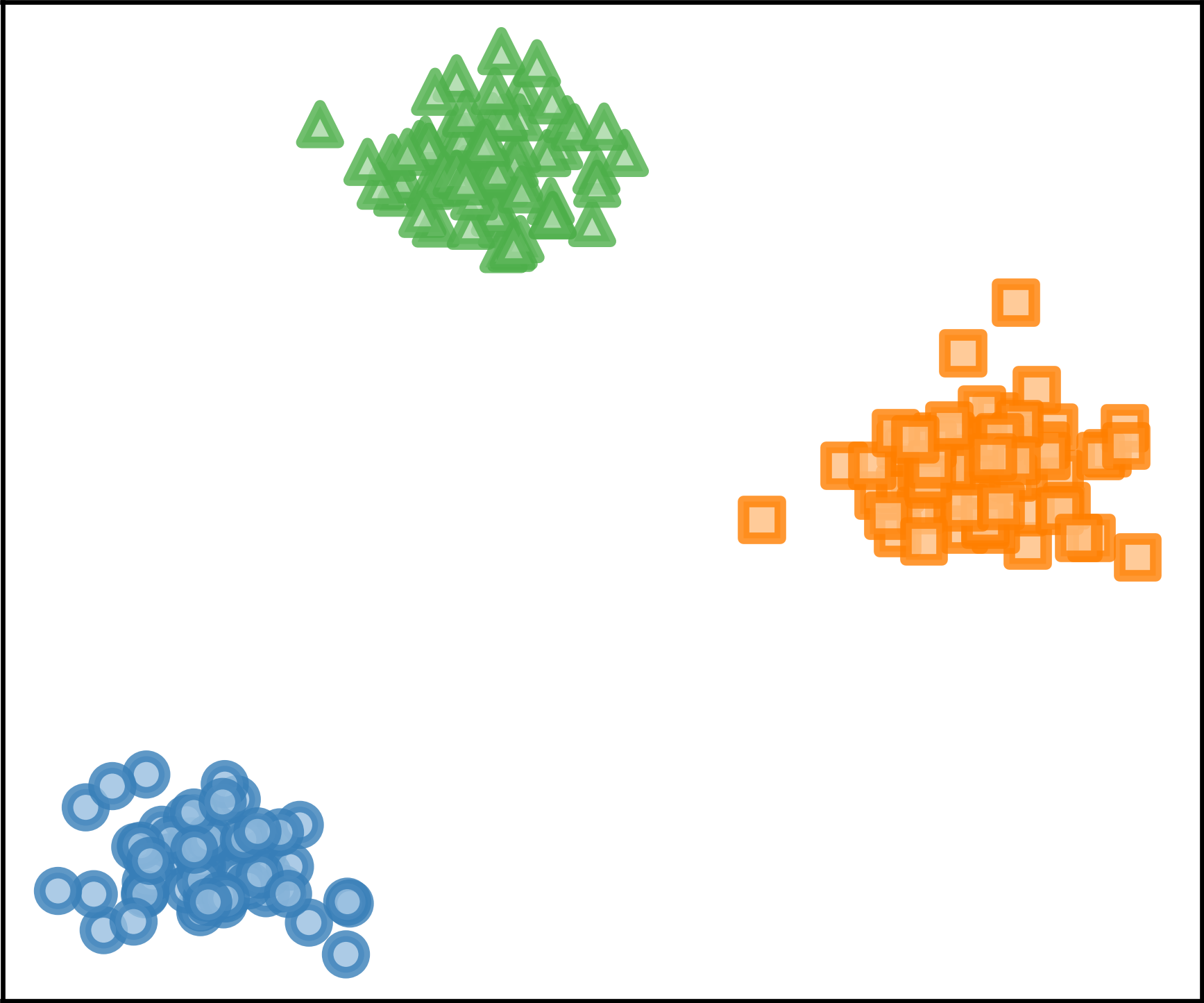}{}
\hfill
\image{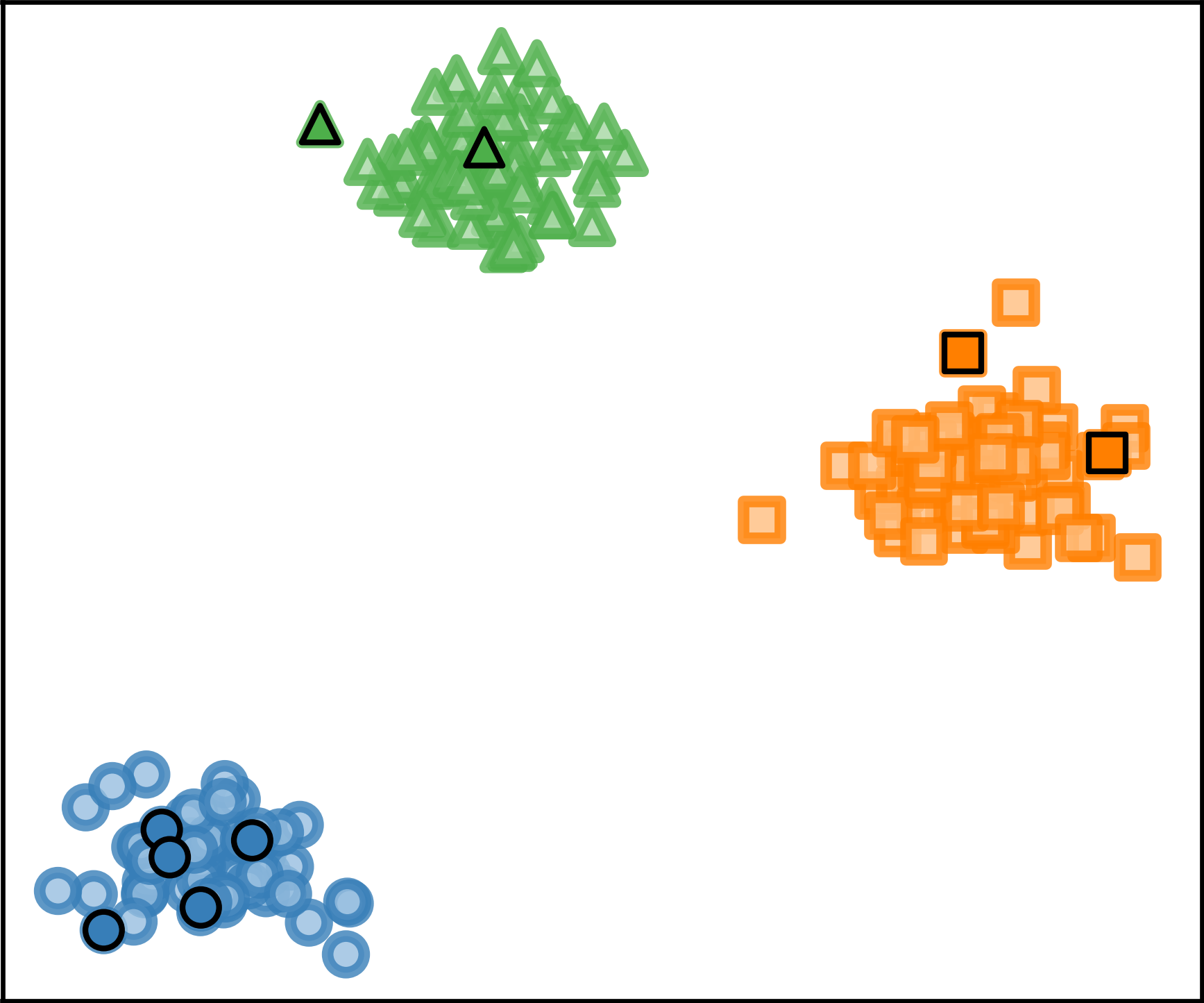}{}
\hfill
\image{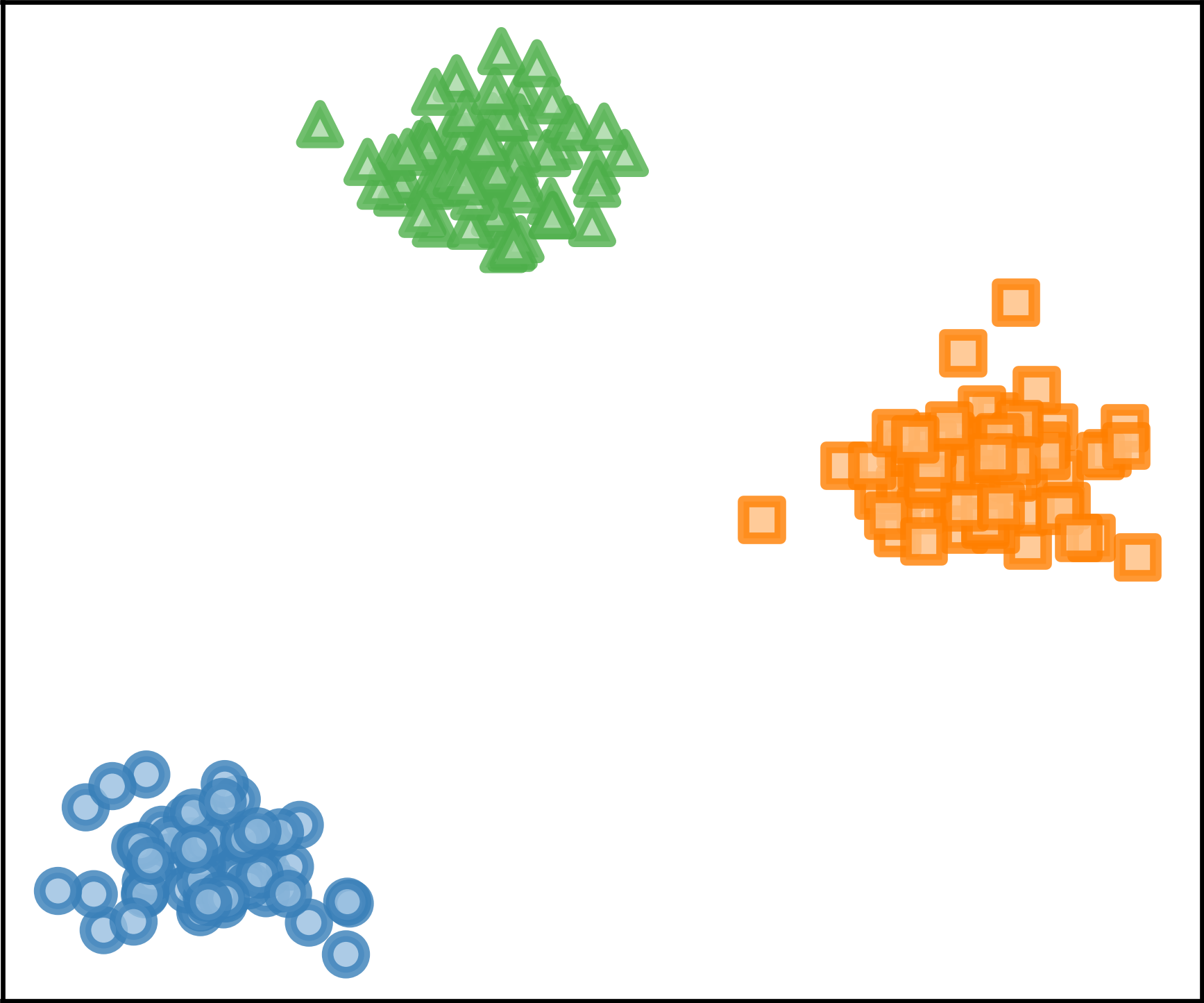}{}

\vspace{1em}

\image{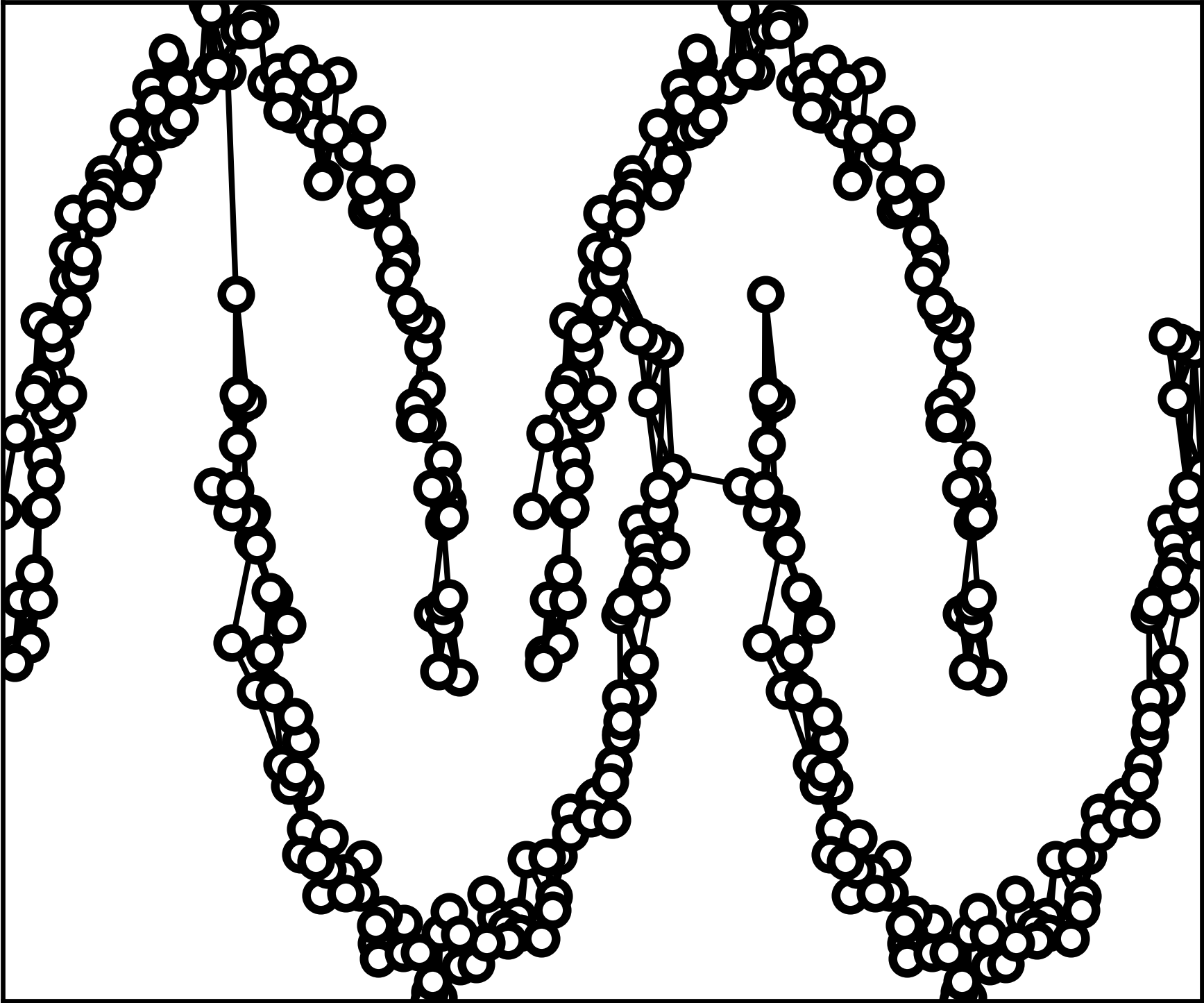}{}
\hfill
\image{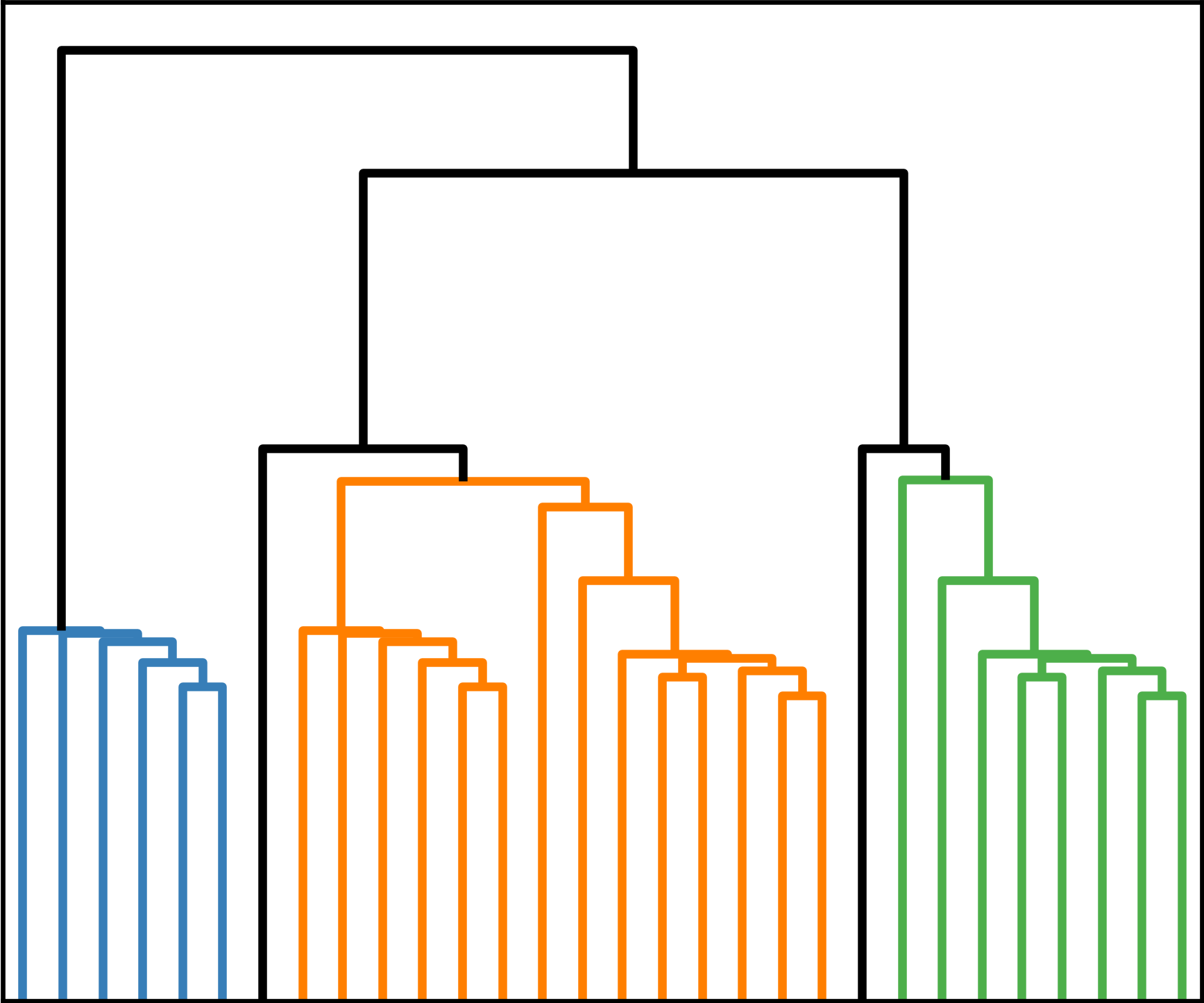}{}
\hfill
\image{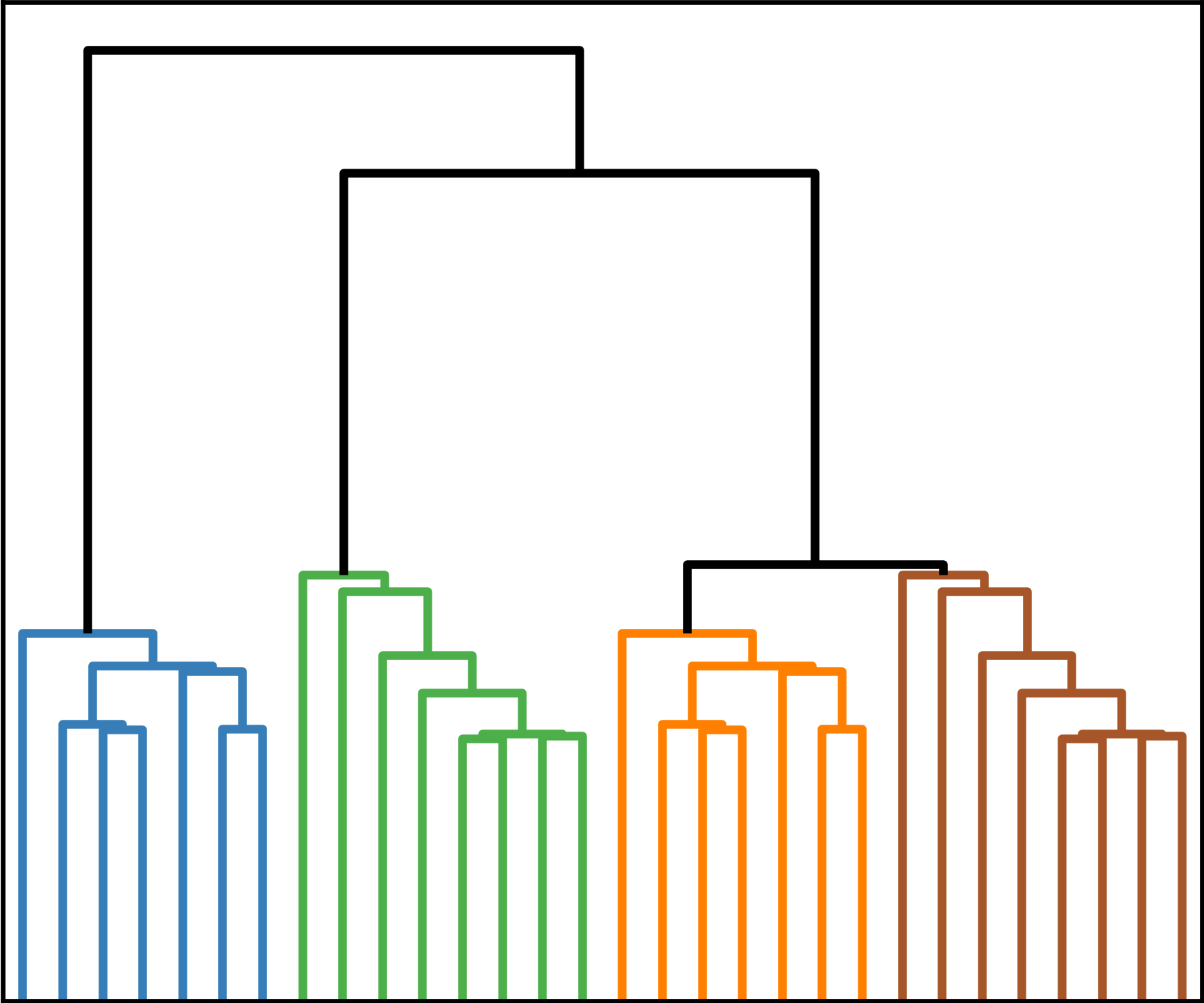}{}
\hfill
\image{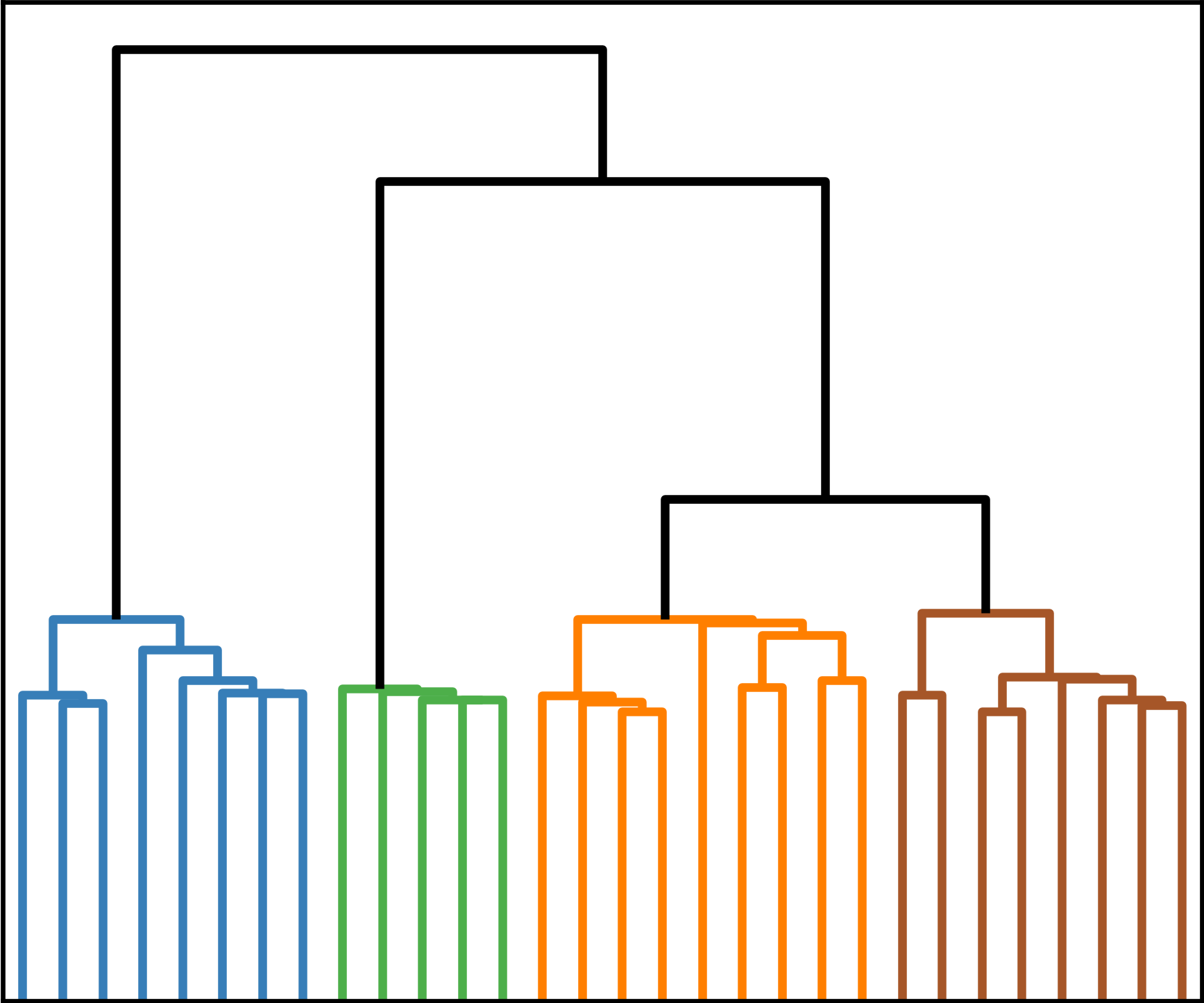}{}
\hfill
\image{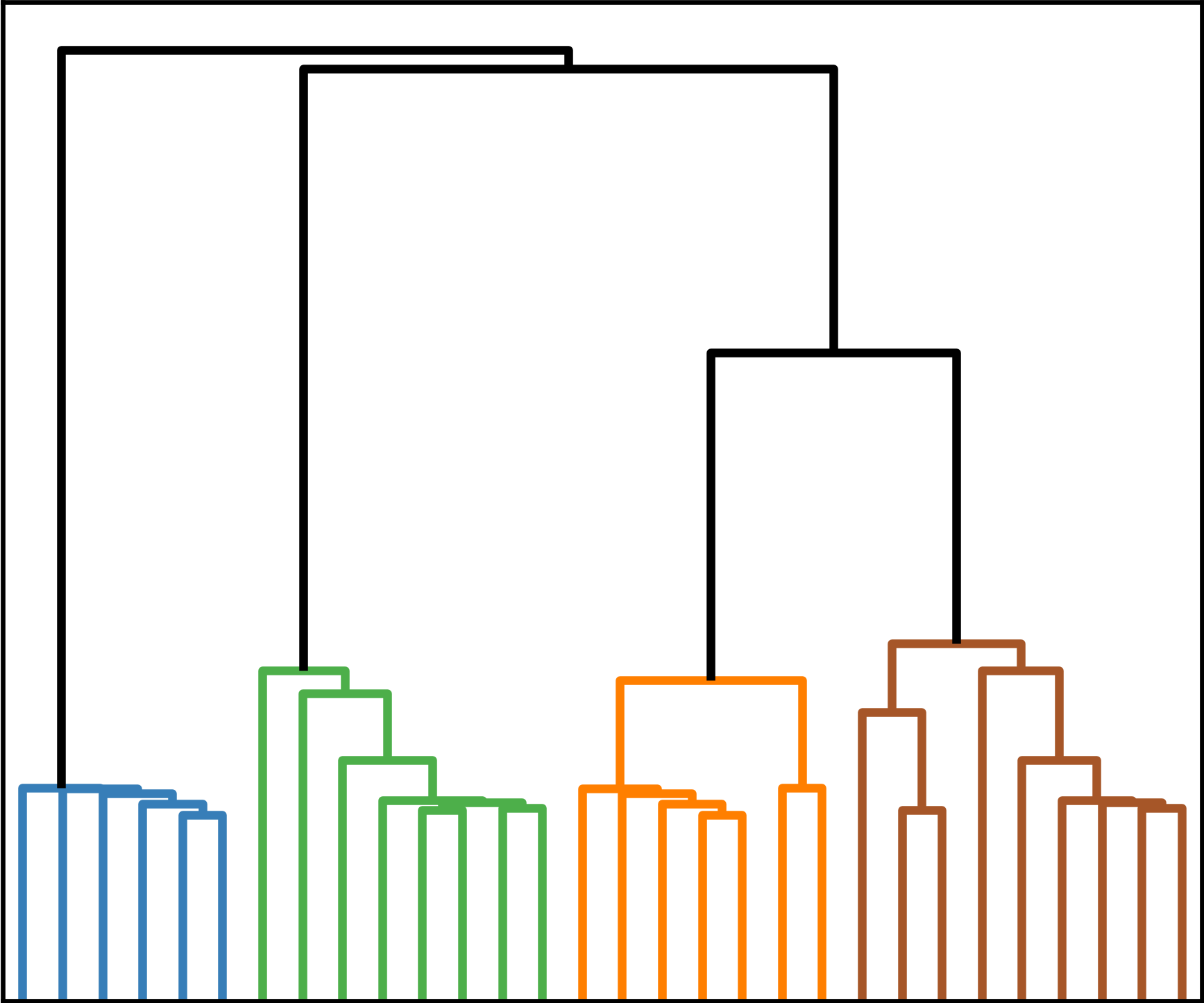}{}
\hfill
\image{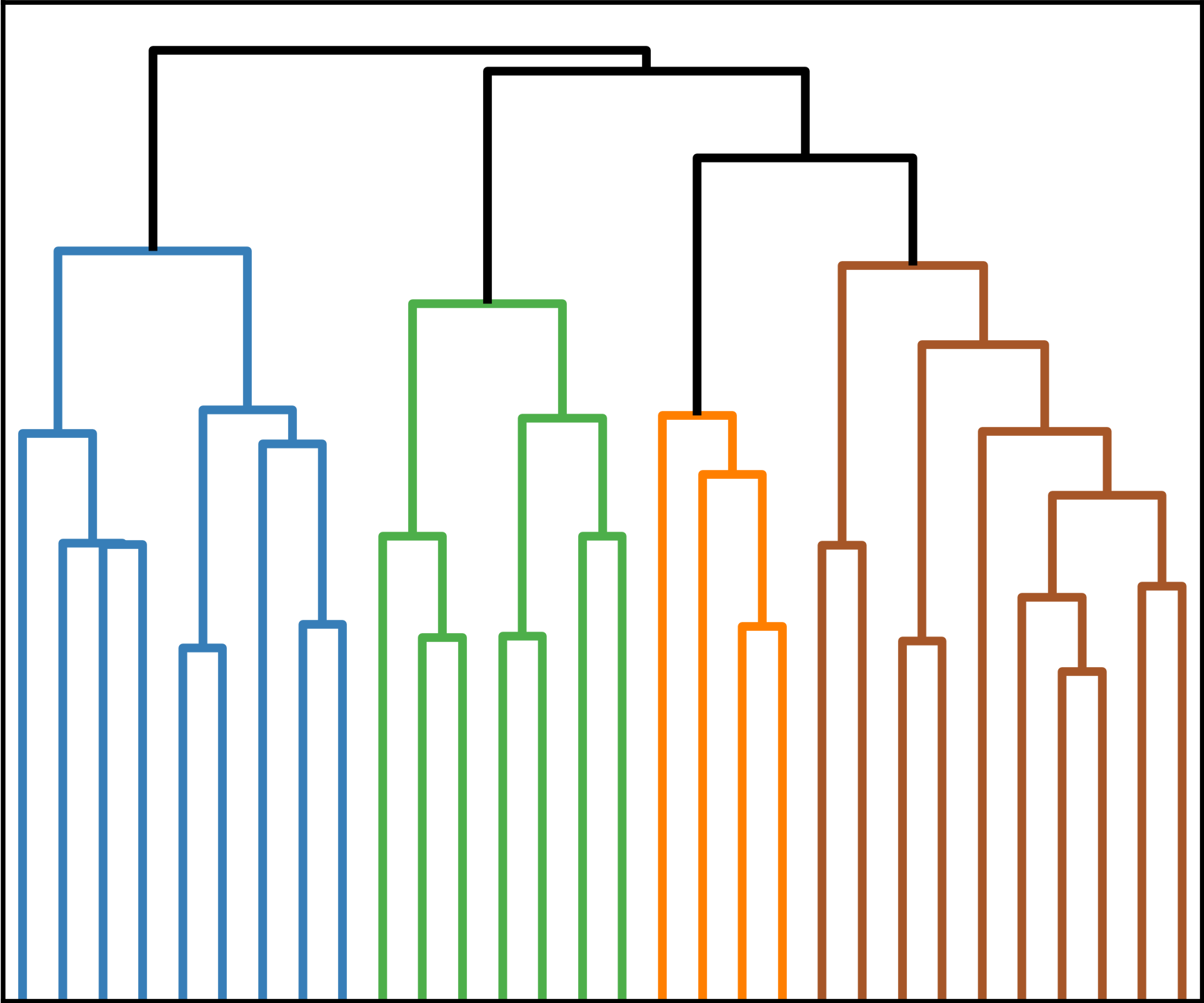}{}

\image[\caption{\footnotesize Graph/Labels}]{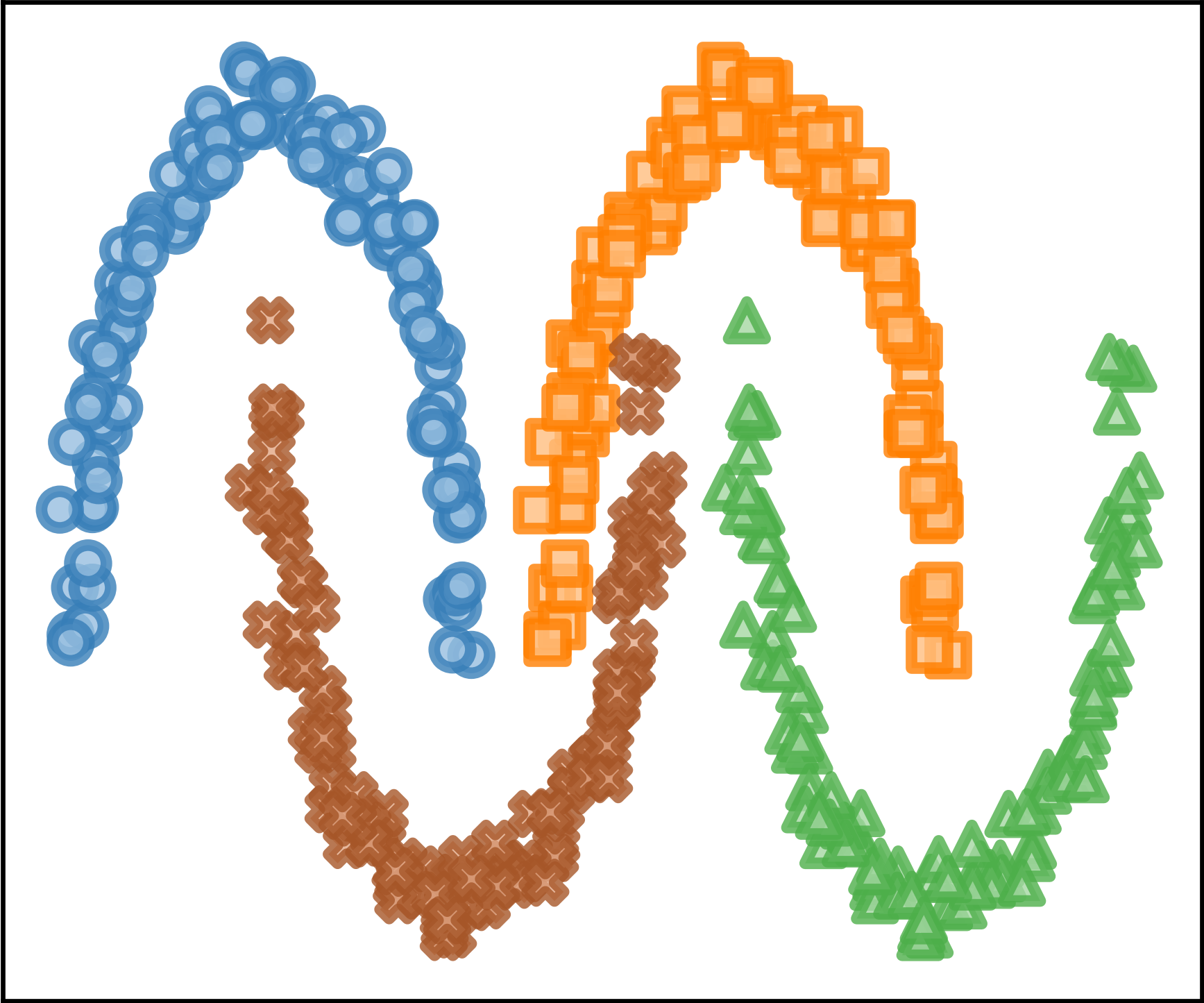}{}
\hfill
\image[\caption{\footnotesize Average link}]{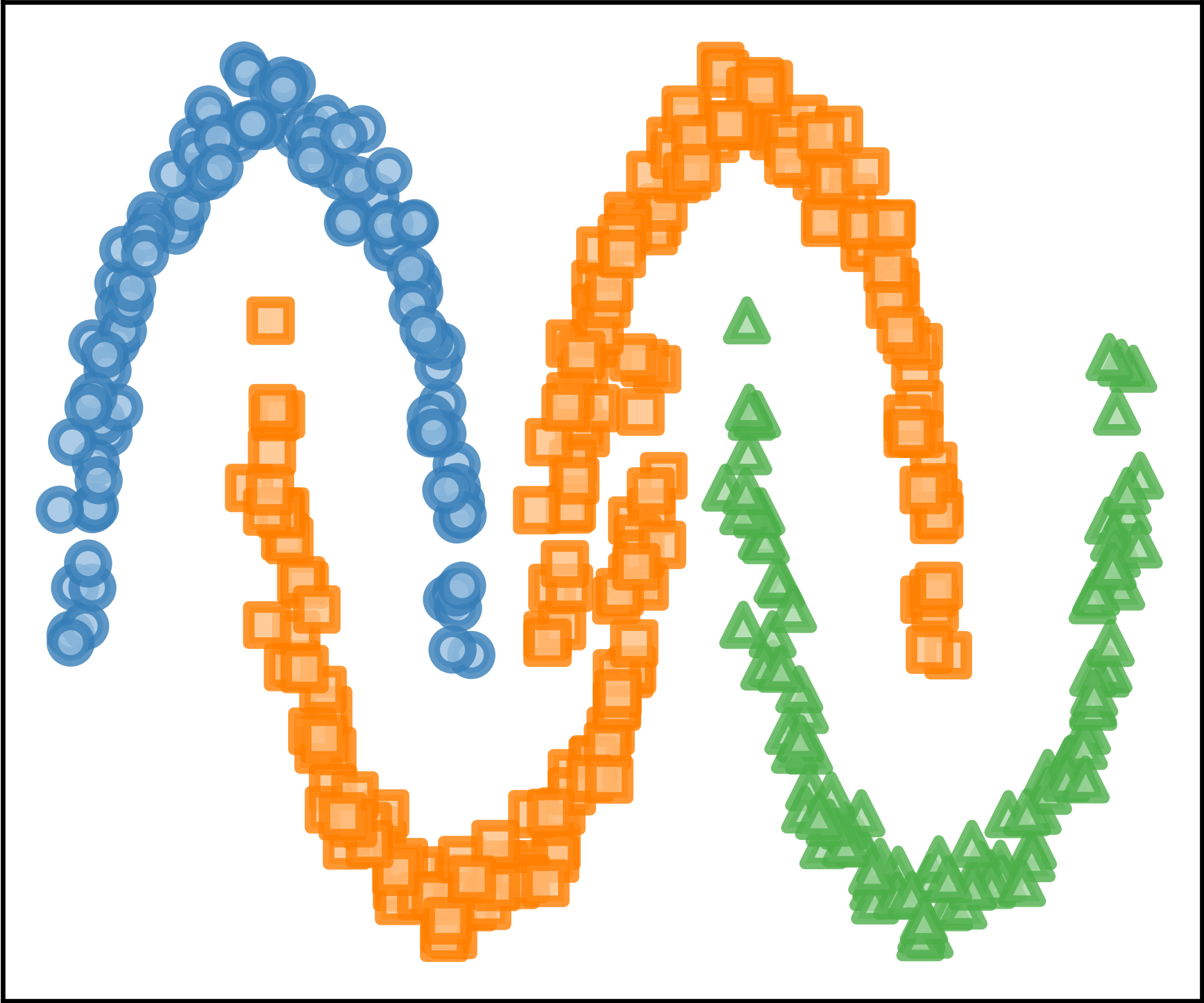}{\label{fig:more_examples:average}}
\hfill
\image[\caption{\footnotesize Closest}]{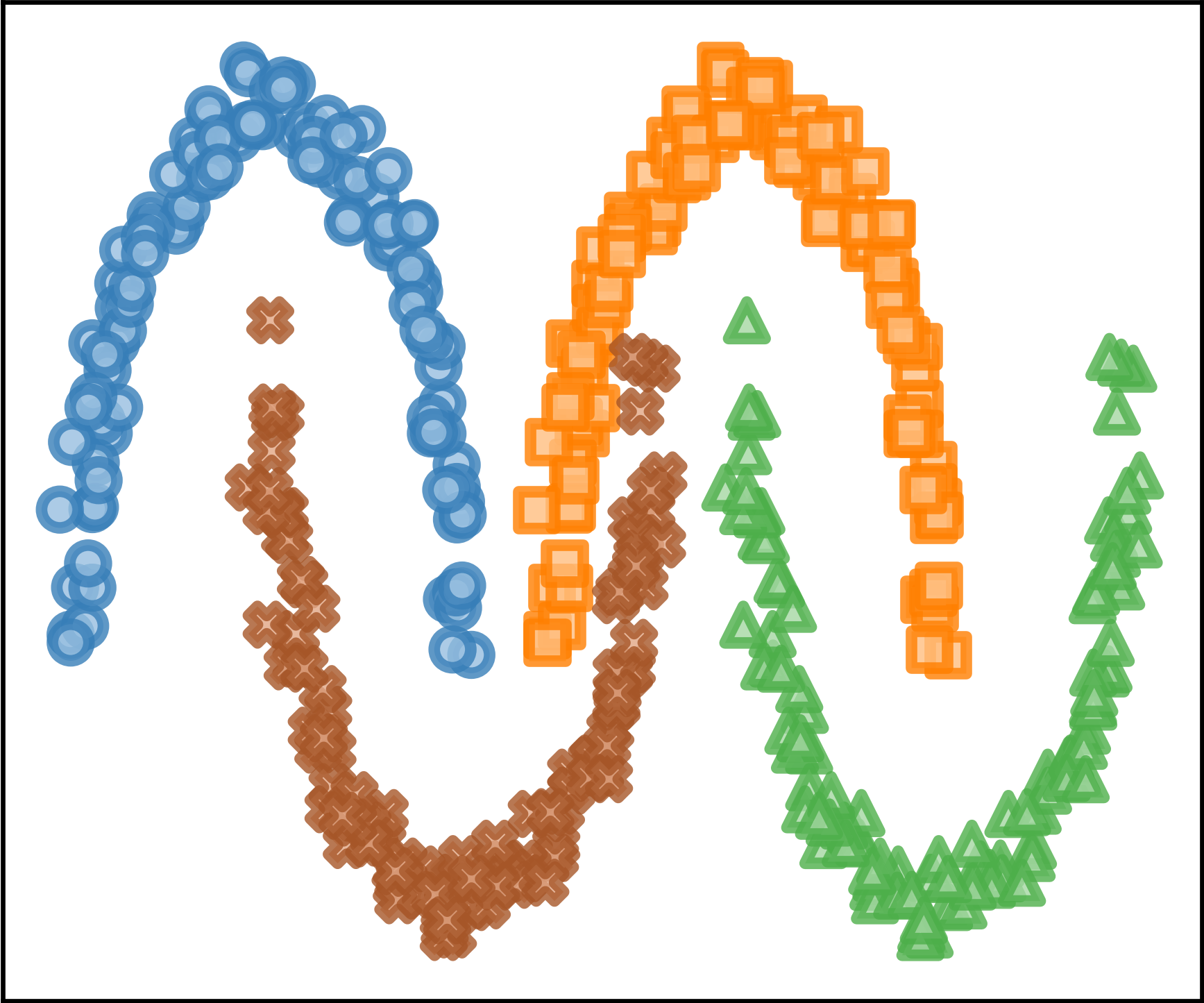}{\label{fig:more_examples:closest}}
\hfill
\image[\caption{\footnotesize Closest+Size}]{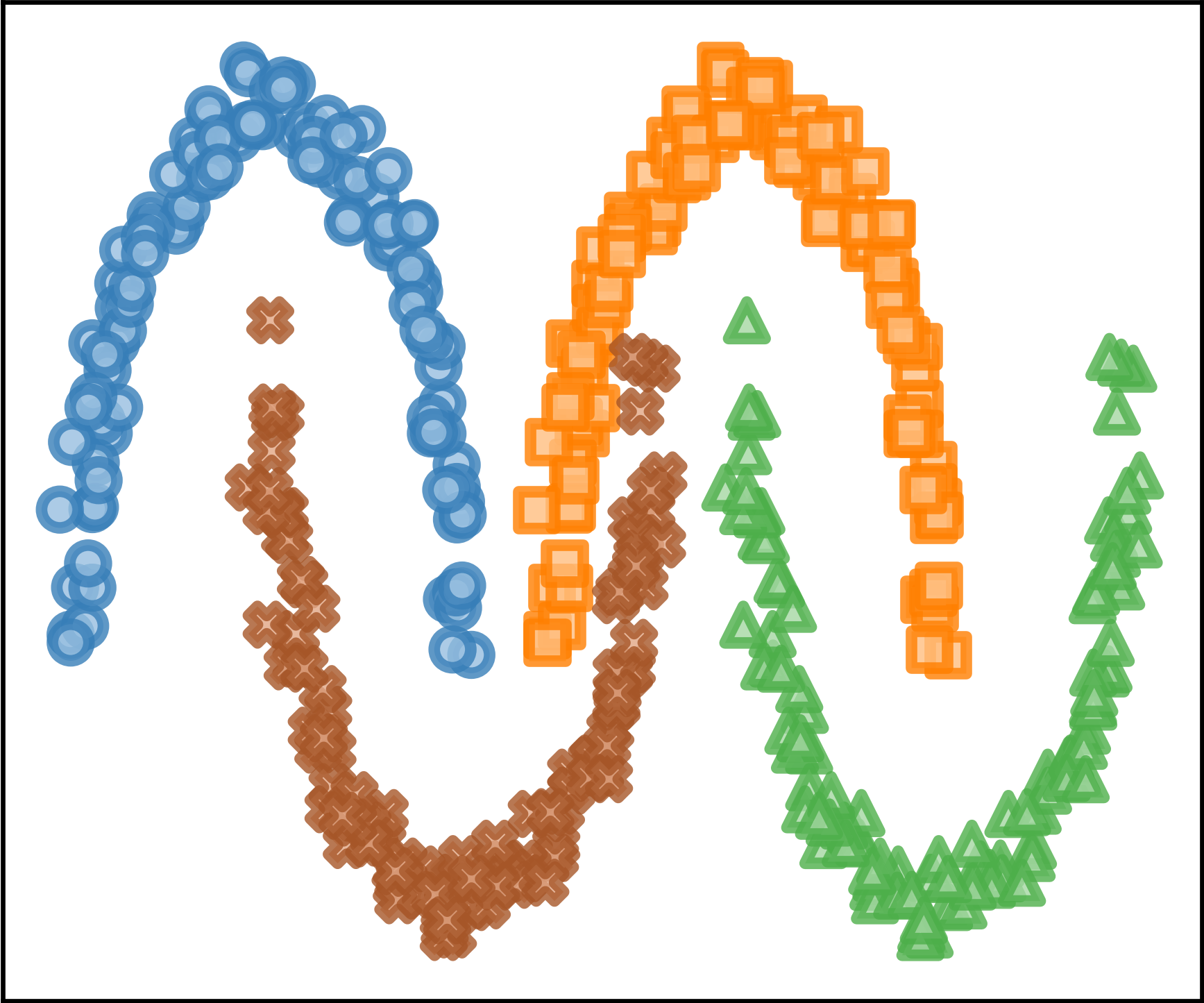}{\label{fig:more_examples:size}}
\hfill
\image[\caption{\scriptsize Closest+Triplet}]{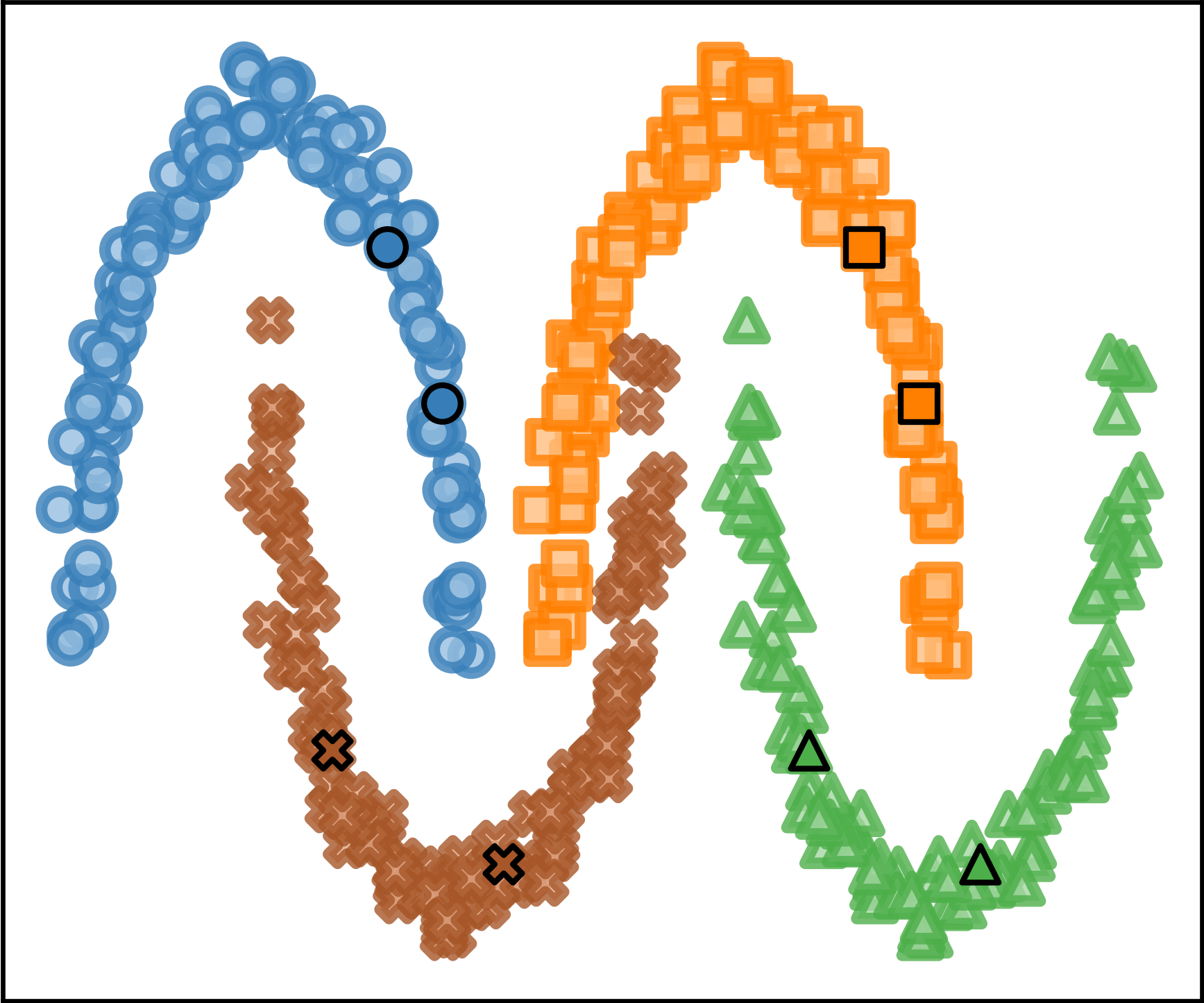}{\label{fig:more_examples:triplet}}
\hfill
\image[\caption{\footnotesize Dasgupta}]{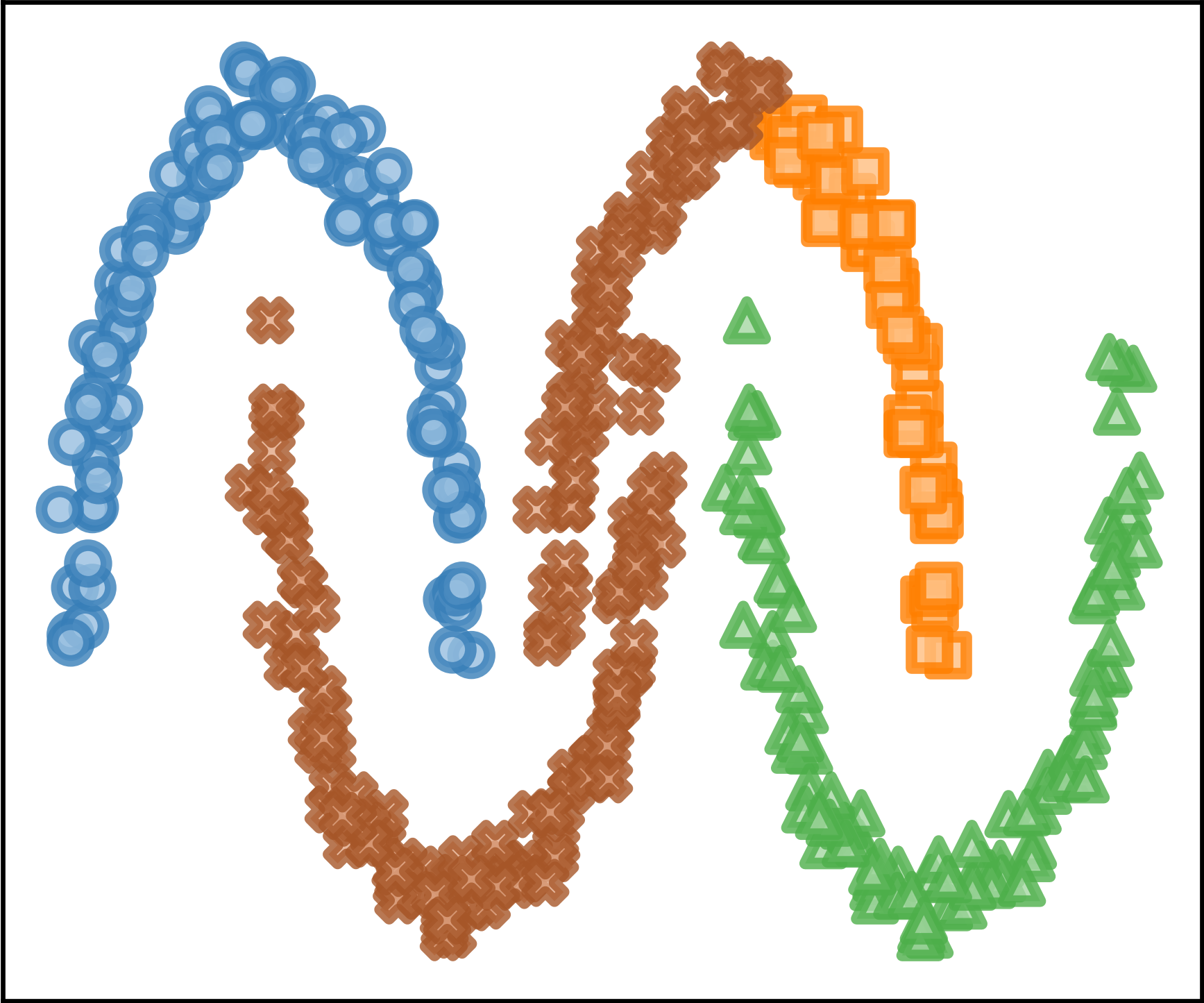}{\label{fig:more_examples:dasgupta}}

\caption{Illustrative examples of hierarchical clustering. \emph{Top rows}: Ultrametrics fitted to the input graph (only the top-30 non-leaf nodes are shown in the dendrograms). \emph{Bottom rows}: Assignments obtained by thresholding the ultrametrics at two, three, or four clusters.}
\label{fig:more_examples}
\end{figure}

Figure \ref{fig:more_examples} shows more illustrative examples of hierarchical clustering. For each dataset, we build a 5-nearest-neighbor graph, to which we add the edges of a minimum spanning tree to ensure the connectivity. Then, we perform hierarchical clustering on this graph, and we threshold the resulting ultrametric at the prescribed number of clusters. The column "Closest" is the solution to Algorithm \ref{algo:gradient_descent} with the cost function $J_{\rm closest}$. The column "Closest+Size" is the solution to Algorithm \ref{algo:gradient_descent} with the cost function $J_{\rm closest} + \lambda J_{\rm size}$ and $\lambda=0.1$, where the regularization is only applied to the top-10 dendogram nodes. The column "Closest+Triplet" is the solution to Algorithm \ref{algo:gradient_descent} with the cost function $J_{\rm closest} + \lambda J_{\rm triplet}$, $\lambda=1$, and $\alpha=3$. The column "Dasgupta" is the solution to Algorithm \ref{algo:gradient_descent} with the cost function $J_{\rm Dasgupta}$. For the optimization, we use the AMSGrad variation\Cite{j.2018on} of the ADAM method with step-size $0.1$.

\subsection{Average linkage approximates the closest ultrametric problem}
To help understand how the closest ultrametric problem relates to average linkage, note that an ultrametric takes a finite set of non-negative values $\{r_1, \dots, r_K\}$ with $K < |V|$. Hence, it can be represented as 
\begin{equation}
(\forall e \in E) \qquad u(e) = \sum_{k=1}^{K} r_k \, \mathcal{L}_k(e),
\end{equation}
where $\mathcal{L}=(\mathcal{L}_1,\dots,\mathcal{L}_K)$ are functions from $E$ to $\{0,1\}$ defining a hierarchical partition. In this setting, the cost function $J_{\rm closest}$ boils down to
\begin{equation}
\widetilde{J}_{\rm closest}(\mathcal{L}, r;\w) = \sum_{e \in E} \sum_{k=1}^{K} \mathcal{L}_k(e)  \big(r_k - \w(e)\big)^2,
\end{equation}
and, for a fixed hierarchical clustering $\bar{\mathcal{L}}$, the optimal altitudes are given by
\begin{equation}
\bar{r}_k = \frac{\sum_{e \in E} \bar{\mathcal{L}}_k(e) \w(e)}{\sum_{e \in E} \bar{\mathcal{L}}_k(e)}.
\end{equation}
This is exactly the criterion used by average linkage to build a hierarchical clustering. We can thus argue that the latter provides an approximate solution to the closest ultrametric problem. As a matter of fact, average linkage and Algorithm \ref{algo:gradient_descent} with $J_{\rm closest}$ produce structurally similar ultrametrics, as shown in Figures~\ref{fig:examples}~and~\ref{fig:more_examples} for illustrative examples of hierarchical clustering.

%% file: main.bbl
\begin{thebibliography}{54}
\providecommand{\natexlab}[1]{#1}
\providecommand{\url}[1]{\texttt{#1}}
\expandafter\ifx\csname urlstyle\endcsname\relax
  \providecommand{\doi}[1]{doi: #1}\else
  \providecommand{\doi}{doi: \begingroup \urlstyle{rm}\Url}\fi

\bibitem[Murtagh and Contreras(2012)]{Murtagh2012}
F.~Murtagh and P.~Contreras.
\newblock Algorithms for hierarchical clustering: an overview.
\newblock \emph{Data Mining and Knowledge Discovery}, 2\penalty0 (1):\penalty0
  86--97, 2012.

\bibitem[Sneath and Sokal(1962)]{Sneath1962}
P.~H.~A. Sneath and R.~R. Sokal.
\newblock Numerical taxonomy.
\newblock \emph{Nature}, 193:\penalty0 855--860, 1962.

\bibitem[Felsenstein(2003)]{Felsenstein2003}
J.~Felsenstein.
\newblock \emph{Inferring phylogenies}.
\newblock Sinauer Associates, 2003.

\bibitem[Gower and Ross(1969)]{Gower1969}
J.~C. Gower and G.~J.~S. Ross.
\newblock Minimum spanning trees and single linkage cluster analysis.
\newblock \emph{Journal of the Royal Statistical Society. Series C (Applied
  Statistics)}, 18\penalty0 (1):\penalty0 54--64, 1969.

\bibitem[Jardine and Sibson(1968)]{Jardine1968}
N.~Jardine and R.~Sibson.
\newblock The construction of hierarchic and non-hierarchic classifications.
\newblock \emph{The Computer Journal}, 11\penalty0 (2):\penalty0 177--184,
  1968.

\bibitem[Ward(1963)]{Ward1963}
J.~H. Ward.
\newblock Hierarchical grouping to optimize an objective function.
\newblock \emph{Journal of the American Statistical Association}, 58\penalty0
  (301):\penalty0 236--244, 1963.

\bibitem[de~Amorim(2015)]{deAmorim2015}
R.~C. de~Amorim.
\newblock Feature relevance in ward's hierarchical clustering using the lp
  norm.
\newblock \emph{Journal of Classification}, 32\penalty0 (1):\penalty0 46--62,
  2015.

\bibitem[Ackerman and Ben-David(2016)]{Ackerman2016}
M.~Ackerman and S.~Ben-David.
\newblock A characterization of linkage-based hierarchical clustering.
\newblock \emph{JMLR}, 17\penalty0 (231):\penalty0 1--17, 2016.

\bibitem[Dasgupta(2016)]{Dasgupta2016}
S.~Dasgupta.
\newblock A cost function for similarity-based hierarchical clustering.
\newblock In \emph{Proc. STOC}, pages 118--127, Cambridge, MA, USA, 2016.

\bibitem[Kobren et~al.(2017)Kobren, Monath, Krishnamurthy, and
  McCallum]{kobren2017}
A.~Kobren, N.~Monath, A.~Krishnamurthy, and A.~McCallum.
\newblock A hierarchical algorithm for extreme clustering.
\newblock In \emph{Proc. ACM SIGKDD}, pages 255--264, 2017.

\bibitem[Cohen-Addad et~al.(2017)Cohen-Addad, Kanade, and
  Mallmann-Trenn]{CohenAddadNIPS2017}
V.~Cohen-Addad, V.~Kanade, and F.~Mallmann-Trenn.
\newblock Hierarchical clustering beyond the worst-case.
\newblock In \emph{Proc. NeurIPS}, pages 6201--6209, 2017.

\bibitem[Chehreghani(2018)]{Chehreghani2018}
M.~H. Chehreghani.
\newblock Reliable agglomerative clustering.
\newblock \emph{Preprint arXiv:1901.02063}, 2018.

\bibitem[Bonald et~al.(2018)Bonald, Charpentier, Galland, and
  Hollocou]{Bonald2018}
T.~Bonald, B.~Charpentier, A.~Galland, and A.~Hollocou.
\newblock Hierarchical graph clustering using node pair sampling.
\newblock In \emph{KDD Workshop}, 2018.

\bibitem[Yarkony and Fowlkes(2015)]{YarkonyNIPS2015}
J.E. Yarkony and C.~Fowlkes.
\newblock Planar ultrametrics for image segmentation.
\newblock In \emph{Proc. NeurIPS}, pages 64--72, 2015.

\bibitem[Di~Summa et~al.(2015)Di~Summa, Pritchard, and Sanità]{DISUMMA2015}
M.~Di~Summa, D.~Pritchard, and L.~Sanità.
\newblock Finding the closest ultrametric.
\newblock \emph{DAM}, 180\penalty0 (10):\penalty0 70--80, 2015.

\bibitem[Roy and Pokutta(2016)]{RoyNIPS2016}
A.~Roy and S.~Pokutta.
\newblock Hierarchical clustering via spreading metrics.
\newblock In \emph{Proc. NeurIPS}, pages 2316--2324, 2016.

\bibitem[De~Soete(1984)]{DeSoete1984}
G.~De~Soete.
\newblock A least squares algorithm for fitting an ultrametric tree to a
  dissimilarity matrix.
\newblock \emph{PRL}, 2\penalty0 (3):\penalty0 133--137, 1984.

\bibitem[Ailon and Charikar(2011)]{Ailon2011}
N.~Ailon and M.~Charikar.
\newblock Fitting tree metrics: Hierarchical clustering and phylogeny.
\newblock \emph{{SIAM} Journal on Computing}, 40\penalty0 (5):\penalty0
  1275--1291, 2011.

\bibitem[Charikar and Chatziafratis(2017)]{Charikar2017}
M.~Charikar and V.~Chatziafratis.
\newblock Approximate hierarchical clustering via sparsest cut and spreading
  metrics.
\newblock In \emph{Proc. SODA}, pages 841--854, 2017.

\bibitem[Monath et~al.(2017)Monath, Kobren, and McCallum]{monath2017}
N.~Monath, A.~Kobren, and A~McCallum.
\newblock Gradient-based hierarchical clustering.
\newblock In \emph{NIPS 2017 Workshop on Discrete Structures in Machine
  Learning}, Long Beach, CA, 2017.

\bibitem[Hartigan(1985)]{Hartigan1985}
J.~A. Hartigan.
\newblock Statistical theory in clustering.
\newblock \emph{Journal of Classification}, 2\penalty0 (1):\penalty0 63--76,
  1985.

\bibitem[Neal(2003)]{Neal2003}
R.~Neal.
\newblock Density modeling and clustering using dirichlet diffusion trees.
\newblock In \emph{Bayesian Statistics}, volume~7, pages 619--629, 2003.

\bibitem[Vikram and Dasgupta(2016)]{vikram2016}
S.~Vikram and S.~Dasgupta.
\newblock Interactive bayesian hierarchical clustering.
\newblock In \emph{Proc. ICML}, volume~48, pages 2081--2090, New York, USA,
  2016.

\bibitem[Cohen-Addad et~al.(2018)Cohen-Addad, Kanade, Mallmann-Trenn, and
  Mathieu]{CohenAddad2018}
V.~Cohen-Addad, V.~Kanade, F.~Mallmann-Trenn, and C.~Mathieu.
\newblock Hierarchical clustering: Objective functions and algorithms.
\newblock In \emph{Proc. SODA}, pages 378--397, 2018.

\bibitem[Moseley and Wang(2017)]{Moseley2017}
B.~Moseley and J.~Wang.
\newblock Approximation bounds for hierarchical clustering: Average linkage,
  bisecting k-means, and local search.
\newblock In \emph{Proc. NeurIPS}, pages 3094--3103, 2017.

\bibitem[Charikar et~al.(2019)Charikar, Chatziafratis, and
  Niazadeh]{Charikar2019}
M.~Charikar, V.~Chatziafratis, and R.~Niazadeh.
\newblock Hierarchical clustering better than average-linkage.
\newblock In \emph{Proc. SODA}, pages 2291--2304, 2019.

\bibitem[Chatziafratis et~al.(2018)Chatziafratis, Niazadeh, and
  Charikar]{Chatziafratis2018}
V.~Chatziafratis, R.~Niazadeh, and M.~Charikar.
\newblock Hierarchical clustering with structural constraints.
\newblock In \emph{Proc. ICML}, volume~80, pages 774--783, Stockholm, Sweden,
  2018.

\bibitem[Turaga et~al.(2009)Turaga, Briggman, Helmstaedter, Denk, and
  Seung]{MALIS2009}
S.C. Turaga, K.L. Briggman, M.~Helmstaedter, W.~Denk, and H.S. Seung.
\newblock Maximin affinity learning of image segmentation.
\newblock In \emph{Proc. NeurIPS}, pages 1865--1873, 2009.

\bibitem[Arbelaez et~al.(2011)Arbelaez, Maire, Fowlkes, and
  Malik]{ArbelaezPAMI2011}
P.~Arbelaez, M.~Maire, C.~Fowlkes, and J.~Malik.
\newblock Contour detection and hierarchical image segmentation.
\newblock \emph{IEEE PAMI}, 33\penalty0 (5):\penalty0 898--916, 2011.

\bibitem[Maninis et~al.(2018)Maninis, Pont-Tuset, Arbel\'{a}ez, and
  Gool]{ManinisPAMI2018}
K.K. Maninis, J.~Pont-Tuset, P.~Arbel\'{a}ez, and L.~Van Gool.
\newblock Convolutional oriented boundaries: From image segmentation to
  high-level tasks.
\newblock \emph{IEEE PAMI}, 40\penalty0 (4):\penalty0 819--833, 2018.

\bibitem[{Funke} et~al.(2018){Funke}, {Tschopp}, {Grisaitis}, {Sheridan},
  {Singh}, {Saalfeld}, and {Turaga}]{DEEPMALIS}
J.~{Funke}, F.~D. {Tschopp}, W.~{Grisaitis}, A.~{Sheridan}, C.~{Singh},
  S.~{Saalfeld}, and S.~C. {Turaga}.
\newblock Large scale image segmentation with structured loss based deep
  learning for connectome reconstruction.
\newblock \emph{IEEE PAMI}, 99:\penalty0 1--12, 2018.

\bibitem[Ishikawa(2003)]{Ishikawa2003}
H.~Ishikawa.
\newblock Exact optimization for markov random fields with convex priors.
\newblock \emph{IEEE PAMI}, 25\penalty0 (10):\penalty0 1333--1336, October
  2003.

\bibitem[Pock et~al.(2008)Pock, Schoenemann, Graber, Bischof, and
  Cremers]{PockECCV2008}
T.~Pock, T.~Schoenemann, G.~Graber, H.~Bischof, and D.~Cremers.
\newblock A convex formulation of continuous multi-label problems.
\newblock In \emph{Proc. ECCV}, pages 792--805, Marseille, France, 2008.

\bibitem[{Pock} et~al.(2009){Pock}, {Chambolle}, {Cremers}, and
  {Bischof}]{PockCVPR2009}
T.~{Pock}, A.~{Chambolle}, D.~{Cremers}, and H.~{Bischof}.
\newblock A convex relaxation approach for computing minimal partitions.
\newblock In \emph{Proc. CVPR}, pages 810--817, Miami, FL, USA, June 2009.

\bibitem[Pock et~al.(2009)Pock, Cremers, Bischof, and Chambolle]{PockECCV2009}
T.~Pock, D.~Cremers, H.~Bischof, and A.~Chambolle.
\newblock An algorithm for minimizing the {M}umford-{S}hah functional.
\newblock In \emph{Proc. ICCV}, pages 1133--1140, September 2009.

\bibitem[M\"ollenhoff et~al.(2016)M\"ollenhoff, Laude, M\"oller, Lellmann, and
  Cremers]{MollenhoffCVPR2016}
T.~M\"ollenhoff, E.~Laude, M.~M\"oller, J.~Lellmann, and D.~Cremers.
\newblock Sublabel-accurate relaxation of nonconvex energies.
\newblock In \emph{Proc. CVPR}, pages 3948--3956, Las Vegas, NV, USA, June
  2016.

\bibitem[Foare et~al.(2018)Foare, Pustelnik, and Condat]{Foare2018}
M.~Foare, N.~Pustelnik, and L.~Condat.
\newblock Semi-linearized proximal alternating minimization for a discrete
  mumford-shah model.
\newblock \emph{Preprint hal-01782346}, 2018.

\bibitem[Carlsson and M{\'e}moli(2010)]{CarlssonJMLR2010}
G.~Carlsson and F.~M{\'e}moli.
\newblock Characterization, stability and convergence of hierarchical
  clustering methods.
\newblock \emph{JMLR}, 11:\penalty0 1425--1470, 2010.

\bibitem[Najman and Schmitt(1996)]{NajmanPAMI1996}
L.~Najman and M.~Schmitt.
\newblock Geodesic saliency of watershed contours and hierarchical
  segmentation.
\newblock \emph{IEEE PAMI}, 18\penalty0 (12):\penalty0 1163--1173, 1996.

\bibitem[K{\v{r}}iv{\'a}nek(1988)]{kvrivanek1988complexity}
M.~K{\v{r}}iv{\'a}nek.
\newblock The complexity of ultrametric partitions on graphs.
\newblock \emph{IPL}, 27\penalty0 (5):\penalty0 265--270, 1988.

\bibitem[Roy and Pokutta(2017)]{RoyJMLR2017}
A.~Roy and S.~Pokutta.
\newblock Hierarchical clustering via spreading metrics.
\newblock \emph{JMLR}, 18\penalty0 (88):\penalty0 1--35, 2017.

\bibitem[Najman et~al.(2013)Najman, Cousty, and Perret]{NajmanISMM2013}
L.~Najman, J.~Cousty, and B.~Perret.
\newblock Playing with kruskal: Algorithms for morphological trees in
  edge-weighted graphs.
\newblock In \emph{ISMM}, volume 7883, pages 135--146, 2013.

\bibitem[Bender and Farach-Colton(2000)]{Bender2000}
M.A. Bender and M.~Farach-Colton.
\newblock The lca problem revisited.
\newblock In Gaston~H. Gonnet and Alfredo Viola, editors, \emph{LATIN 2000:
  Theoretical Informatics}, pages 88--94. Springer Berlin Heidelberg, 2000.

\bibitem[Perret et~al.(2019)Perret, Chierchia, Cousty, Guimar{\~a}es, Kenmochi,
  and Najman]{Perret2019softwarex}
B.~Perret, G.~Chierchia, J.~Cousty, S.~J.~F. Guimar{\~a}es, Y.~Kenmochi, and
  L.~Najman.
\newblock Higra: Hierarchical graph analysis.
\newblock \emph{SoftwareX}, 10:\penalty0 1--6, 2019.

\bibitem[Paszke et~al.(2017)Paszke, Gross, Chintala, Chanan, Yang, DeVito, Lin,
  Desmaison, Antiga, and Lerer]{paszke2017automatic}
A.~Paszke, S.~Gross, S.~Chintala, G.~Chanan, E.~Yang, Z.~DeVito, Z.~Lin,
  A.~Desmaison, L.~Antiga, and A.~Lerer.
\newblock Automatic differentiation in {PyTorch}.
\newblock In \emph{NIPS Autodiff Workshop}, 2017.

\bibitem[{Dollár} and {Zitnick}(2015)]{DollarPAMI2015}
P.~{Dollár} and C.~L. {Zitnick}.
\newblock Fast edge detection using structured forests.
\newblock \emph{IEEE PAMI}, 37\penalty0 (8):\penalty0 1558--1570, 2015.

\bibitem[Carleo et~al.(2019)Carleo, Cirac, Cranmer, Daudet, Schuld, Tishby,
  Vogt-Maranto, and Zdeborov{\'a}]{carleo2019}
G.~Carleo, I.~Cirac, K.~Cranmer, L.~Daudet, M.~Schuld, N.~Tishby,
  L.~Vogt-Maranto, and L.~Zdeborov{\'a}.
\newblock Machine learning and the physical sciences.
\newblock \emph{Preprint arXiv:1903.10563}, 2019.

\bibitem[Grossman(1989)]{Grossman1989}
B.~Grossman.
\newblock The origin of the ultrametric topology of spin glasses.
\newblock \emph{Journal of Physics A: Mathematical and General}, 22\penalty0
  (1):\penalty0 L33--L39, January 1989.

\bibitem[Leuzzi(1999)]{leuzzi1999}
L.~Leuzzi.
\newblock Critical behaviour and ultrametricity of {I}sing spin-glass with
  long-range interactions.
\newblock \emph{Journal of Physics A: Mathematical and General}, 32\penalty0
  (8):\penalty0 1417, 1999.

\bibitem[Katzgraber and Hartmann(2009)]{katzgraber2009}
H.~G. Katzgraber and A.~K. Hartmann.
\newblock Ultrametricity and clustering of states in spin glasses: A
  one-dimensional view.
\newblock \emph{Physical review letters}, 102\penalty0 (3):\penalty0 037207,
  2009.

\bibitem[Katzgraber et~al.(2012)Katzgraber, J{\"o}rg, Krz{\k{a}}ka{\l}a, and
  Hartmann]{katzgraber2012}
H.~G. Katzgraber, T.~J{\"o}rg, F.~Krz{\k{a}}ka{\l}a, and A.~K. Hartmann.
\newblock Ultrametric probe of the spin-glass state in a field.
\newblock \emph{Physical Review B}, 86\penalty0 (18):\penalty0 184405, 2012.

\bibitem[Baviera and Virasoro(2015)]{baviera2015}
R.~Baviera and M.~A. Virasoro.
\newblock A method that reveals the multi-level ultrametric tree hidden in
  p-spin-glass-like systems.
\newblock \emph{Journal of Statistical Mechanics: Theory and Experiment},
  2015\penalty0 (12):\penalty0 P12007, 2015.

\bibitem[Jagannath(2017)]{jagannath2017}
A.~Jagannath.
\newblock Approximate ultrametricity for random measures and applications to
  spin glasses.
\newblock \emph{Communications on Pure and Applied Mathematics}, 70\penalty0
  (4):\penalty0 611--664, 2017.

\bibitem[Reddi et~al.(2018)Reddi, Kale, and Kumar]{j.2018on}
S.J. Reddi, S.~Kale, and S.~Kumar.
\newblock On the convergence of adam and beyond.
\newblock In \emph{Proc. ICLR}, 2018.

\end{thebibliography}
